\documentclass[utf8]{FrontiersinHarvard} % for articles in journals using the Harvard Referencing Style (Author-Date), for Frontiers Reference Styles by Journal: https://zendesk.frontiersin.org/hc/en-us/articles/360017860337-Frontiers-Reference-Styles-by-Journal
\usepackage{url,hyperref,lineno,microtype,subcaption}
\usepackage{booktabs} 
\usepackage{stmaryrd}
\usepackage{graphicx}

\definecolor{car}{RGB}{100,150,245}
\definecolor{bicycle}{RGB}{100,230,245}
\definecolor{motorcycle}{RGB}{30,60,150}
\definecolor{truck}{RGB}{80,30,180}
\definecolor{othervehicle}{RGB}{100,80,250}
\definecolor{person}{RGB}{255,30,30}
\definecolor{bicyclist}{RGB}{255,40,200}
\definecolor{motorcyclist}{RGB}{150,30,90}
\definecolor{road}{RGB}{255,0,255}
\definecolor{parking}{RGB}{255,150,255}
\definecolor{sidewalk}{RGB}{75,0,75}
\definecolor{otherground}{RGB}{175,0,75}
\definecolor{building}{RGB}{255,200,0}
\definecolor{fence}{RGB}{255,120,50}
\definecolor{vegetation}{RGB}{0,175,0}
\definecolor{trunk}{RGB}{135,60,0}
\definecolor{terrain}{RGB}{150,240,80}
\definecolor{pole}{RGB}{255,240,150}
\definecolor{trafficsign}{RGB}{255,0,0}
\usepackage{float}

\def\firstAuthorLast{Zhang {et~al.}} % use et al. for >1 author

\def\Authors{%
Dunxing Zhang\,$^{3,\dagger}$, %
Jiachen Lu\,$^{3,\dagger}$, %
Han Yang\,$^{1,2,*}$, %
Lei Bao\,$^{1,2}$ and %
Bo Song\,$^{1,2}$%
}

\def\Address{%
$^{1}$National Science Center for Earthquake Engineering, Tianjin University, Tianjin 300350, China\\
$^{2}$School of Civil Engineering, Tianjin University, Tianjin 300350, China\\
$^{3}$Chair of Robotics, Artificial Intelligence and Real-time Systems, Technical University of Munich, Munich, Germany
}

\def\corrAuthor{Han Yang}
\def\corrEmail{professorhansolo@tju.edu.cn} % TODO: 

\DeclareGraphicsExtensions{.pdf,.jpg,.jpeg,.png}

\begin{document}
\onecolumn
\firstpage{1}

\title[3D Semantic Scene Completion with Refinement
]{Enhancing 3D Semantic Scene Completion with Refinement Module} 

\author[\firstAuthorLast ]{\Authors} %This field will be automatically populated
\address{} %This field will be automatically populated
\correspondance{} %This field will be automatically populated

\extraAuth{}% If there are more than 1 corresponding author, comment this line and uncomment the next one.
%\extraAuth{corresponding Author2 \\ Laboratory X2, Institute X2, Department X2, Organization X2, Street X2, City X2 , State XX2 (only USA, Canada and Australia), Zip Code2, X2 Country X2, email2@uni2.edu}
\maketitle

\begin{abstract}
We propose ESSC-RM, a plug-and-play \textbf{E}nhancing framework for \textbf{S}emantic \textbf{S}cene \textbf{C}ompletion with a \textbf{R}efinement \textbf{M}odule, which can be seamlessly integrated into existing SSC models. ESSC-RM operates in two phases: a baseline SSC network first produces a coarse voxel prediction, which is subsequently refined by a 3D U-Net–based Prediction Noise-Aware Module (PNAM) and Voxel-level Local Geometry Module (VLGM) under multiscale supervision. Experiments on SemanticKITTI show that ESSC-RM consistently improves semantic prediction performance. When integrated into CGFormer and MonoScene, the mean IoU increases from $16.87\%$ to $17.27\%$ and from $11.08\%$ to $11.51\%$, respectively. These results demonstrate that ESSC-RM serves as a general refinement framework applicable to a wide range of SSC models. Project page: \url{https://github.com/LuckyMax0722/ESSC-RM} and \url{https://github.com/LuckyMax0722/VLGSSC}.
\end{abstract}

\section{Introduction}\label{chapter:introduction}

Accurate 3D scene understanding is fundamental to autonomous driving, robotics, and embodied perception, where downstream tasks such as detection, reconstruction, mapping, and planning rely on complete geometric and semantic representations of the environment \cite{2019_DL43DPC,2020_SAD,9787342,2022_3DOD4AD,2024_SADP,10584449}. However, real-world sensors (LiDAR and RGB cameras) provide only sparse, noisy, and partial observations due to occlusions, limited resolution, restricted field of view, and missing depth information, resulting in incomplete voxelized scenes \cite{2021_3DSSC,10292927}. To address this, 3D  semantic scene completion (SSC) aims to jointly infer voxel occupancy and semantic labels, a task first formalized by SSCNet \cite{2016_SSCNet}.

Despite extensive progress in both LiDAR-based \cite{2020_LMSCNet,2020_JS3CNet,2023_SCPNet,2024_TALoS} and vision-based SSC \cite{2021_MonoScene,2023_VoxFormer,2023_Symphonies,2023_CGFormer}, a considerable gap remains between predictions and ground truth. LiDAR-based models suffer from sparsity; BEV-based methods \cite{2021_SSASC} lose fine-grained details; RGB-based approaches degrade due to depth ambiguity and unclear 2D--3D projection \cite{2024_SemCity}; and distillation pipelines depend heavily on task-specific teacher designs \cite{2023_SCPNet}. Moreover, SSC architectures differ substantially, making it difficult to develop a unified refinement strategy that generalizes across models without modifying their internal structures.

To bridge these limitations, this paper proposes \textbf{ESSC-RM}, a unified coarse-to-fine refinement framework that directly enhances the voxel predictions of arbitrary SSC models. ESSC-RM performs multi-scale geometric–semantic aggregation, integrates auxiliary priors, and introduces a model-agnostic refinement pipeline that requires no architectural modification to the baseline. It supports both end-to-end joint training and fully independent plug-and-play deployment.

\textbf{The main contributions of this paper are as follows:}
\begin{itemize}
    \item We introduce \textbf{ESSC-RM}, a general refinement framework designed to improve heterogeneous SSC baselines via coarse-to-fine multi-scale error reduction, applicable to both LiDAR-based and vision-based methods.
    \item We develop two complementary training paradigms: a \emph{joint training} mode that co-optimizes the refinement and baseline networks, and a \emph{separate training} mode enabling true plug-and-play enhancement without modifying the original SSC architecture.
    \item We propose a neighborhood-attention-based multi-scale aggregation module that adaptively fuses geometric and semantic features, improving voxel-level reasoning across scales.
    \item We introduce a novel vision–language guidance module that injects text-derived semantic priors to compensate for missing geometric cues and ambiguous visual projections, enhancing cross-modal scene understanding.
    \item Extensive experiments on SemanticKITTI~\cite{2019_SemanticKITTI} demonstrate that \textbf{ESSC-RM} consistently improves strong baselines such as CGFormer and MonoScene, validating its generality, flexibility, and effectiveness.
\end{itemize}

% !TeX spellcheck = en_US
\section{Related Work}\label{chapter:Related_Work}%

In this section, we review \emph{LiDAR- and camera-based 3D perception}, then summarize advances in \emph{3D SSC}, and finally discuss recent progress in \emph{vision–language models (VLMs)} and \emph{text-driven multimodal fusion}.

\subsection{LiDAR-based 3D Perception}
LiDAR provides accurate 3D geometry for autonomous driving perception, enabling detection, tracking, and mapping, and has become a core sensing modality \cite{2019_DL43DPC,2020_SAD,2022_3DOD4AD,2024_SADP,2022_LiDAR_Survey,2025_LiDAR_Survey}.

Early point-based and voxel-based detectors—PointNet \cite{2016_PointNet}, VoxelNet \cite{2017_VoxelNet}, SECOND \cite{2018_SECOND}, PointPillars \cite{2018_PointPillars}, PointRCNN \cite{2018_PointRCNN}, PV-RCNN \cite{2019_PVRCNN}, and Voxel R-CNN \cite{2020_VoxelRCNN}—established effective feature extraction paradigms. Tracking frameworks such as AB3DMOT \cite{2020_AB3DMOT} and \cite{2023_3DMOT} leverage motion models and geometric association. Semantic segmentation approaches including PointNet++ \cite{2017_PointNet_pp}, RangeNet++ \cite{2019_RangeNet_pp}, and Cylinder3D \cite{2020_Cylinder3D} demonstrate point-based, projection-based, and cylindrical-voxel inference strategies.

% These methods offer geometric priors and representations foundational to LiDAR-based SSC.

\subsection{Camera-based 3D Perception}
Camera-based perception offers a cost-efficient alternative with rich semantic cues. Monocular approaches extend 2D detectors \cite{2019_M3DRPN,2019_CenterNet,2018_ROI_10D} or rely on pseudo-depth and geometric priors \cite{2018_MultiFusion,2018_Pseudo_LiDAR,2014_3DB,2016_3DBB,2018_JM3D}, yet remain affected by depth ambiguity. Stereo-based methods \cite{2018_PyramidStero,2019_SteroRCNN,2019_Pseudo_LiDAR,2020_DSGN} mitigate this by enforcing geometric consistency \cite{2023_3DOD}.

With multi-camera setups becoming standard, multi-view 3D detection methods have evolved rapidly. LSS-based pipelines \cite{2020_LSS,2021_BEVDet} lift image features to Bird's-Eye View (BEV), while transformer-based designs such as DETR3D \cite{2021_DETR3D} and BEVFormer \cite{2022_BEVFormer} aggregate cross-view features using 3D object queries. Spatiotemporal attention mechanisms \cite{2017_AIAYN,2022_SDETR,2023_3DOD} further enhance robustness.

\subsection{Semantic Scene Completion}
SSC jointly predicts occupancy and voxel-level semantics. SSCNet \cite{2016_SSCNet} established the task on indoor data \cite{2012_NYUDV2}; outdoor datasets such as KITTI and SemanticKITTI \cite{2012_KITTI,2019_SemanticKITTI,2021_SemanticKITTI_2,2024_SSCBench} introduce sparsity and large-scale variability.

\subsection{Vision–Language Models}
Vision–language models (VLMs) provide strong semantic priors through aligned image–text representations \cite{2025_VLME2E}. CLIP \cite{2021_CLIP} and EVACLIP \cite{2023_EVACLIP,2024_EVACLIP} learn powerful contrastive embeddings, while LongCLIP \cite{2024_LongCLIP} and JinaCLIP \cite{2024_JinaCLIP,2024_JinaCLIP_2} improve long-text modeling.

Models such as BLIP2 \cite{2023_BLIP2}, InstructBLIP \cite{2023_InstructBLIP}, MiniGPT-4 \cite{2023_MiniGPT}, and LLaVA \cite{2023_LLaVA,2024_LLaVA_2} leverage frozen Large Language Models (LLMs) to build efficient multimodal reasoning pipelines \cite{2024_GPT4}. Text-conditioned segmentation models such as LSeg \cite{2022_LSEG} and Grounded-SAM \cite{2024_GSAM} further highlight the utility of text in perception tasks \cite{2023_GDINO,2023_SAM}.

\subsection{Multimodal Fusion and Text Modality}
Multimodal fusion traditionally combines 3D geometry (LiDAR, stereo) with rich 2D semantics. With the emergence of LLMs and VLMs, text has become a scalable, low-cost semantic modality for describing road scenes \cite{2024_MMAF,2025_VLME2E}.

Attention-based fusion \cite{2017_AIAYN,9546775}—as in \cite{2020_CMA,cao2024embracing,2025_CDScene}—captures long-range cross-modal dependencies but can be computationally heavy. Learnable fusion strategies such as Text-IF \cite{2024_Text_IF} and VLScene \cite{2025_VLScene} use trainable coefficients to balance visual and linguistic cues.

\section{Methodology}\label{sec:methodology}

ESSC-RM refines the coarse voxel predictions produced by any SSC backbone. 
We now present the problem formulation and describe the architecture components of our refinement module, including the 3D U-Net backbone, the progressive neighbourhood attention module (PNAM), and the vision–language guidance module (VLGM), as illustrated in Figure~\ref{fig:method_overall}.

\subsection{Problem Statement}\label{sec:problem}

Given an RGB image $\mathbf{I}_{t} \in \mathbb{R}^{H \times W \times 3}$ and a LiDAR point cloud $\mathbf{P}_{t} \in \mathbb{R}^{N \times 3}$ at time $t$, 3D SSC aims to predict a dense semantic voxel grid $\hat{\mathbf{Y}}_{t} \in \{c_{0}, c_{1}, \dots, c_{C}\}^{H \times W \times Z}$ defined in the vehicle coordinate system, where each voxel is either empty ($c_{0}$) or belongs to one of the $C$ semantic classes $\{c_{1}, \dots, c_{C}\}$ and $H,W,Z$ denote the voxel grid dimensions. A standard SSC backbone learns $\hat{\mathbf{Y}}_{t} = f_{\theta}(\mathbf{I}_{t}, \mathbf{P}_{t})$, but the coarse prediction $\hat{\mathbf{Y}}_{t}$ often exhibits broken surfaces, incomplete structures, and semantic confusions. We therefore introduce a refinement module $g_{\phi}$ that treats $\hat{\mathbf{Y}}_{t}$ as a noisy discrete volume and outputs a refined prediction $\hat{\mathbf{Y}}_{t}^{\prime} = g_{\phi}(\hat{\mathbf{Y}}_{t}, \text{aux})$, where \emph{aux} denotes additional cues (multi-scale voxel features and text semantics) extracted within the refinement module. The objective is to bring $\hat{\mathbf{Y}}_{t}^{\prime}$ closer to the ground truth $\mathbf{Y}_{t}$ in both geometry and semantics while remaining compatible with heterogeneous SSC backbones.

\subsection{Overall Architecture}\label{sec:overall}

\begin{figure}[h!]
  \centering
  \includegraphics[width=\textwidth]{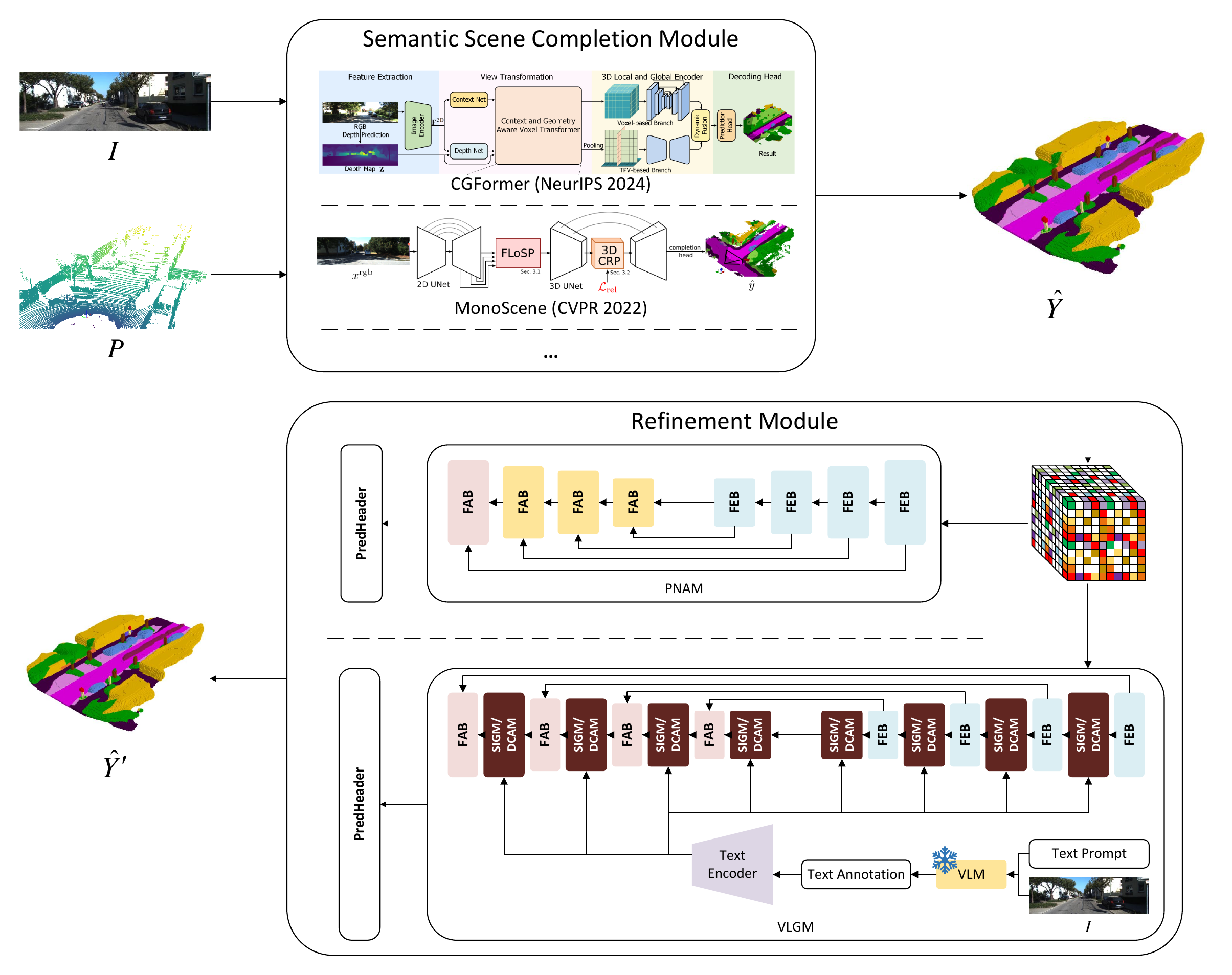}
  \caption{Overall architecture of ESSC-RM.
  An SSC backbone first predicts a coarse semantic voxel grid from image and/or LiDAR input.
  The refinement module embeds the discrete voxel labels into continuous features and processes them with a 3D U-Net enhanced by PNAM and VLGM, finally producing a refined semantic volume $\hat{\mathbf{Y}}^{\prime}$.}
  \label{fig:method_overall}
\end{figure}

As shown in Figure~\ref{fig:method_overall}, ESSC-RM has two decoupled parts:
\begin{itemize}
  \item \textbf{SSC backbone:} maps $(\mathbf{I}_{t}, \mathbf{P}_{t})$ to a coarse voxel grid $\hat{\mathbf{Y}}_{t}$.
  \item \textbf{Refinement module:} operates purely in voxel space, refining $\hat{\mathbf{Y}}_{t}$ into $\hat{\mathbf{Y}}_{t}^{\prime}$ using multi-scale U-Net features, neighbourhood attention, and vision–language guidance.
\end{itemize}

This separation allows us to plug in backbones of different quality while focusing the design of $g_{\phi}$ on correcting geometric and semantic errors at the voxel level using additional structural and semantic cues.

\subsection{SSC Backbone}

ESSC-RM is model-agnostic and can refine the output of any SSC backbone. 
To demonstrate generality, we instantiate two monocular SSC models with 
different coarse prediction qualities: CGFormer~\cite{2023_CGFormer} and 
MonoScene~\cite{2021_MonoScene}. CGFormer represents a strong backbone 
with accurate voxel lifting, while MonoScene produces notably noisier 
volumes, providing a more challenging setting for refinement. All 
architectural details follow the original papers, as our refinement module 
does not modify or depend on the internal design of the backbone.

\subsection{3D U-Net Refinement Backbone}\label{sec:3dunet}

\begin{figure}[h!]
  \centering
  \includegraphics[width=\textwidth]{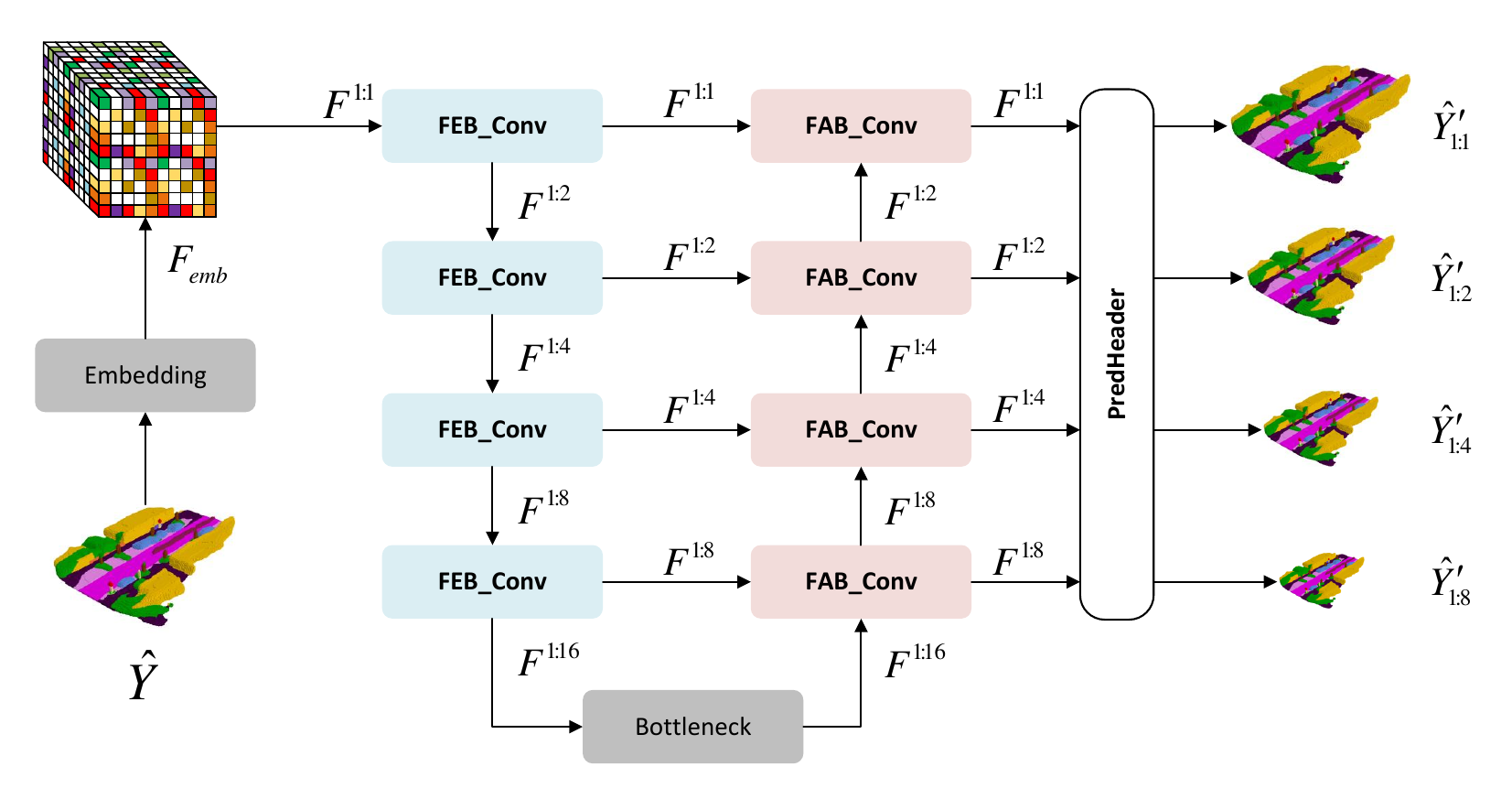}
  \caption{\textbf{Three-dimensional U-shaped neural network (3D U-Net) backbone used in the refinement module.}
  The encoder consists of stacked feature encoding blocks (FEBs) that progressively downsample the input and extract multi-scale features, a bottleneck aggregates global context, and the decoder consists of stacked feature aggregation blocks (FABs) that progressively upsample the features and predict semantic voxel outputs at four spatial scales.}
  \label{fig:3DUNet_arch}
\end{figure}

\begin{figure}[h!]
  \centering
  \includegraphics[width=\textwidth]{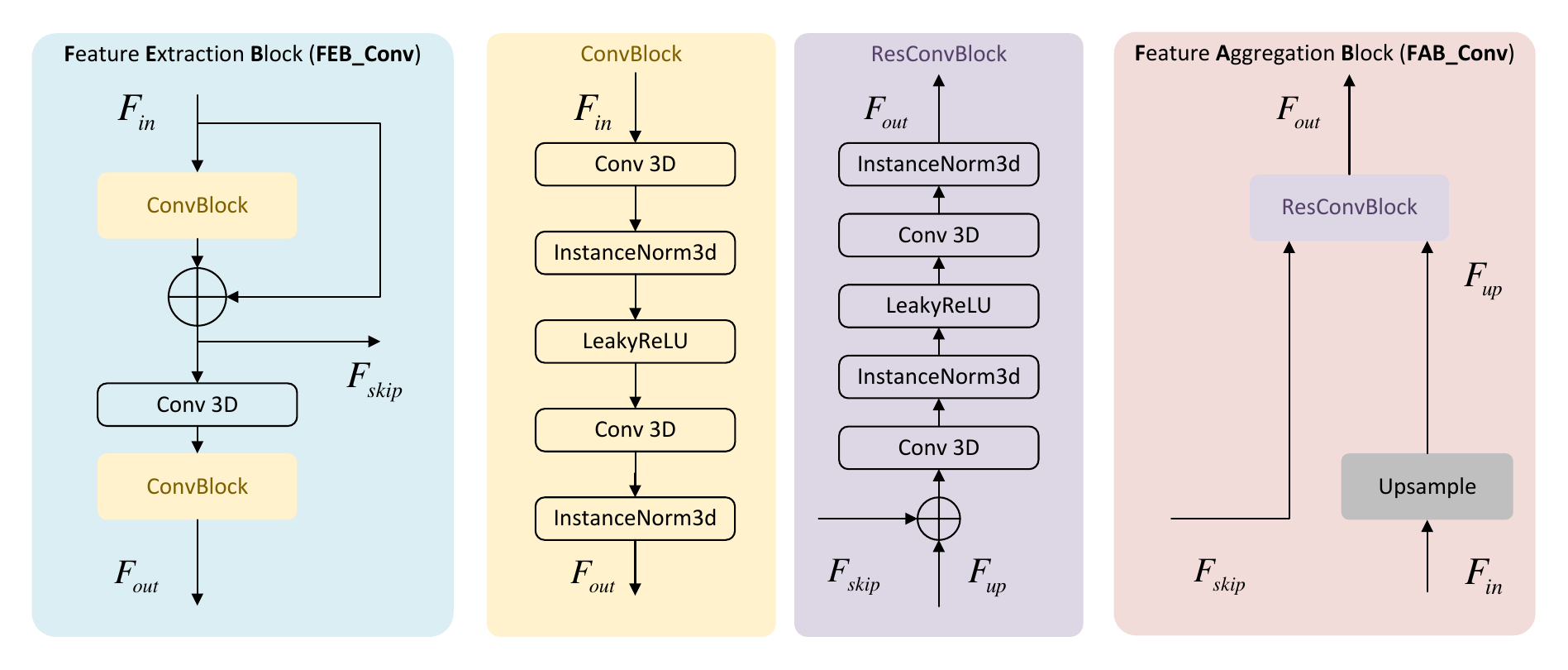}
  \caption{\textbf{Core components of the three-dimensional U-shaped neural network (3D U-Net).}
  The feature encoding block (FEB) applies three-dimensional convolutions, normalization, and residual connections to encode geometric and semantic information, while the feature aggregation block (FAB) upsamples low-resolution features and fuses them with skip connections to recover fine spatial details.}
  \label{fig:3DUNet_blocks}
\end{figure}

The refinement module receives the coarse discrete volume $\hat{\mathbf{Y}}$ and must
(i) map it into a continuous feature space,
(ii) aggregate multi-scale contextual information, and
(iii) reconstruct a refined voxel grid $\hat{\mathbf{Y}}^{\prime}$.
To accomplish these steps, we adopt a three-dimensional U-shaped neural network (3D U-Net) backbone~\cite{2016_3DUNET,2015_2DUNET}, whose overall encoder–bottleneck–decoder structure is illustrated in Figure~\ref{fig:3DUNet_arch}.
The specific computational blocks that constitute the encoder and decoder, namely the feature encoding block (FEB) and the feature aggregation block (FAB), are detailed in Figure~\ref{fig:3DUNet_blocks}.

\subsubsection{Voxel Embedding and Encoder–Decoder}\label{sec:3dunet-main}

We first embed the discrete labels of $\hat{\mathbf{Y}}$ into a continuous feature map:
\begin{equation}\label{eq:Embedding}
  F_{\mathrm{emb}} = \mathrm{Embedding}(\hat{\mathbf{Y}}),
\end{equation}
where $F_{\mathrm{emb}} \in \mathbb{R}^{C \times H \times W \times Z}$. 
A $1 \times 1 \times 1$ 3D convolution then produces the input feature:
\begin{equation}\label{eq:input}
  F_{\mathrm{in}} = \mathrm{Conv}_{\mathrm{in}}(F_{\mathrm{emb}}).
\end{equation}

The encoder uses four stacked feature encoding blocks (FEBs) (Figure~\ref{fig:3DUNet_blocks}) to extract multi-scale features
$F^{1:s}$ at progressively lower resolutions. 
For a voxel grid of size $H \times W \times Z$ and feature dimension $G$, the encoder outputs
\begin{equation}
  \label{eq:3DUNet_encoder}
  F^{1:s} \in \mathbb{R}^{G \times \tfrac{H}{s} \times \tfrac{W}{s} \times \tfrac{Z}{s}},
  \quad s \in \{1, 2, 4, 8, 16\}.
\end{equation}
A bottleneck processes $F^{1:16}$, and the decoder then upsamples via four stacked feature aggregation blocks (FABs), followed by a shared prediction head that produces voxel logits at multiple scales:
\begin{equation}
  \label{eq:3DUNet_decoder}
  \hat{Y}_{1:s}^{\prime} \in \mathbb{R}^{C \times \tfrac{H}{s} \times \tfrac{W}{s} \times \tfrac{Z}{s}},
  \quad s \in \{1, 2, 4, 8\},
\end{equation}
where $C$ is the number of semantic classes.
At inference time, we use
\begin{equation}
  \label{eq:Pred_2}
  \hat{\mathbf{Y}}^{\prime} = \arg\max_{c} \hat{Y}_{1:1,c}^{\prime}
\end{equation}
as the final refined semantic voxel prediction.

\subsubsection{Feature Encoding Block (FEB)}\label{sec:FEB}

Each FEB refines features at a given scale and produces both a skip feature and a downsampled feature. 
As in Figure~\ref{fig:3DUNet_blocks}, an FEB applies two 3D convolutions with InstanceNorm3D~\cite{2016_InstanceNorm3D} and LeakyReLU~\cite{2015_LeakyReLU}, followed by a residual skip and a stride-2 convolution:
\begin{equation}
  \label{eq:FEB_CONV}
  \begin{cases}
    F_{\mathrm{skip}}^{1:\ell} = \mathrm{ConvBlock}(F_{\mathrm{in}}^{1:\ell}) + F_{\mathrm{in}}^{1:\ell},\\
    F^{1:2\ell}               = \mathrm{Conv}_{2 \times 2 \times 2}(F_{\mathrm{skip}}^{1:\ell}),\\
    F_{\mathrm{out}}^{1:2\ell} = \mathrm{ConvBlock}(F^{1:2\ell}),
  \end{cases}
  \quad \forall\,\ell \in \{1,2,4,8\}.
\end{equation}

\subsubsection{Feature Aggregation Block (FAB) and Multi-scale Supervision}\label{sec:FAB}

Each FAB upsamples low-resolution features and fuses them with encoder skip features:
\begin{equation}
  \label{eq:FAB_CONV}
  \begin{cases}
    F_{\mathrm{up}}^{1:2\ell} = \mathrm{UpSample}(F_{\mathrm{in}}^{1:2\ell}),\\
    F_{\mathrm{out}}^{1:\ell} = \mathrm{ResConvBlock}(F_{\mathrm{skip}}^{1:\ell}, F_{\mathrm{up}}^{1:2\ell}),
  \end{cases}
  \quad \forall\,\ell \in \{1,2,4,8\}.
\end{equation}

Following PaSCo~\cite{2024_PaSCo}, each decoder feature map $F_{\mathrm{out}}^{1:\ell}$ is mapped to logits by a $1 \times 1 \times 1$ 3D convolution:
\begin{equation}\label{eq:Pred}
  \hat{Y}_{1:\ell}^{\prime} = \mathrm{PredHead}(F_{\mathrm{out}}^{1:\ell}),
  \quad \forall\,\ell \in \{1,2,4,8\},
\end{equation}
and all scales are supervised during training.
This encourages coarse-to-fine refinement and stabilises optimisation.

\subsection{Progressive Neighborhood Attention Module (PNAM)}\label{sec:PNAM}

\begin{figure}[t]
  \centering
  \includegraphics[width=\textwidth]{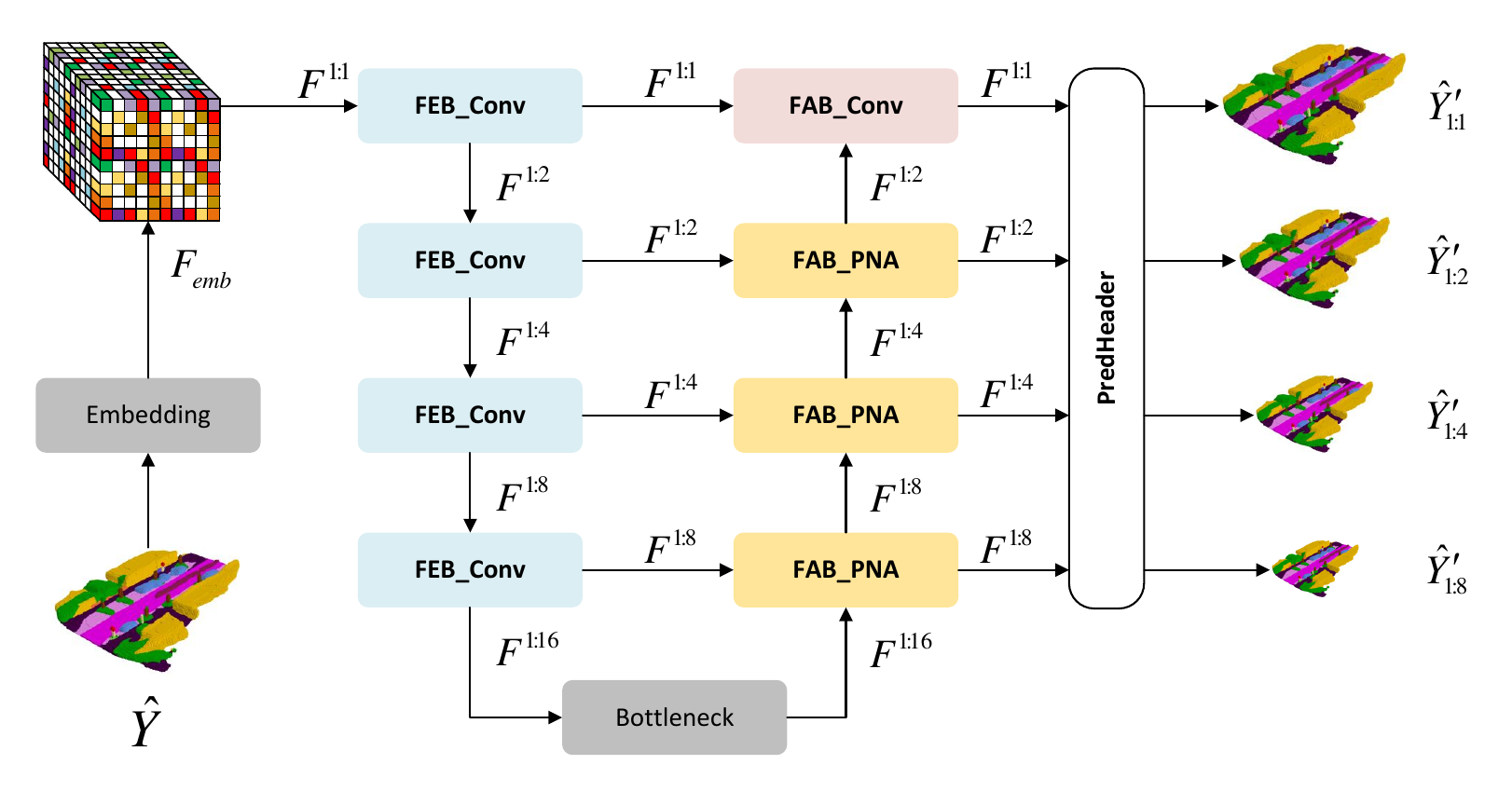}
  \caption{\textbf{Progressive Neighborhood Attention U-Net (PNA U-Net)~\cite{2023_PNA}.}
  PNAM replaces the feature aggregation blocks (FABs) at intermediate scales ($1{:}2$, $1{:}4$, $1{:}8$) with attention-based FABs, while the finest-scale FAB remains convolutional for efficiency.}
  \label{fig:PNA_UNet_arch}
\end{figure}

Purely convolutional decoders aggregate context only within fixed local windows, limiting their ability to capture long-range and structure-aware voxel relations. To address this, we integrate the Progressive Neighborhood Attention Module (PNAM)~\cite{2023_PNA} into the decoder of our refinement network.

As illustrated in Figure~\ref{fig:PNA_UNet_arch}, the FABs at scales $1{:}2$, $1{:}4$, and $1{:}8$ are replaced with PNA-based FABs, while the finest-scale FAB remains convolutional for efficiency. PNAM enhances multi-scale voxel reasoning by combining global self-attention~\cite{2017_AIAYN} with localized neighborhood aggregation~\cite{hassani2022dilated,hassani2023neighborhood,hassani2024faster}.

\begin{figure}[t]
  \centering
  \includegraphics[width=\textwidth]{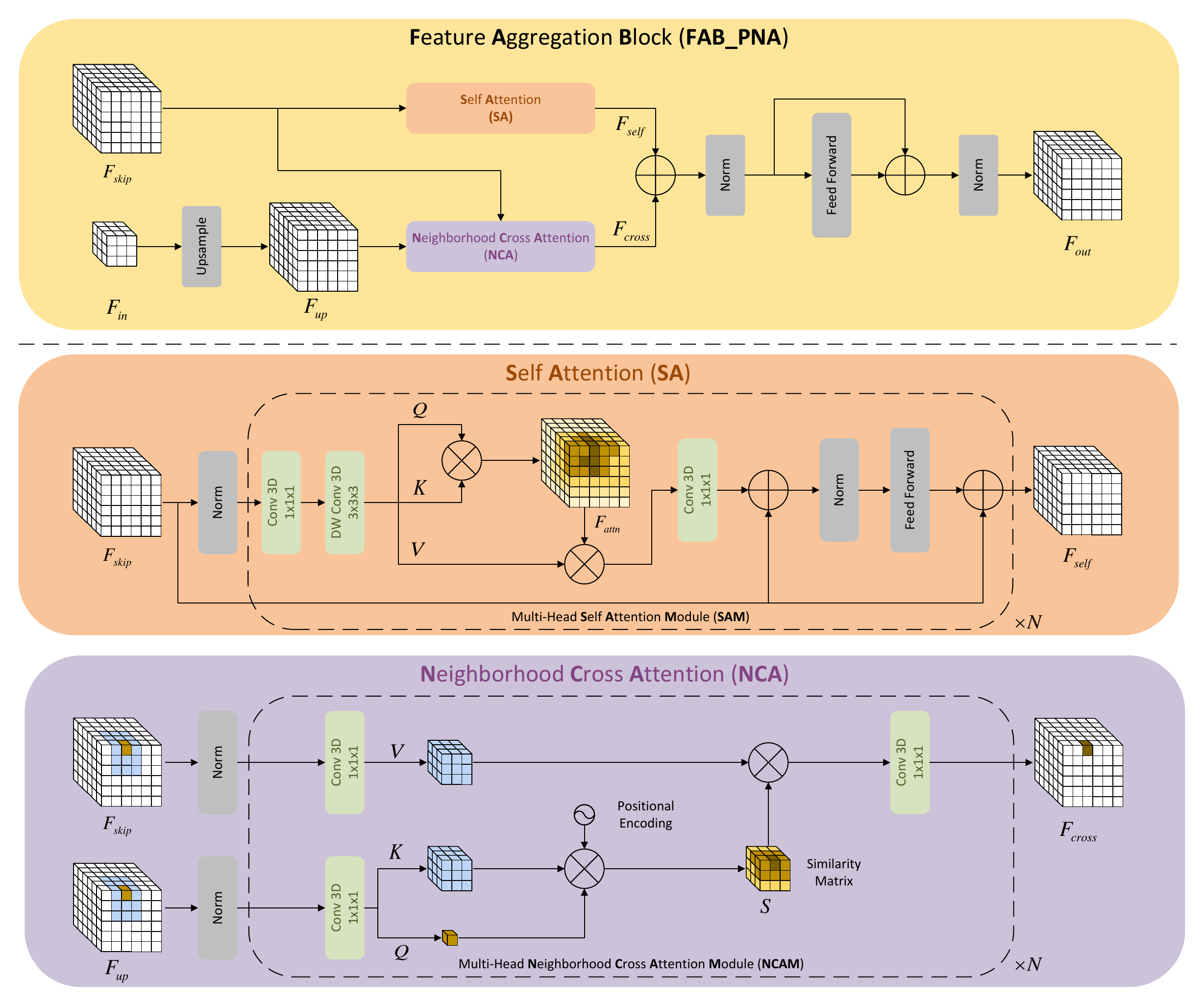}
  \caption{\textbf{Core components of the PNA-based Feature Aggregation Block (FAB).}
  Each block combines self-attention (SA) and neighborhood cross-attention (NCA) to jointly model long-range dependencies and local geometric consistency.}
  \label{fig:PNA_UNet_blocks}
\end{figure}

\subsubsection{PNA-based Feature Aggregation Block}

As illustrated in Figure~\ref{fig:PNA_UNet_blocks}, a PNA-based FAB consists of two branches:  
(1) a self-attention (SA) branch operating on $F_{\mathrm{up}}$, and  
(2) a neighborhood cross-attention (NCA) branch operating between $F_{\mathrm{skip}}$ and $F_{\mathrm{up}}$.

Given the upsampled feature $F_{\mathrm{up}}^{1:\ell}$ and the corresponding skip feature $F_{\mathrm{skip}}^{1:\ell}$, the two attention responses are computed as:
\[
F_{\mathrm{self}}^{1:\ell}=\mathrm{SA}(F_{\mathrm{up}}^{1:\ell}), \qquad
F_{\mathrm{cross}}^{1:\ell}=\mathrm{NCA}(F_{\mathrm{skip}}^{1:\ell},F_{\mathrm{up}}^{1:\ell}),
\]
for $\ell\in\{2,4,8\}$.  
The outputs are fused and refined via normalization and a lightweight feed-forward network:
\[
F_{\mathrm{out}}^{1:\ell}=\mathrm{FFN}\big(\mathrm{Norm}(F_{\mathrm{self}}^{1:\ell}+F_{\mathrm{cross}}^{1:\ell})\big).
\]

\subsubsection{Self-Attention (SA)}

SA refines the upsampled voxel features by capturing long-range dependencies.  
Following the standard multi-head attention formulation~\cite{2017_AIAYN}, we use $1^3$ and depthwise $3^3$ convolutions to compute $Q,K,V$, followed by attention and a residual FFN.  
This propagates global geometric–semantic cues, compensating for missing structures in the coarse prediction.

\subsubsection{Neighborhood Cross-Attention (NCA)}

NCA enforces local geometric consistency.  
Inspired by the NATTEN family of neighborhood attention operators~\cite{hassani2022dilated,hassani2023neighborhood,hassani2024faster}, it restricts attention to a 3D neighborhood window, enabling each voxel to aggregate high-confidence structural cues from spatially adjacent voxels.  
This makes PNAM particularly effective at restoring fine structures such as object boundaries and thin geometry.

Overall, PNAM strengthens the refinement network’s ability to jointly model global context and local voxel continuity across scales.

\subsection{Vision–Language Guidance Module (VLGM)}\label{sec:VLGM}

\begin{figure}[t]
  \centering
  \includegraphics[width=\textwidth]{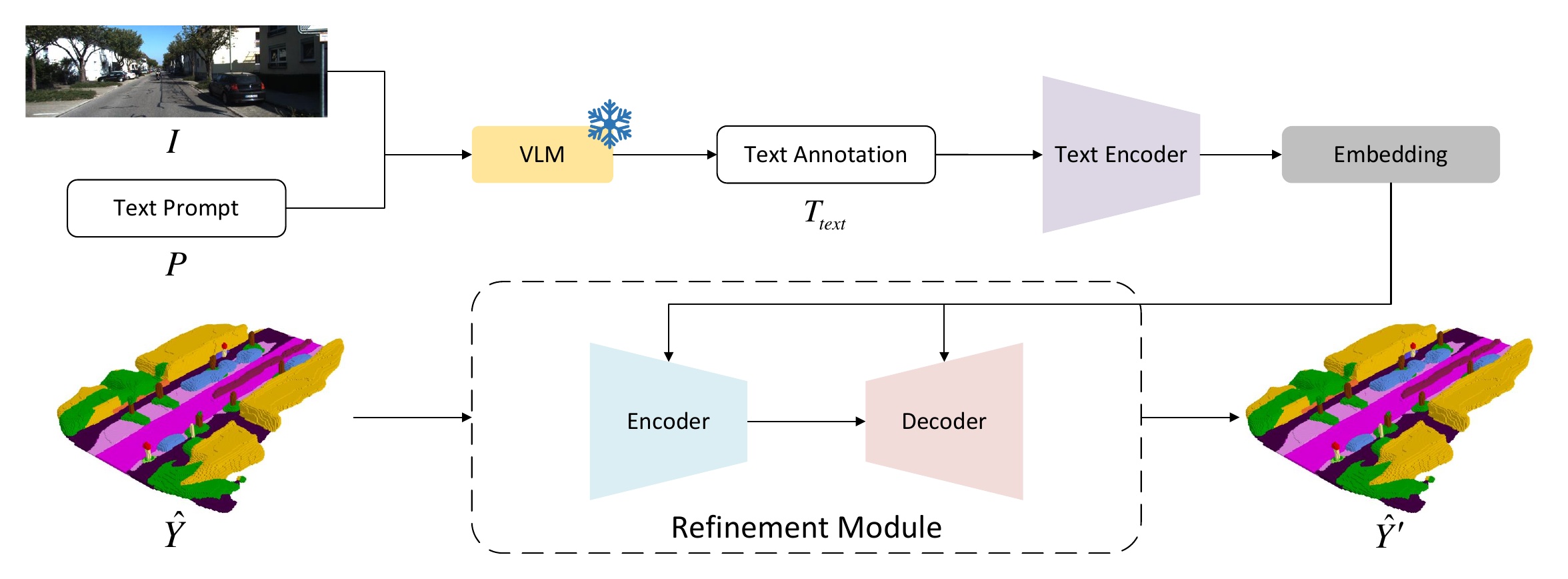}
  \caption{\textbf{VLGM pipeline}. A frozen VLM produces a free-form scene description, which is encoded and fused with voxel features.}
  \label{fig:VLGM}
\end{figure}

Even with stronger voxel–voxel reasoning, SSC remains ambiguous in occluded or sparsely observed regions. To inject high-level scene priors—such as road layout, object co-occurrence patterns, or typical urban structures—we introduce the Vision–Language Guidance Module (VLGM). As illustrated in Figure~\ref{fig:VLGM}, the module leverages a frozen vision–language model (VLM) to produce a free-form scene description, whose textual semantics are encoded and fused into the voxel refinement pipeline.

\subsubsection{Text Acquisition and Semantic Encoding}

Given an input image $I$ and prompt $P$, a frozen VLM such as LLaVA~\cite{2023_LLaVA,2024_LLaVA_2} or InstructBLIP~\cite{2023_InstructBLIP} generates a scene description
\[
T_{\mathrm{text}}=\mathrm{VLM}(I,P),
\]
which is precomputed offline to avoid training overhead.  

To capture different levels of textual semantics, we employ two complementary encoders.  
(1) \textbf{JinaCLIP}~\cite{2024_JinaCLIP,2024_JinaCLIP_2} extracts a global embedding  
\[
F_{\mathrm{text}}^{\mathrm{JinaCLIP}}=\mathrm{JinaCLIP}(T_{\mathrm{text}}),
\]
providing holistic scene cues.  
(2) A \textbf{Q-Former}~\cite{2023_BLIP2} produces token-level embeddings  
\[
F_{\mathrm{text}}^{\mathrm{Q\text{-}Former}}=\mathrm{Q\text{-}Former}(T_{\mathrm{text}}),
\]
which enable fine-grained cross-modal alignment.  
This design follows instruction-style prompting practices used in~\cite{2023_InstructBLIP,2025_VLME2E}.

\subsubsection{Text–Voxel Fusion Modules}

To integrate text cues into voxel refinement, we build a Text U-Net by inserting lightweight fusion blocks after each FEB and FAB. Each fusion block consists of two components:

\paragraph{Semantic Interaction Guidance Module (SIGM).}
Following Text-IF~\cite{2024_Text_IF}, global JinaCLIP features are mapped to affine parameters $(\gamma_m,\beta_m)$ via MLPs. Voxel features are modulated as  
\[
F_{\mathrm{out}} = (1+\gamma_m)\odot F_{\mathrm{in}} + \beta_m,
\]
injecting scene-level priors that guide early geometric reasoning.

\paragraph{Dual Cross-Attention Module (DCAM).}
Inspired by BLIP-2~\cite{2023_BLIP2}, SAM~\cite{2023_SAM}, and MultiRAtt-RSSC~\cite{2024_MultiAtt_RSSC}, DCAM alternates self- and cross-attention between Q-Former tokens and voxel features. Text self-attention yields $F_{\mathrm{text}}^{\mathrm{attn}}$, followed by text-to-voxel cross-attention producing $F_{\mathrm{text}}^{\mathrm{enhanced}}$, and voxel-to-text cross-attention generating $F_{\mathrm{voxel}}^{\mathrm{enhanced}}$. A residual update produces  
\[
F_{\mathrm{out}}=\mathrm{LayerNorm}(F_{\mathrm{voxel}}^{\mathrm{enhanced}}+F_{\mathrm{in}}).
\]

SIGM injects global scene priors (e.g., “urban street with parked vehicles”), while DCAM provides fine-grained token-level alignment. As visualized in Figure~\ref{fig:VLGM}, the two components operate synergistically to improve geometric completeness and semantic coherence, especially in occluded and ambiguous regions.

\subsection{Loss Function}\label{sec:loss}

ESSC-RM performs coarse-to-fine refinement across multiple spatial scales.  
We therefore supervise both voxel-wise predictions and scene-level consistency using two complementary terms: a class-weighted cross-entropy loss and the scene--class affinity loss (SCAL)~\cite{2021_MonoScene,2023_CGFormer}.  
This combination stabilises multi-scale refinement while encouraging globally coherent semantics.

% -------------------------
% CE LOSS
% -------------------------
\subsubsection{Cross-Entropy Loss}\label{sec:CE}

At each refinement scale $l$, voxel predictions are supervised using a class-weighted cross-entropy:
\begin{equation}\label{eq:CE}
\mathcal{L}_{l} = 
-\frac{1}{C}
\sum_{i=0}^{N}\sum_{c=0}^{C}
w_{c}\, y_{i,c}\,
\log\frac{\exp(\hat{y}^{\prime}_{i,c})}
{\sum_{c^{\prime}=0}^{C}\exp(\hat{y}^{\prime}_{i,c^{\prime}})},
\end{equation}
where $\hat{y}^{\prime}$ denotes refinement logits and $w_{c}$ compensates for class imbalance~\cite{2020_LMSCNet}.  
Aggregating all scales yields:
\begin{equation}\label{eq:CE_Loss}
\mathcal{L}_{ce} = \sum_{l=0}^{L}\mathcal{L}_{l}.
\end{equation}

% -------------------------
% SCAL
% -------------------------
\subsubsection{Scene--Class Affinity Loss (SCAL)}\label{sec:SCAL}

To promote globally consistent refinement—particularly under sparsity or ambiguous projections—we adopt SCAL~\cite{2021_MonoScene}, which optimises class-wise precision ($P_{c}$), recall ($R_{c}$), and specificity ($S_{c}$).  
Let $p_{i}$ denote the ground-truth class for voxel $i$, and $\hat{p}_{i,c}$ the predicted probability for class $c$.  
Using Iverson brackets $\llbracket \cdot \rrbracket$, the metrics are:
\begin{align}\label{eq:SCA}
P_{c}(\hat{p},p) &=
\log\frac{\sum_{i}\hat{p}_{i,c}\llbracket p_{i}=c\rrbracket}
{\sum_{i}\hat{p}_{i,c}}, \\[0.4em]
R_{c}(\hat{p},p) &=
\log\frac{\sum_{i}\hat{p}_{i,c}\llbracket p_{i}=c\rrbracket}
{\sum_{i}\llbracket p_{i}=c\rrbracket}, \\[0.4em]
S_{c}(\hat{p},p) &=
\log\frac{\sum_{i}(1-\hat{p}_{i,c})(1-\llbracket p_{i}=c\rrbracket)}
{\sum_{i}(1-\llbracket p_{i}=c\rrbracket)}.
\end{align}

The per-scale affinity loss is:
\begin{equation}\label{eq:SCA_level_Loss}
\mathcal{L}_{l}(\hat{p},p) =
-\frac{1}{C}
\sum_{c=1}^{C}\big(
P_{c}(\hat{p},p)
+ R_{c}(\hat{p},p)
+ S_{c}(\hat{p},p)
\big).
\end{equation}

SCAL is applied to both semantic and geometric predictions across all refinement scales:
\begin{align}\label{eq:SCA_Loss}
\mathcal{L}_{scal}^{sem} &= \sum_{l=0}^{L}\mathcal{L}_{l}(\hat{y}^{\prime},y), \\[-0.4em]
\mathcal{L}_{scal}^{geo} &= \sum_{l=0}^{L}\mathcal{L}_{l}(\hat{y}^{\prime,geo},y^{geo}).
\end{align}

% -------------------------
% FINAL LOSS
% -------------------------
\subsubsection{Overall Objective}\label{sec:overall_loss}

The total training loss is:
\begin{equation}\label{eq:Loss}
\mathcal{L} =
\lambda_{ce}\mathcal{L}_{ce}
+ \lambda_{scal}^{geo}\mathcal{L}_{scal}^{geo}
+ \lambda_{scal}^{sem}\mathcal{L}_{scal}^{sem},
\end{equation}
with all coefficients set to $1$ in our experiments, providing a balanced supervision over voxel-wise accuracy, geometric completion, and scene-level semantic consistency.

\section{Experiment}\label{sec:experiment}%
This section evaluates ESSC-RM on the SemanticKITTI benchmark~\cite{2019_SemanticKITTI,2021_SemanticKITTI_2}.
We first describe the experimental setup (datasets, metrics, and implementation), then report quantitative and qualitative results on strong and weak semantic scene completion baselines (CGFormer and MonoScene).
Comprehensive ablation studies that analyze the refinement framework, the neighborhood-attention-based aggregation module, and the vision–language guidance module are provided in the supplementary material.

\subsection{Experimental Setup}

\subsubsection{Datasets}\label{sec:datasets}%
We adopt the SemanticKITTI semantic scene completion benchmark~\cite{2019_SemanticKITTI,2021_SemanticKITTI_2}, which extends the KITTI odometry dataset~\cite{2012_KITTI} with dense semantic labels for each LiDAR scan.
The dataset contains $22$ outdoor sequences; following the official split, sequences 00--07 and 09--10 are used for training, 08 for validation, and 11--21 as a hidden test set.

For semantic scene completion, a 3D volume around the ego-vehicle is considered: $51.2\,\mathrm{m}$ in front, $25.6\,\mathrm{m}$ to each side (total width $51.2\,\mathrm{m}$), and $6.4\,\mathrm{m}$ in height~\cite{2019_SemanticKITTI}.
This volume is voxelized into a $256 \times 256 \times 32$ grid with voxel size $0.2\,\mathrm{m}^3$.
Each voxel is assigned one of $20$ classes (19 semantic classes and 1 free-space), obtained by voxelizing aggregated, registered semantic point clouds~\cite{2023_VoxFormer}.

We conduct all experiments on SemanticKITTI, following its established voxelization protocol and official evaluation scripts, which provides a standardized testbed for semantic scene completion.

\subsubsection{Evaluation Metrics}\label{sec:metrics}%
We follow standard practice~\cite{2021_MonoScene,2023_VoxFormer,2023_CGFormer} and report intersection-over-union (IoU) for 3D scene completion (SC) and mean intersection-over-union (mIoU) for semantic scene completion (SSC).

For SC, evaluation is binary (occupied vs.\ free) and uses IoU over the occupancy grid:
\begin{equation}\label{eq:IoU}
    \text{IoU} = \frac{\text{completion\_TP}}{\text{completion\_TP} + \text{completion\_FP} + \text{completion\_FN}},
\end{equation}
where TP, FP, and FN denote true positives, false positives, and false negatives on the occupancy grid.

For SSC, we evaluate per-class IoU over $C=19$ semantic classes and report mean IoU:
\begin{equation}\label{eq:mIoU}
    \text{mIoU} = \frac{1}{C} \sum_{c=1}^{C} \frac{\text{TP}_{c}}{\text{TP}_{c} + \text{FP}_{c} + \text{FN}_{c} + \epsilon},
\end{equation}
where $\text{TP}_{c}$, $\text{FP}_{c}$ and $\text{FN}_{c}$ are computed for class $c$, and evaluation is carried out in known space as in~\cite{2021_3DSSC}.
IoU primarily reflects geometric completion quality, whereas mIoU captures voxel-wise semantic accuracy; both are reported to assess overall scene understanding.

\subsubsection{Implementation Details}\label{sec:impl}%
We consider two training paradigms for ESSC-RM: 
(1) joint training, where the semantic scene completion backbone is switched to inference mode while the refinement module is trained on-the-fly from its predictions; and 
(2) separate training, where semantic scene completion predictions are pre-computed and stored, and the refinement module is trained purely as a plug-and-play post-processor without modifying the original semantic scene completion architecture.

Unless otherwise stated, experiments are conducted on two NVIDIA RTX A5000 GPUs, with 10 epochs and a batch size of $1$ per GPU.
We use AdamW~\cite{2017_AdamW} with $\beta_{1}=0.9$, $\beta_{2}=0.99$, and a peak learning rate of $5 \times 10^{-5}$.
A cosine schedule~\cite{2017_OneCycleLR} with $5\%$ warm-up is applied.
The refinement module follows a 3D U-Net~\cite{2016_3DUNET} backbone; encoder and decoder feature-enhancement blocks (FEB/FAB) are adapted from SemCity~\cite{2024_SemCity}, and neighborhood-attention-based variants from NATTEN~\cite{hassani2022dilated,hassani2023neighborhood,hassani2024faster} and PNA~\cite{2023_PNA}.
The vision–language guidance module (VLGM) uses frozen vision–language models (InstructBLIP~\cite{2023_BLIP2,2023_InstructBLIP} and LLaVA~\cite{2023_LLaVA,2024_LLaVA_2}) together with text–voxel fusion modules inspired by Text-IF~\cite{2024_Text_IF} and MultiAtt-RSSC~\cite{2024_MultiAtt_RSSC}.
Following PaSCo~\cite{2024_PaSCo} and HybridOcc~\cite{2024_HybridOcc}, we apply coarse-to-fine multi-level supervision in the decoder.
Training losses are described in Sec.~\ref{sec:loss}.

\subsection{Evaluation Results}\label{sec:eval_results}
We evaluate ESSC-RM as a refinement module on strong and weak SSC baselines and analyze its efficiency and qualitative behavior.

\subsubsection{Quantitative Results}\label{sec:quant_results}
ESSC-RM is designed to prioritize voxel-wise semantic
correctness (mIoU) over boundary-sensitive binary occupancy
smoothness (IoU); therefore, minor IoU drops may accompany
consistent mIoU gains.

\paragraph{3D SSC performance.}
Table~\ref{tab:QuantitativeResults} reports SSC performance on SemanticKITTI, including representative image-based SSC baselines and our ESSC-RM variants.
Among the listed baselines without ESSC-RM (upper block), DepthSSC~\cite{2024_DepthSSC} achieves the best SC-IoU (45.84\%), while Symphonize~\cite{2023_Symphonies} attains the highest mIoU (14.89\%).
In addition to these method-level comparisons, we evaluate ESSC-RM as a plug-and-play refinement module on top of two representative SSC backbones:
CGFormer~\cite{2023_CGFormer} as a strong baseline (45.99\% IoU, 16.87\% mIoU) and MonoScene~\cite{2021_MonoScene} as a widely used weaker baseline (36.86\% IoU, 11.08\% mIoU).
Due to training and storage overhead of voxel-level refinement, we instantiate ESSC-RM on these two backbones to demonstrate generality across different performance regimes; extending the plug-in evaluation to additional backbones is left for future work (see Sec.~\ref{sec:conclusion}).

\begin{table}[htbp]
    \centering
    \resizebox{\textwidth}{!}{%
        \begin{tabular}{@{}p{4cm}|cc|lllllllllllllllllll@{}}
        \toprule
        Methods & IoU & mIoU & 
        \rotatebox{90}{{\color{car}\rule{1ex}{1ex}}\, car (3.92\%)} & 
        \rotatebox{90}{{\color{bicycle}\rule{1ex}{1ex}}\, bicycle (0.03\%)} & 
        \rotatebox{90}{{\color{motorcycle}\rule{1ex}{1ex}}\, motorcycle (0.03\%)} & 
        \rotatebox{90}{{\color{truck}\rule{1ex}{1ex}}\, truck (0.16\%)} & 
        \rotatebox{90}{{\color{othervehicle}\rule{1ex}{1ex}}\, other-vehicle (0.20\%)} & 
        \rotatebox{90}{{\color{person}\rule{1ex}{1ex}}\, person (0.07\%)} & 
        \rotatebox{90}{{\color{bicyclist}\rule{1ex}{1ex}}\, bicyclist (0.07\%)} & 
        \rotatebox{90}{{\color{motorcyclist}\rule{1ex}{1ex}}\, motorcyclist (0.05\%)} & 
        \rotatebox{90}{{\color{road}\rule{1ex}{1ex}}\, road (15.30\%)} & 
        \rotatebox{90}{{\color{parking}\rule{1ex}{1ex}}\, parking (1.12\%)} & 
        \rotatebox{90}{{\color{sidewalk}\rule{1ex}{1ex}}\, sidewalk (11.13\%)} & 
        \rotatebox{90}{{\color{otherground}\rule{1ex}{1ex}}\, other-ground (0.56\%)} & 
        \rotatebox{90}{{\color{building}\rule{1ex}{1ex}}\, building (14.10\%)} & 
        \rotatebox{90}{{\color{fence}\rule{1ex}{1ex}}\, fence (3.90\%)} & 
        \rotatebox{90}{{\color{vegetation}\rule{1ex}{1ex}}\, vegetation (39.3\%)} & 
        \rotatebox{90}{{\color{trunk}\rule{1ex}{1ex}}\, trunk (0.51\%)} & 
        \rotatebox{90}{{\color{terrain}\rule{1ex}{1ex}}\, terrain (9.17\%)} & 
        \rotatebox{90}{{\color{pole}\rule{1ex}{1ex}}\, pole (0.29\%)} & 
        \rotatebox{90}{{\color{trafficsign}\rule{1ex}{1ex}}\, traffic-sign (0.08\%)} \\ \midrule

        \multicolumn{22}{l}{\emph{Baselines (without ESSC-RM)}} \\ \midrule
        TPVFormer \cite{2023_TPVFormer} 
        & 35.61 & 11.36 & 23.81 & 0.36 & 0.05 & 8.08 & 4.35 & 0.51 & 0.89 & 0.00 & 56.50 & \textbf{20.60} & 25.87 & 0.85 & 13.88 & 5.94 & 16.92 & 2.26 & 30.38 & 3.14 & 1.52 \\ [3pt]
        OccFormer \cite{2023_OccFormer} 
        & 36.50 & 13.46 & 25.09 & 0.81 & \underline{1.19} & \textbf{25.53} & \underline{8.52} & 2.78 & 2.82 & 0.00 & \textbf{58.85} & \underline{19.61} & 26.88 & 0.31 & 14.40 & 5.61 & 19.63 & 3.93 & 32.62 & 4.26 & 2.86 \\ [3pt]
        IAMSSC \cite{2024_IAMSSC} 
        & 44.29 & 12.45 & 26.26 & 0.60 & 0.15 & 8.74 & 5.06 & 1.32 & 3.46 & \textbf{0.01} & 54.55 & 16.02 & 25.85 & 0.70 & 17.38 & 6.86 & 24.63 & 4.95 & 30.13 & 6.35 & 3.56 \\ [3pt]
        VoxFormer-S \cite{2023_VoxFormer} 
        & 44.02 & 12.35 & 25.79 & 0.59 & 0.51 & 5.63 & 3.77 & 1.78 & 3.32 & 0.00 & 54.76 & 15.50 & 26.35 & 0.70 & 17.65 & 7.64 & 24.39 & 5.08 & 29.96 & 7.11 & 4.18 \\ [3pt]
        DepthSSC \cite{2024_DepthSSC} 
        & \textbf{45.84} & 13.28 & 25.94 & 0.35 & 1.16 & 6.02 & 7.50 & 2.58 & \textbf{6.32} & 0.00 & 55.38 & 18.76 & 27.04 & 0.92 & 19.23 & \textbf{8.46} & \textbf{26.37} & 4.52 & 30.19 & 7.42 & 4.09 \\ [3pt]
        Symphonize \cite{2023_Symphonies} 
        & 41.92 & \textbf{14.89} & \textbf{28.68} & \textbf{2.54} & \textbf{2.82} & \underline{20.44} & \textbf{13.89} & \textbf{3.52} & 2.24 & 0.00 & 56.37 & 15.28 & 27.58 & \underline{0.95} & \textbf{21.64} & \underline{8.40} & 25.72 & \underline{6.60} & 30.87 & \textbf{9.57} & \textbf{5.76} \\ [3pt]
        HASSC-S \cite{2024_HASSC} 
        & \underline{44.82} & 13.48 & 27.23 & \underline{0.92} & 0.86 & 9.91 & 5.61 & \underline{2.80} & \underline{4.71} & 0.00 & \underline{57.05} & 15.90 & \underline{28.25} & \textbf{1.04} & 19.05 & 6.58 & 25.48 & 6.15 & \underline{32.94} & 7.68 & 4.05 \\ [3pt]
        H2GFormer-S \cite{2024_H2GFormer} 
        & 44.57 & \underline{13.73} & \underline{28.21} & 0.50 & 0.47 & 10.00 & 7.39 & 1.54 & 2.88 & \underline{0.00} & 56.08 & 17.83 & \textbf{29.12} & 0.45 & \underline{19.74} & 7.24 & \underline{26.25} & \textbf{6.80} & \textbf{34.42} & \underline{7.88} & \underline{4.68} \\ [3pt] \midrule

        \multicolumn{22}{l}{\emph{MonoScene and ESSC-RM variants}} \\ \midrule
        MonoScene \cite{2021_MonoScene} 
        & \textbf{36.86} & 11.08 & \underline{23.26} & \textbf{0.61} & 0.45 & 6.98 & 1.48 & 1.86 & 1.20 & 0.00 & \textbf{56.52} & 14.27 & \textbf{26.72} & 0.46 & \textbf{14.09} & 5.84 & \textbf{17.89} & 2.81 & \textbf{29.64} & \textbf{4.14} & 2.25 \\ [3pt]
        MonoScene + 3D U-Net 
        & 35.70 & 11.47 & \textbf{23.46} & 0.41 & \textbf{0.87} & 10.95 & \textbf{3.69} & \underline{2.98} & \underline{1.64} & 0.00 & 56.24 & \textbf{14.95} & 26.63 & \underline{1.42} & 13.11 & 6.19 & 16.75 & 2.73 & \underline{29.57} & 3.77 & 2.62 \\ [3pt]
        MonoScene + VLGM 
        & 35.62 & \underline{11.49} & 22.76 & \underline{0.44} & 0.71 & \textbf{12.45} & 3.12 & \textbf{3.04} & \underline{1.64} & 0.00 & \underline{56.48} & 14.35 & 26.64 & \underline{1.42} & \underline{13.55} & \textbf{6.28} & 16.44 & \textbf{2.97} & 29.50 & \underline{3.85} & \underline{2.65} \\ [3pt]
        MonoScene + PNAM 
        & \underline{36.44} & \textbf{11.51} & 23.11 & 0.40 & \underline{0.73} & \underline{11.38} & \underline{3.59} & 2.95 & \textbf{1.69} & 0.00 & 56.27 & \underline{14.65} & \underline{26.71} & \textbf{1.45} & 13.48 & \underline{6.20} & \underline{17.08} & \underline{2.96} & 29.45 & 3.84 & \textbf{2.69} \\ [3pt] \midrule

        \multicolumn{22}{l}{\emph{CGFormer and ESSC-RM variants}} \\ \midrule
        CGFormer \cite{2023_CGFormer} 
        & \textbf{45.99} & 16.87 & \underline{34.32} & 4.61 & 2.71 & 19.44 & 7.67 & 2.38 & 4.08 & 0.00 & \underline{65.51} & \underline{20.82} & \textbf{32.31} & \textbf{0.16} & \textbf{23.52} & \textbf{9.20} & \textbf{26.93} & 8.83 & \textbf{39.54} & 10.67 & \textbf{7.84} \\ [3pt]
        CGFormer + 3D U-Net 
        & 43.53 & 17.17 & 33.99 & \underline{5.28} & \textbf{3.11} & \underline{22.39} & \underline{8.22} & \underline{2.65} & 4.05 & 0.00 & 65.29 & 20.26 & \underline{32.14} & \underline{0.13} & 23.11 & 8.93 & \underline{26.84} & \underline{11.17} & 38.99 & \textbf{11.93} & \textbf{7.84} \\ [3pt]
        CGFormer + VLGM 
        & 43.20 & \underline{17.21} & \textbf{34.33} & 5.24 & \underline{3.01} & 22.33 & 7.81 & \textbf{2.70} & \underline{4.12} & 0.00 & \textbf{65.52} & 20.79 & \textbf{32.31} & \underline{0.13} & \underline{23.27} & 8.95 & 26.69 & 10.73 & \underline{39.29} & \textbf{11.93} & \underline{7.82} \\ [3pt]
        CGFormer + PNAM 
        & \underline{44.33} & \textbf{17.27} & 34.11 & \textbf{5.69} & 2.94 & \textbf{23.71} & \textbf{8.36} & 2.64 & \textbf{4.37} & 0.00 & 65.27 & \textbf{20.87} & 31.90 & \textbf{0.16} & 22.70 & \underline{9.08} & 26.63 & \textbf{11.42} & 38.91 & \underline{11.78} & 7.66 \\ \bottomrule
        \end{tabular}
    }
    \caption[Quantitative results on the SemanticKITTI validation set]{Quantitative results on the SemanticKITTI validation set. The upper block lists baseline stereo-based SSC methods without ESSC-RM. The middle and lower blocks show MonoScene- and CGFormer-based ESSC-RM variants, respectively. Within each block, the best and second-best results are shown in \textbf{bold} and \underline{underlined}, respectively.}
    \label{tab:QuantitativeResults}%
\end{table}

To assess the generality of ESSC-RM, we plug it on top of both CGFormer and MonoScene, progressively adding
(i) a plain 3D U-Net refinement head,
(ii) the proposed neighborhood-attention-based refinement module (PNAM),
and (iii) the vision--language guidance module (VLGM).
The MonoScene and CGFormer blocks in Table~\ref{tab:QuantitativeResults} summarize these ablation results.

\paragraph{ESSC-RM on CGFormer.}
As shown in the CGFormer block of Table~\ref{tab:QuantitativeResults}, adding a 3D U-Net refinement head improves mIoU from $16.87\%$ to $17.17\%$.
Equipping the refinement with VLGM further increases mIoU to $17.21\%$, while PNAM achieves the best mIoU of $17.27\%$ with only a modest IoU drop.
The gains are more apparent on small and medium-scale categories (e.g., truck, bicycle, trunk, pole), suggesting that coarse-to-fine decoding and neighborhood-aware aggregation help correct local ambiguities and recover thin structures that are challenging for the backbone alone.

Despite consistent improvements, the absolute mIoU gain on CGFormer remains moderate (from $16.87\%$ to $17.27\%$, +0.40, i.e., $\sim$2\% relative).
This is mainly because ESSC-RM performs refinement \emph{in the voxel-prediction space} by design: it takes the discrete semantic occupancy $\hat{\mathbf{Y}}$ predicted by the backbone, embeds it into a continuous feature map, and refines it via a 3D U-Net style encoder--decoder (with PNAM/VLGM as optional enhancements).
Consequently, the global occupancy layout and object extents remain largely inherited from the backbone prediction, while ESSC-RM mainly improves local semantic consistency and boundary delineation (e.g., thin objects and class-confusing regions), which naturally limits the headroom when the backbone output is already geometrically plausible.

\paragraph{ESSC-RM on MonoScene.}
The MonoScene block of Table~\ref{tab:QuantitativeResults} shows that ESSC-RM also improves the weaker MonoScene baseline.
VLGM increases mIoU from $11.08\%$ to $11.49\%$, and PNAM further pushes it to $11.51\%$ with comparable IoU.
These consistent gains across CGFormer and MonoScene support the plug-and-play nature of ESSC-RM, indicating that the refinement is not tied to a specific SSC backbone.

On MonoScene, ESSC-RM improves mIoU from $11.08\%$ to $11.51\%$ (+0.43, $\sim$4\% relative).
Since the refinement module does not introduce additional sensor-level geometric observations beyond the backbone output, its improvement is mainly achieved by enforcing multi-scale voxel consistency and reducing local misclassifications.
When large missing structures are completely absent in $\hat{\mathbf{Y}}$, post-hoc voxel refinement cannot fully recover them, whereas it remains effective at sharpening boundaries and improving local semantic coherence.

\paragraph{SC-IoU trade-off.}
Although refinement improves mIoU, SC-IoU (occupied vs.\ free) can slightly decrease in some cases.
For example, on CGFormer, ESSC-RM increases mIoU by +0.40 (16.87\%$\rightarrow$17.27\%) while SC-IoU decreases by 1.66 (45.99\%$\rightarrow$44.33\%).
This behavior is expected because SC-IoU is a \emph{binary} occupancy metric that is particularly sensitive to boundary voxels: semantic refinement around thin structures and object borders may flip a small fraction of occupied/free decisions, increasing FP/FN near boundaries even when per-class semantics improve.
As SC-IoU aggregates over the entire occupancy grid, such boundary perturbations can lead to a measurable IoU change, reflecting a mild trade-off between semantic correction (mIoU) and boundary-sensitive binary occupancy under discrete voxel predictions.

\paragraph{Refinement module efficiency.}
We further analyze the computational overhead of ESSC-RM on top of CGFormer (Table~\ref{tab:EfficienctResults}).
CGFormer itself has 122.42M parameters, requires about $19.3$ GB memory during training and $6.55$ GB at inference, and runs at approximately $205$ ms per frame.
The 3D U-Net refinement head adds only 13.36M parameters and can be trained jointly with CGFormer on a 24GB GPU when the backbone is set to inference mode.
VLGM and PNAM increase parameter counts and inference time more noticeably, but remain practical for offline refinement or two-stage pipelines.

\begin{table}[htbp]
  \centering
  \resizebox{\textwidth}{!}{%
    \begin{tabular}{@{}c|cc|cccc@{}}
      \toprule
      Model & IoU & mIoU & Params (M) & Train Memory (M) & Infer. Memory (M) & Infer. Time (ms) \\
      \midrule
      CGFormer & \textbf{45.99} & 16.87 & 122.42 & 19330 & 6550 & 205 \\
      \midrule
      +3D U-Net & 43.53 & 17.17 & 13.36 & 12726 & 4904 & 215 \\
      [3pt]
      +VLGM & 43.20 & 17.21 & 43.96 & 18942 & 5382 & 340 \\
      [3pt]
      +PNAM & 44.33 & \textbf{17.27} & 9.59 & 20664 & 5042 & 265 \\
      \bottomrule
    \end{tabular}%
  }
  \caption{Ablation study on the efficiency of the refinement module with CGFormer as backbone.}
  \label{tab:EfficienctResults}
\end{table}

\begin{figure}[h!]
  \centering
  \setlength{\tabcolsep}{2pt}
  \renewcommand{\arraystretch}{0.0}

  \newcommand{\rgbpic}[1]{%
    \raisebox{5ex}{\includegraphics[width=0.15\textwidth]{#1}}%
  }

  \begin{tabular}{cccccc}
    % ---- Row 1 ----
    \rgbpic{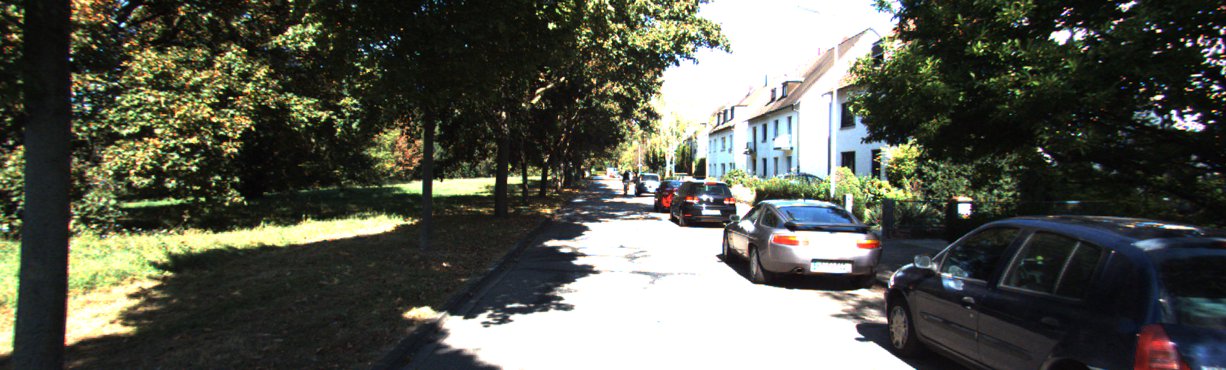} &
    \includegraphics[width=0.15\textwidth]{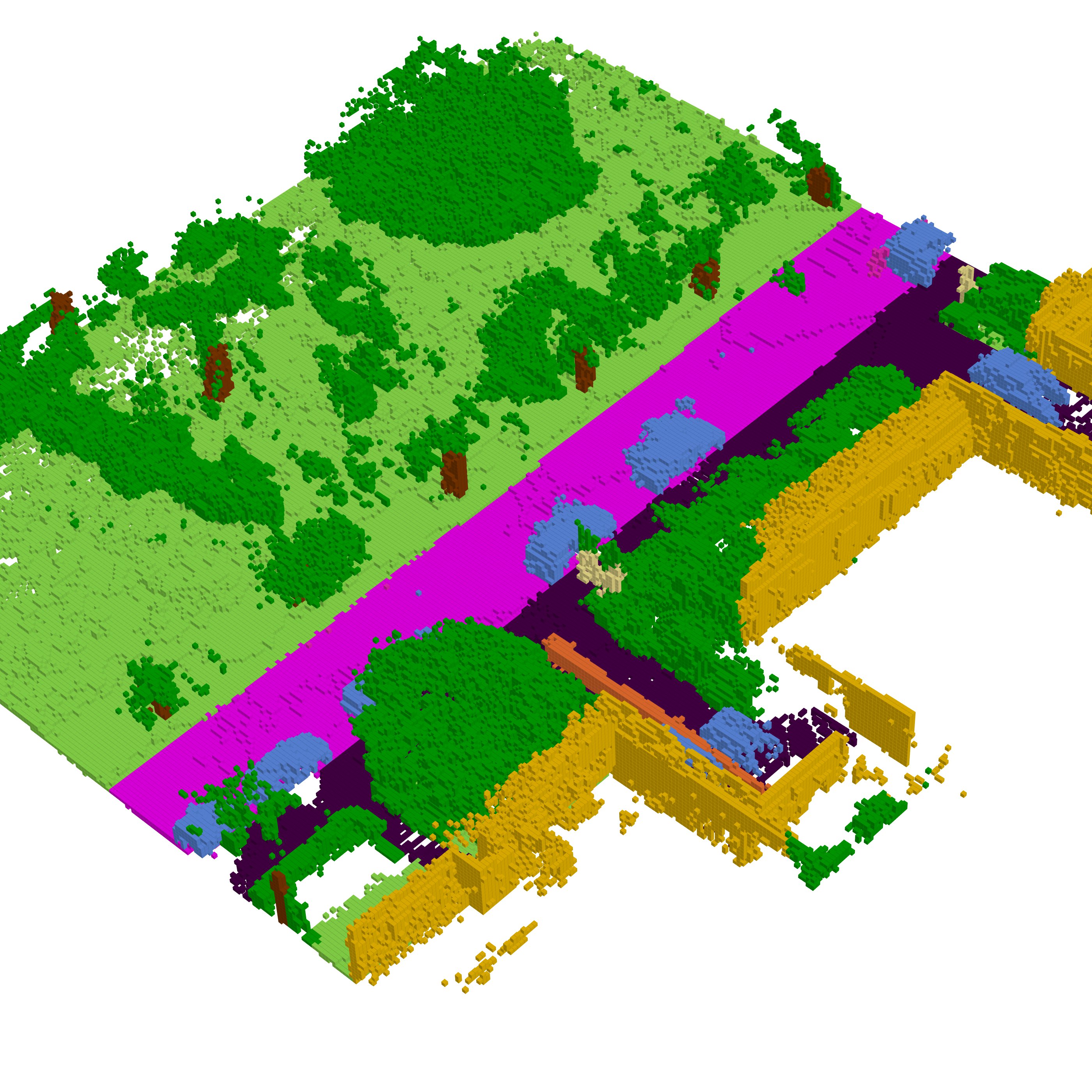} &
    \includegraphics[width=0.15\textwidth]{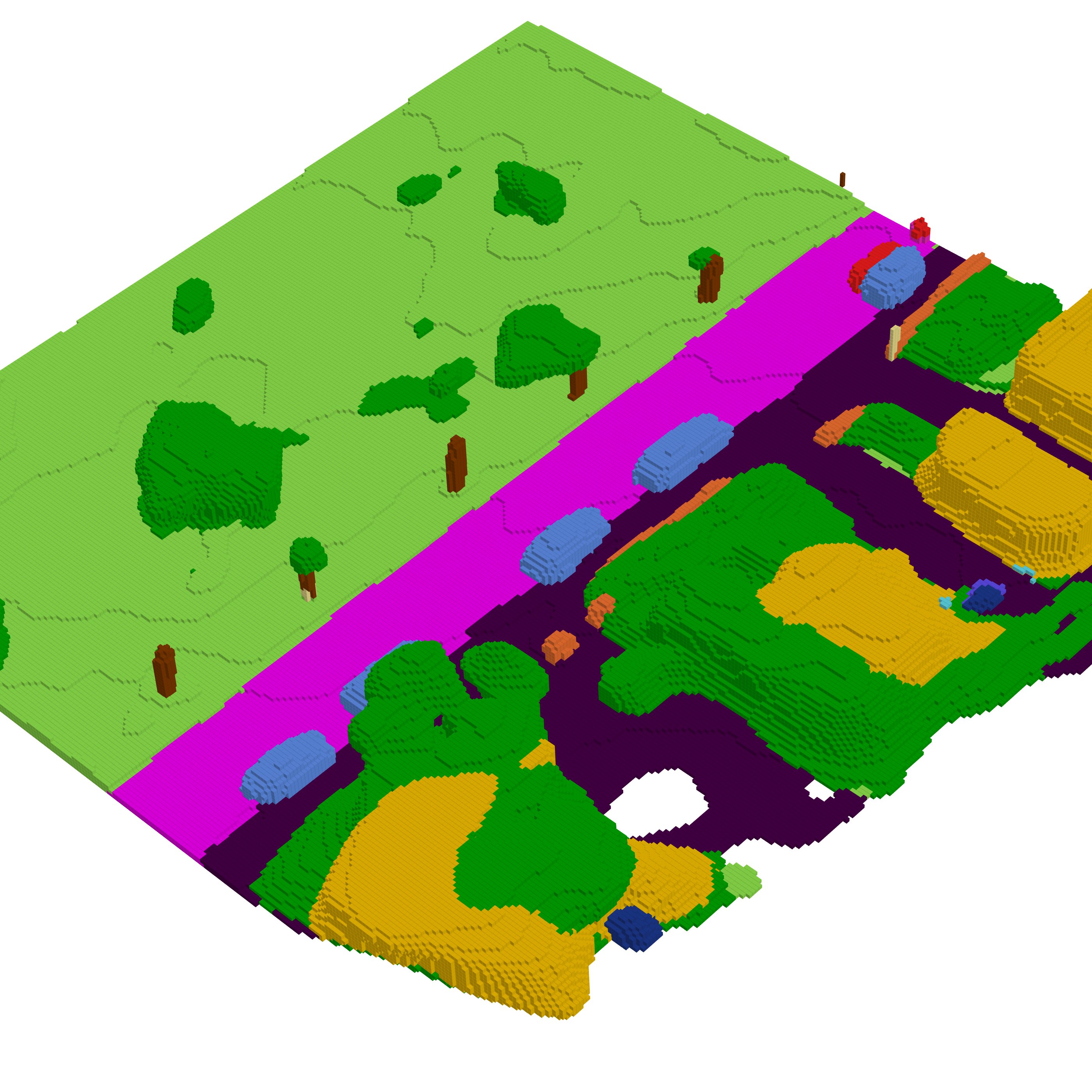} &
    \includegraphics[width=0.15\textwidth]{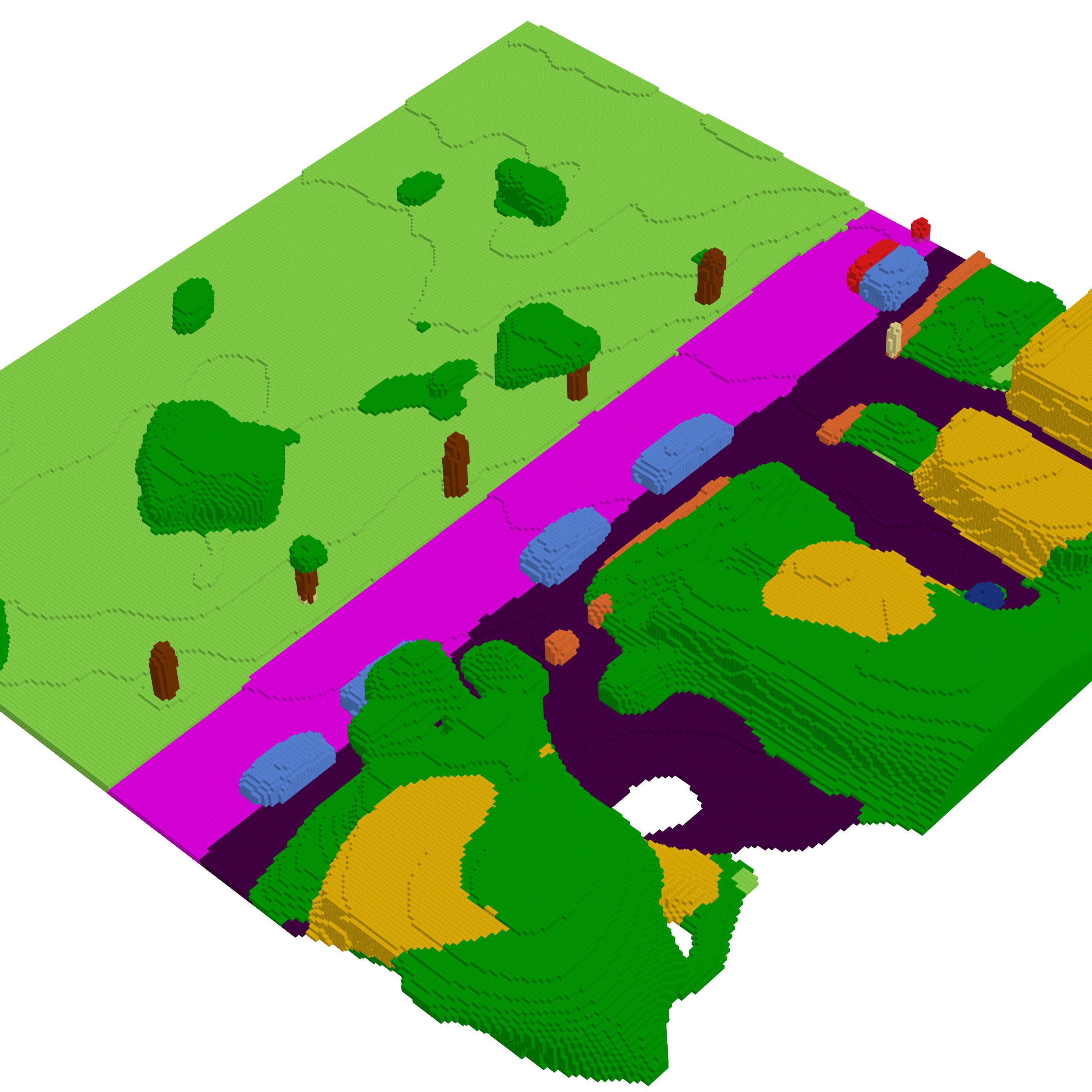} &
    \includegraphics[width=0.15\textwidth]{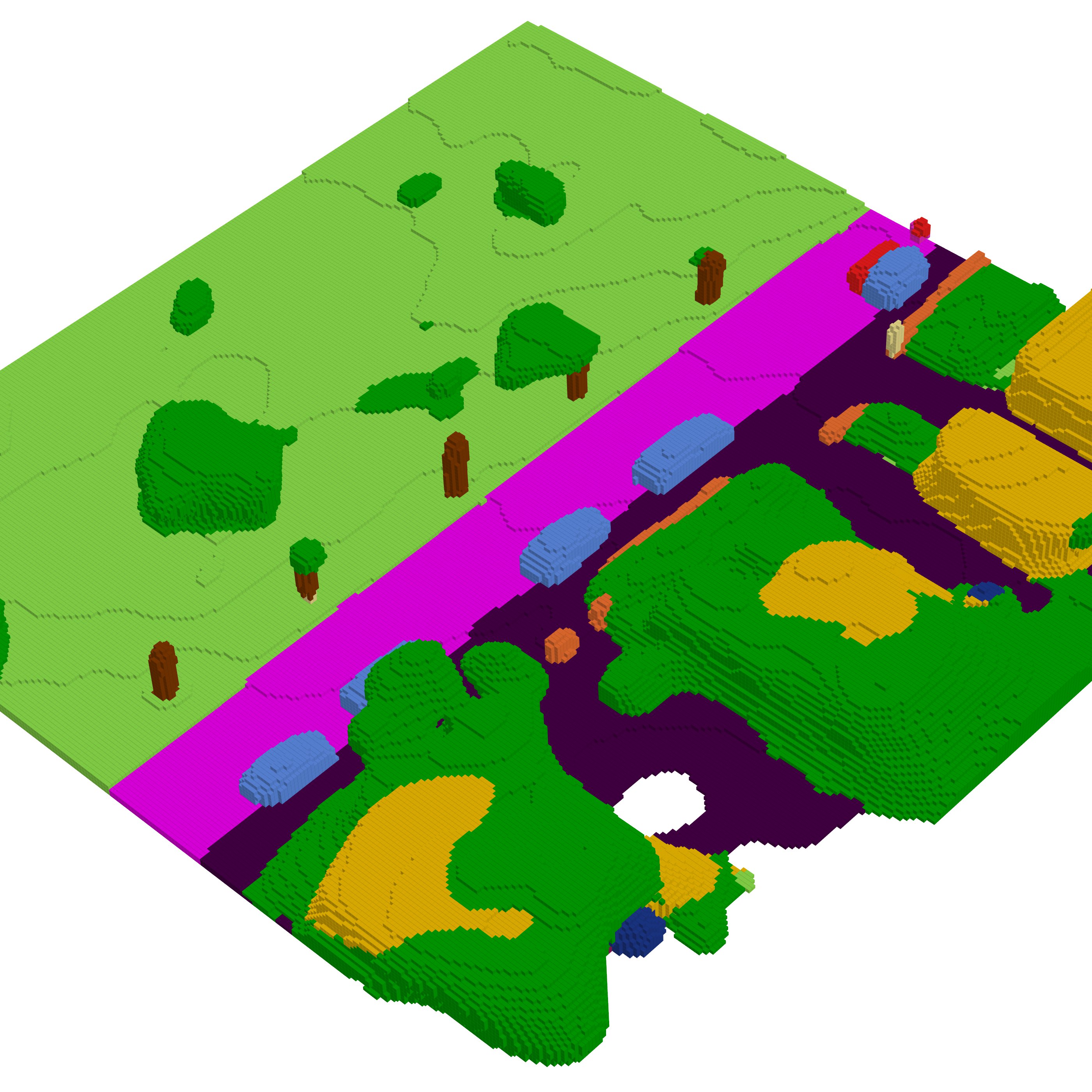} &
    \includegraphics[width=0.15\textwidth]{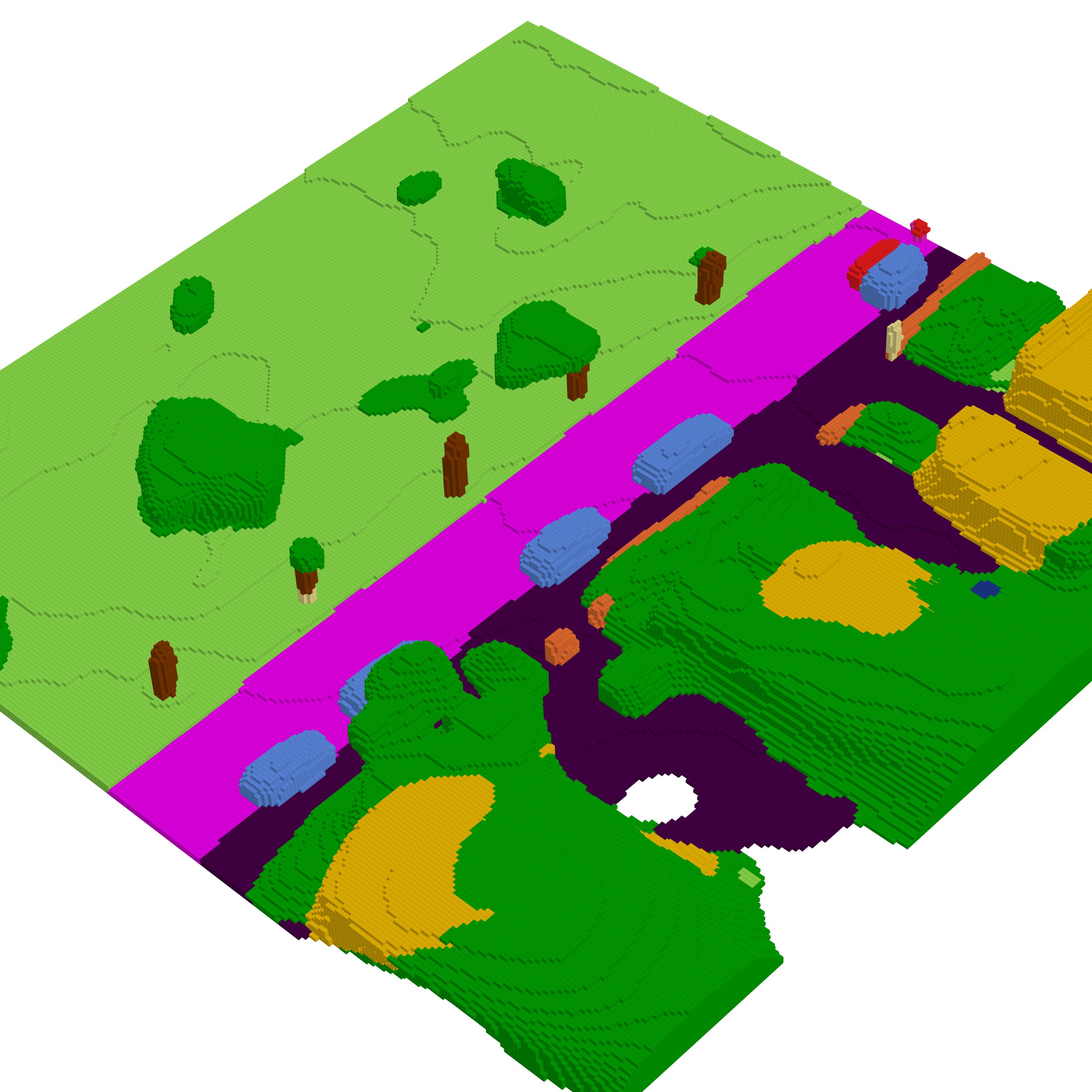} \\
    [0.3em]
    % ---- Row 2 ----
    \rgbpic{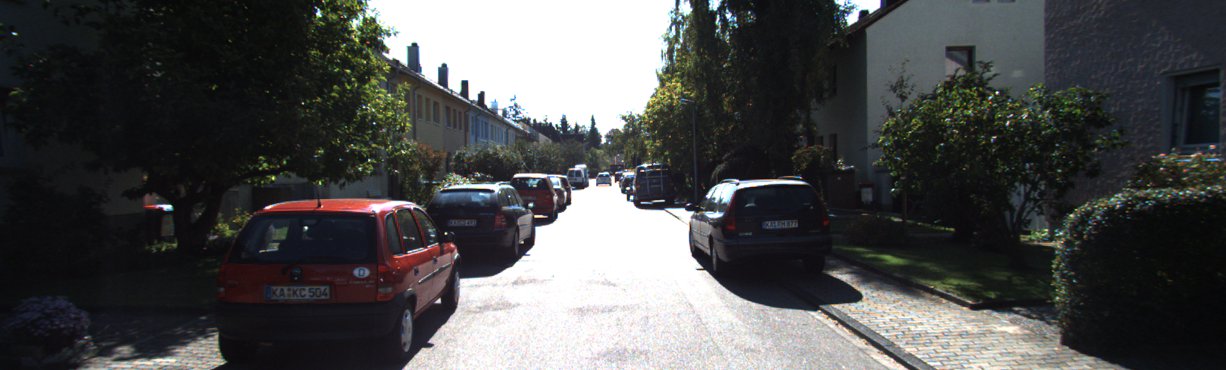} &
    \includegraphics[width=0.15\textwidth]{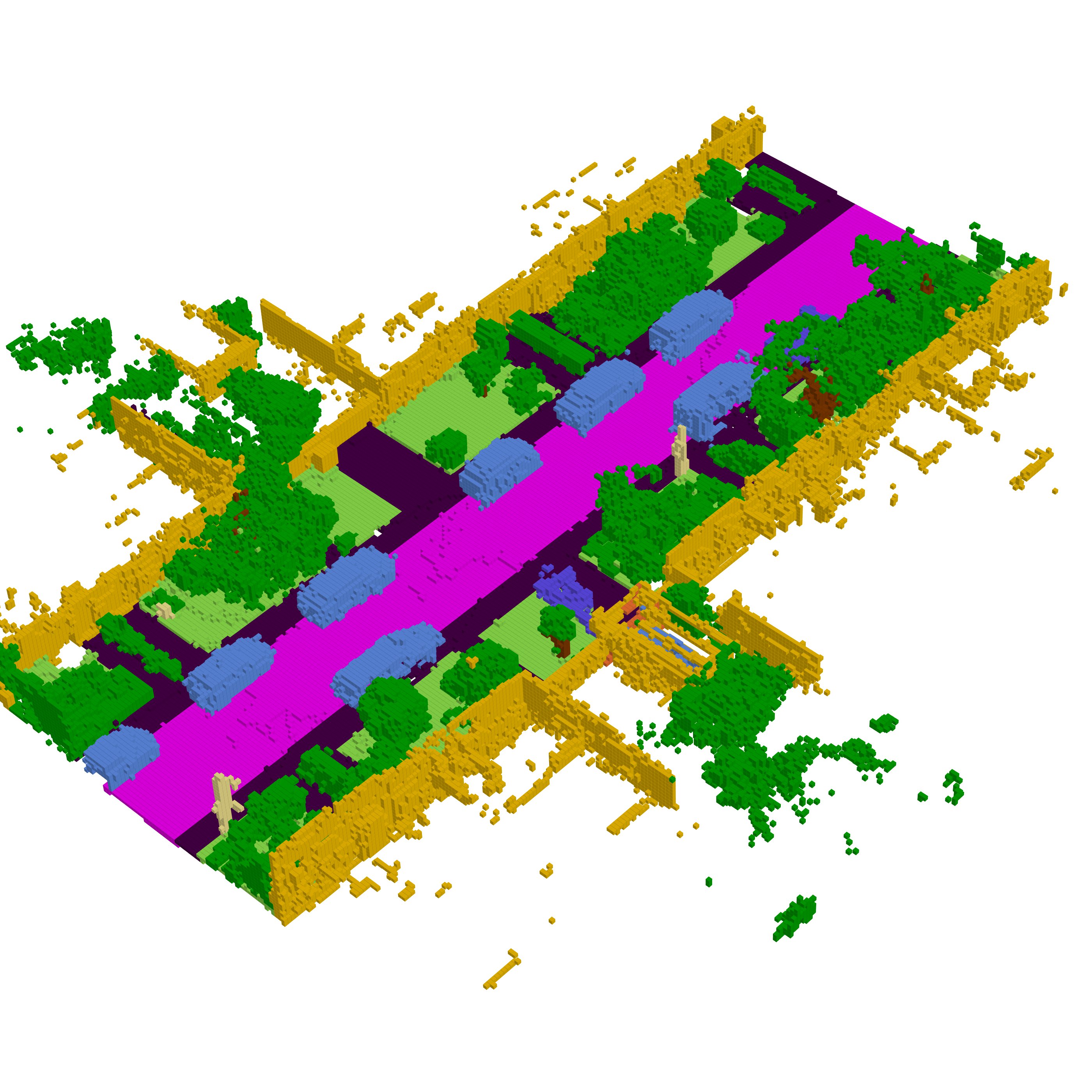} &
    \includegraphics[width=0.15\textwidth]{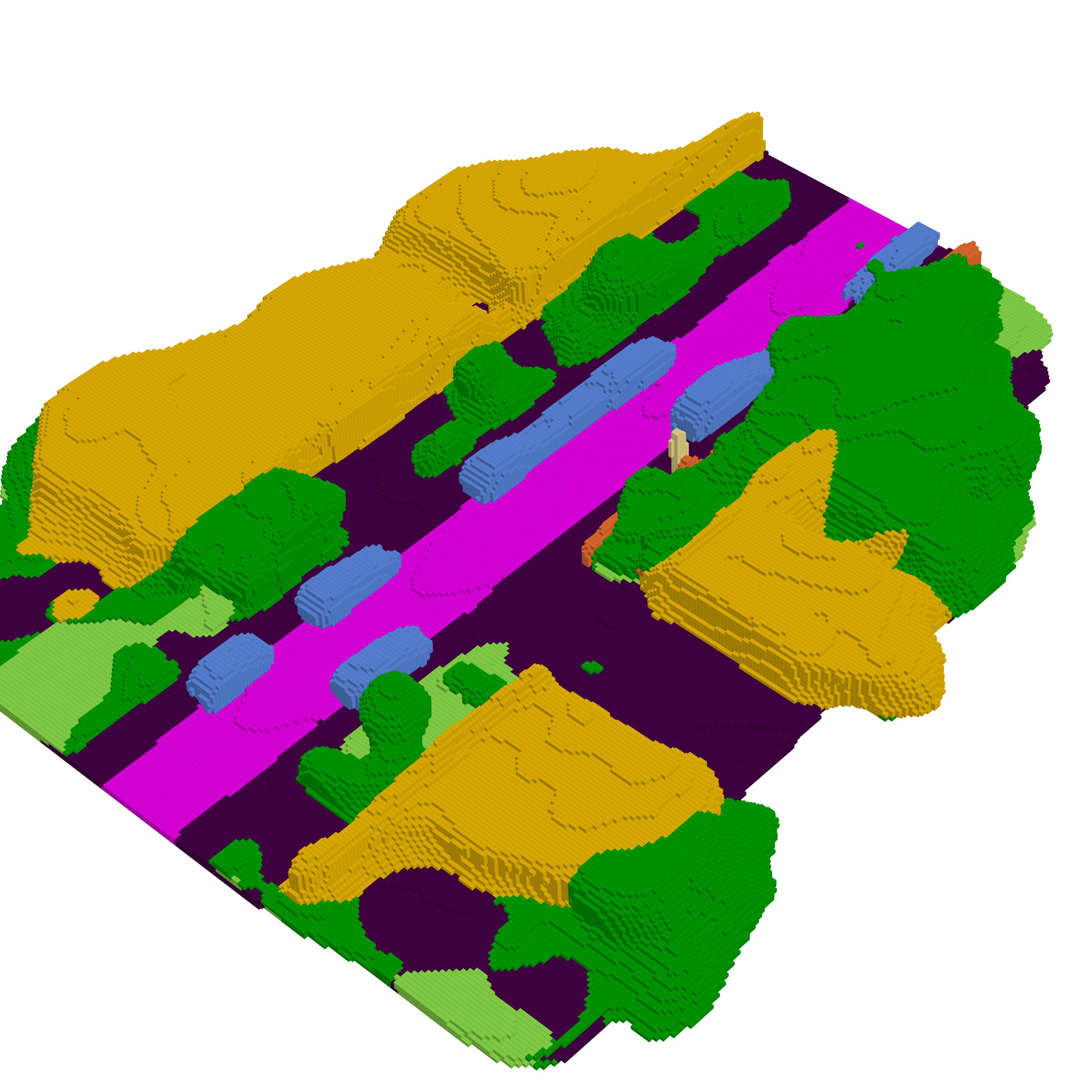} &
    \includegraphics[width=0.15\textwidth]{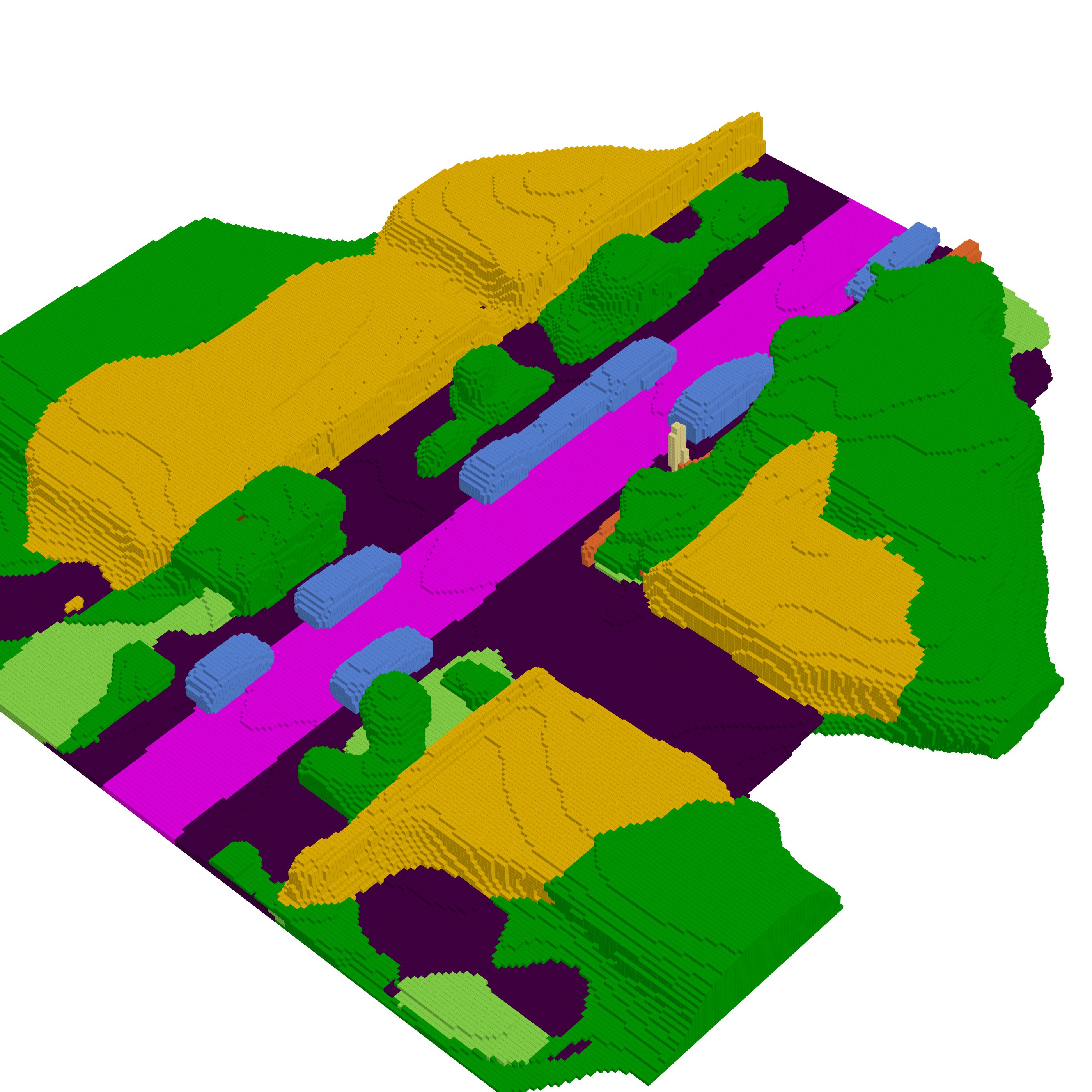} &
    \includegraphics[width=0.15\textwidth]{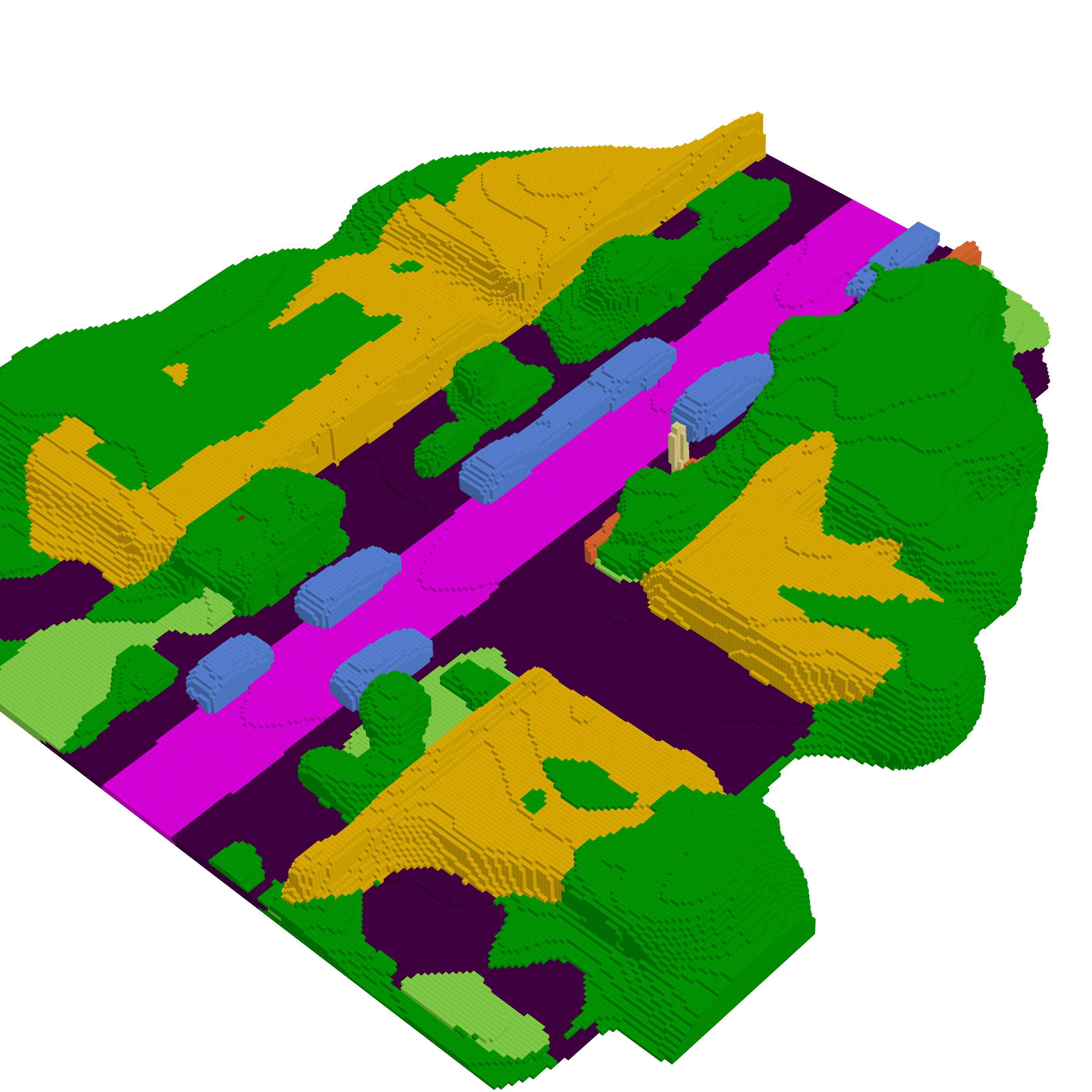} &
    \includegraphics[width=0.15\textwidth]{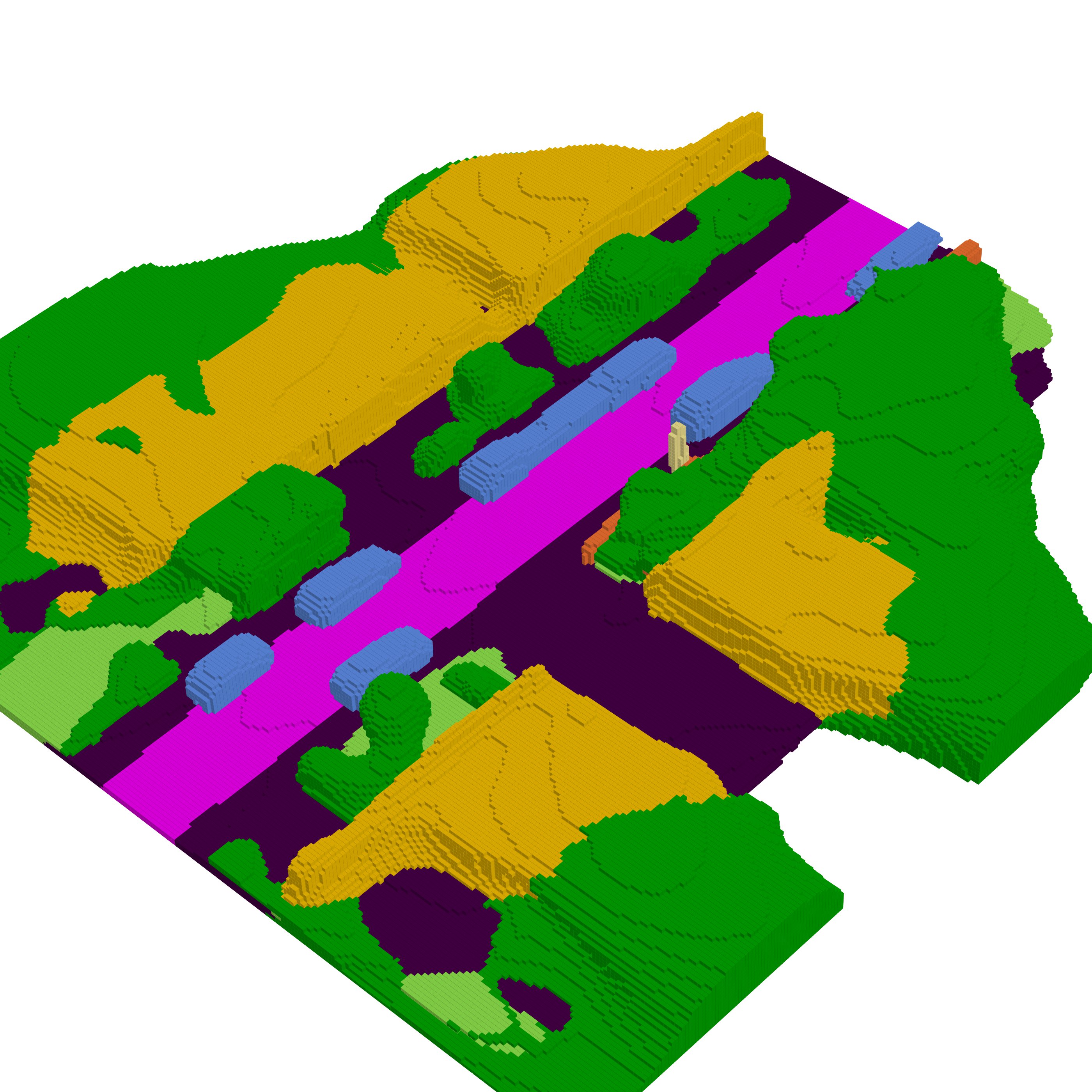} \\
    [0.3em]
    % ---- Row 3 ----
    \rgbpic{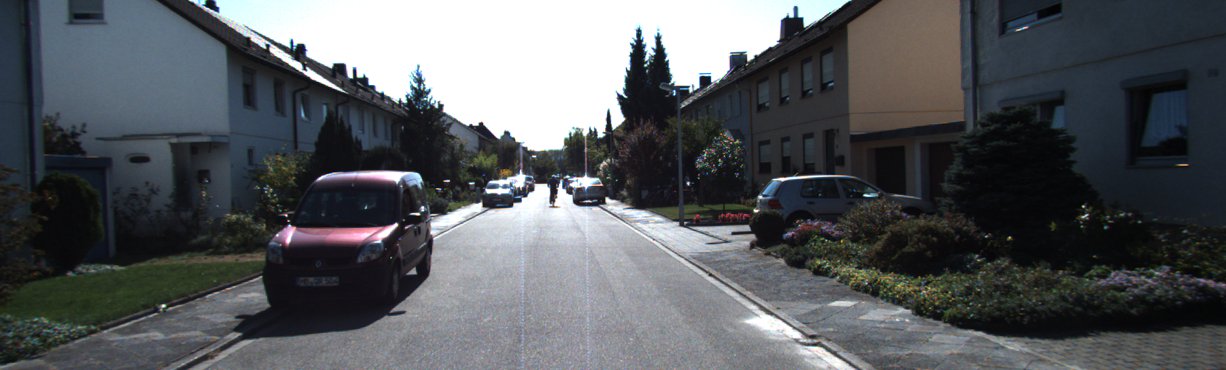} &
    \includegraphics[width=0.15\textwidth]{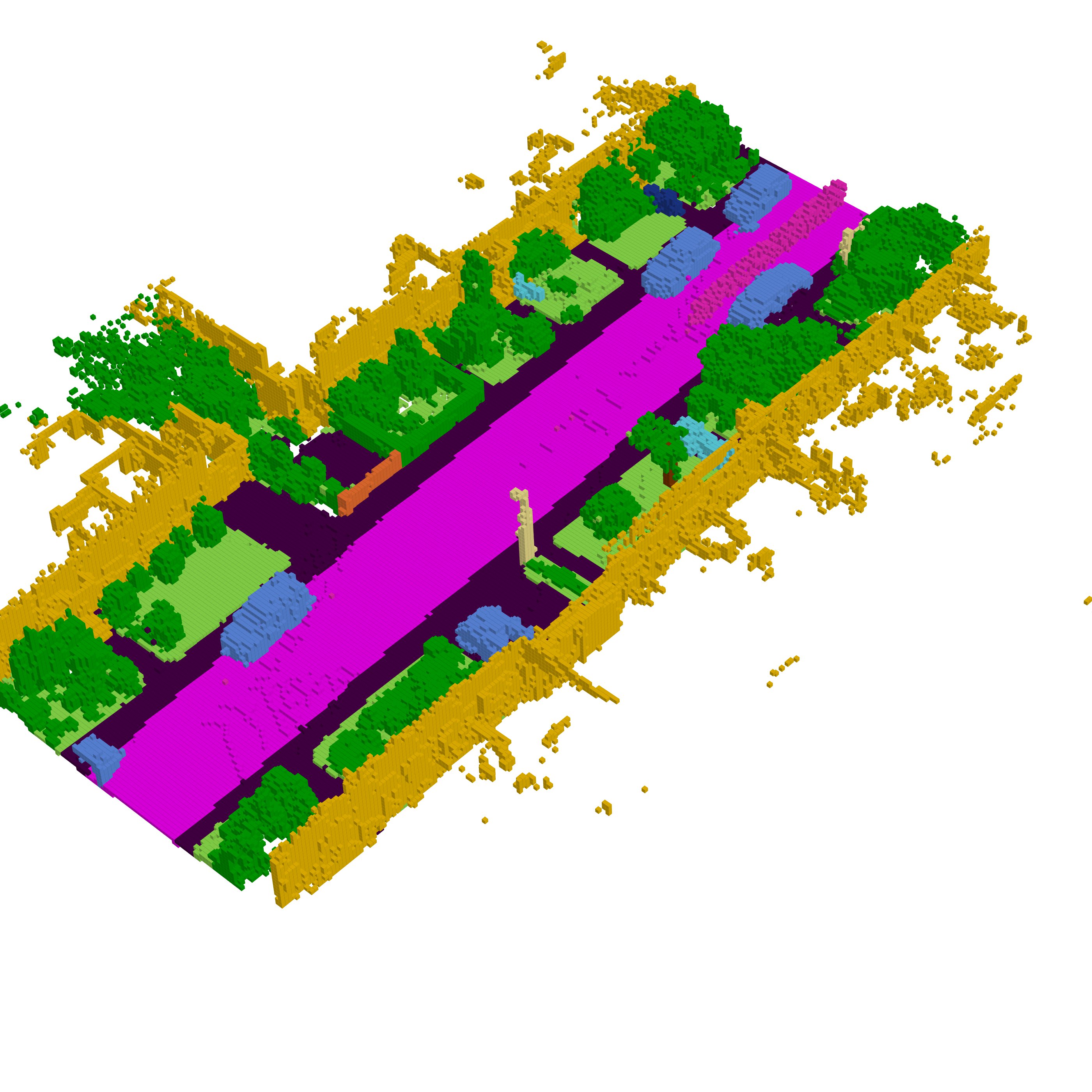} &
    \includegraphics[width=0.15\textwidth]{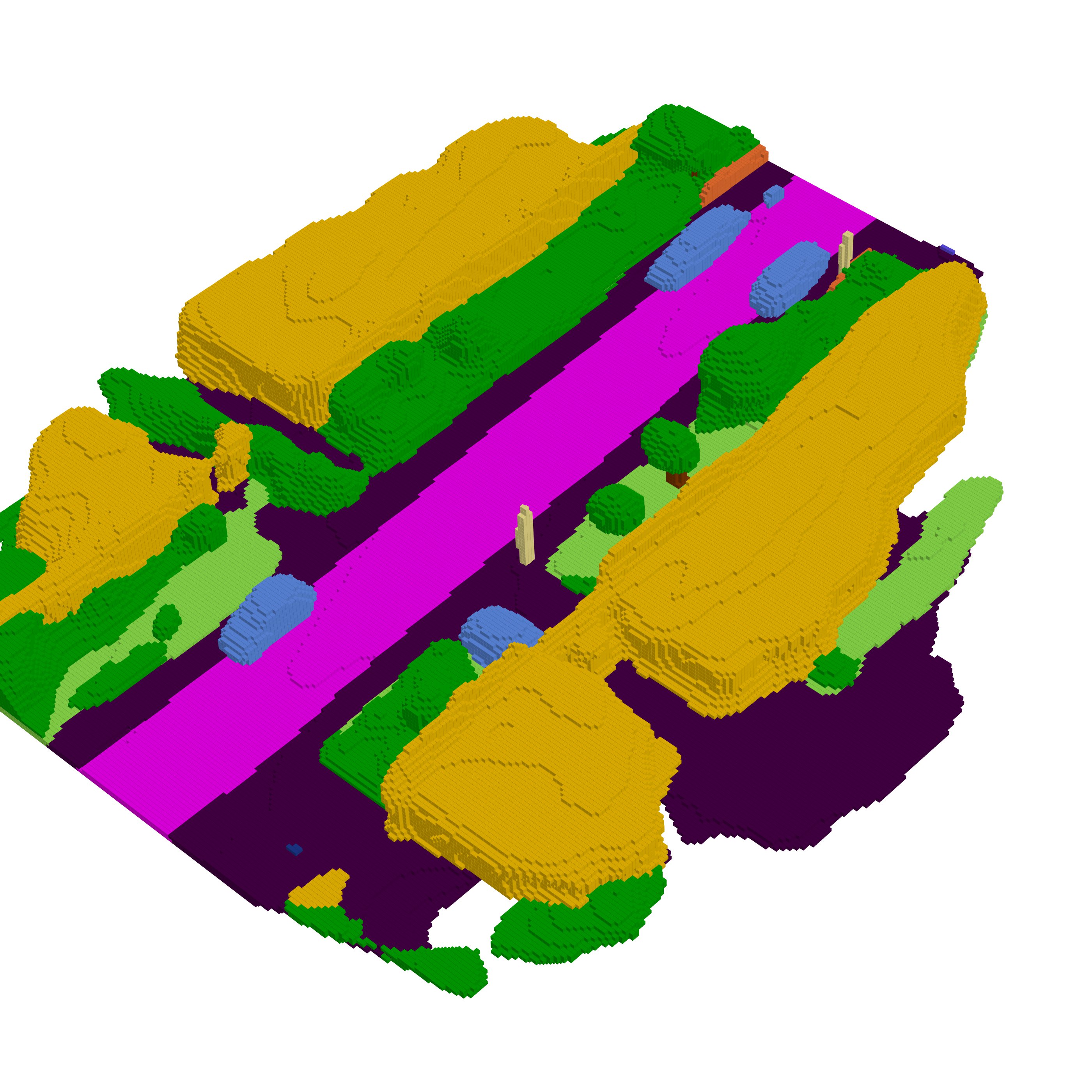} &
    \includegraphics[width=0.15\textwidth]{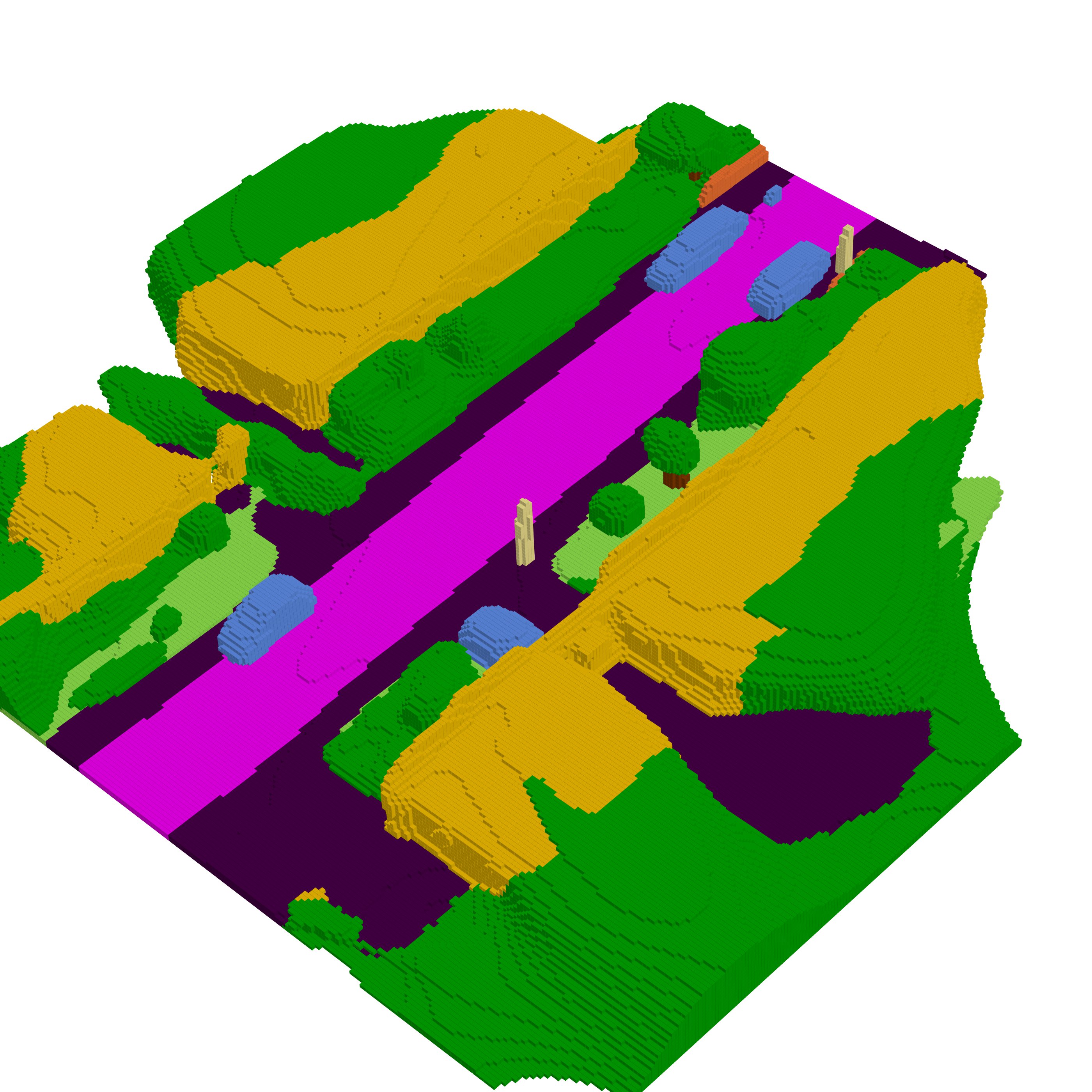} &
    \includegraphics[width=0.15\textwidth]{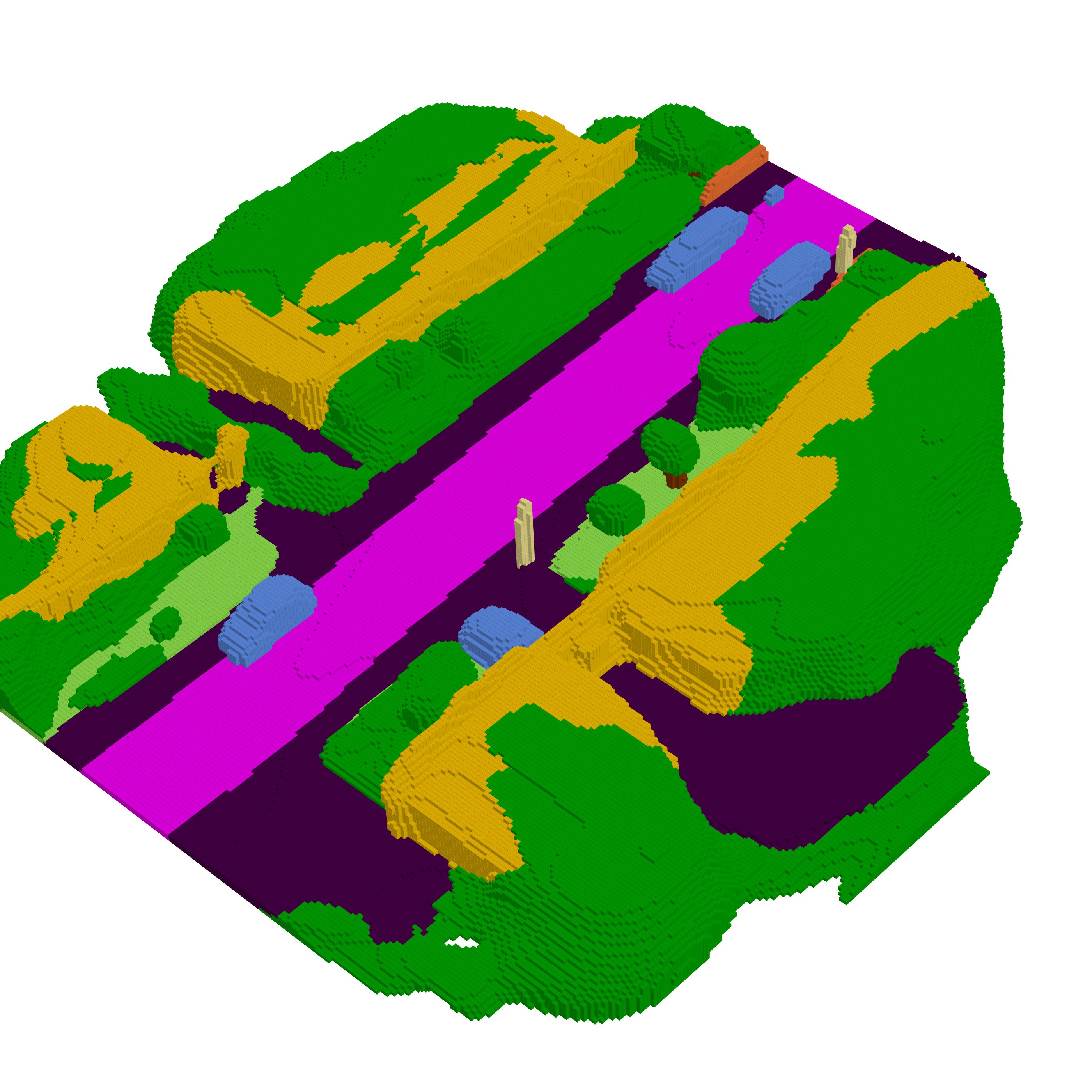} &
    \includegraphics[width=0.15\textwidth]{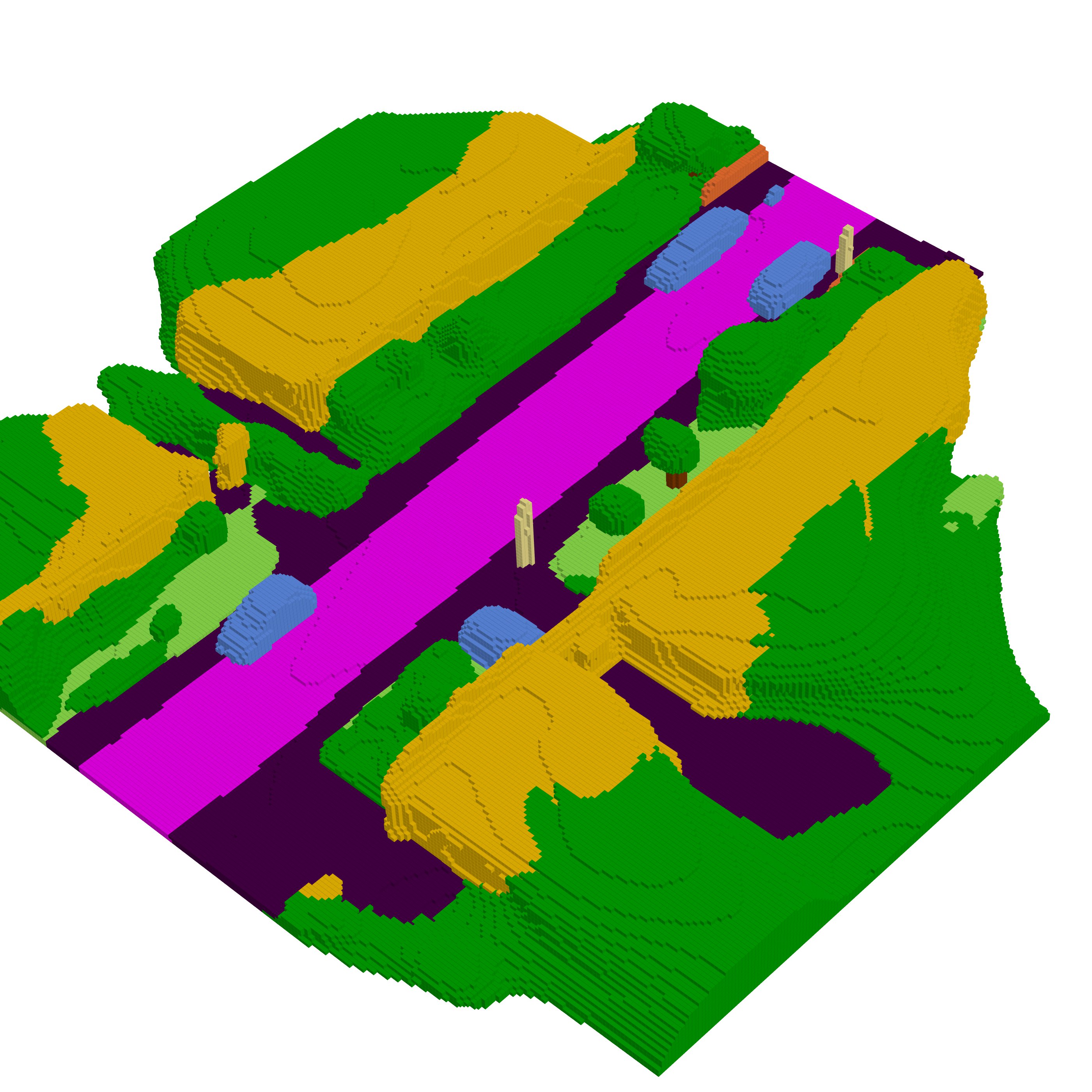} \\
    [0.4em]
    % ---- Legend row ----
    {\small RGB} &
    {\small Ground Truth} &
    {\small CGFormer} &
    {\small + 3D U\mbox{-}Net} &
    {\small + PNAM} &
    {\small + VLGM}
  \end{tabular}

  \caption{Qualitative results of ESSC-RM on CGFormer~\cite{2023_CGFormer}
  on the SemanticKITTI~\cite{2019_SemanticKITTI,2021_SemanticKITTI_2} validation set.
  ESSC-RM progressively refines the baseline prediction, filling missing regions
  and improving object shapes and small semantic structures.}
  \label{fig:qualitative_CGFormer}
\end{figure}

\begin{figure}[ht!]
  \centering
  \setlength{\tabcolsep}{2pt}
  \renewcommand{\arraystretch}{0.0}

  \newcommand{\rgbpic}[1]{%
    \raisebox{5ex}{\includegraphics[width=0.15\textwidth]{#1}}%
  }

  \begin{tabular}{cccccc}
    % ---- Row 1 ----
    \rgbpic{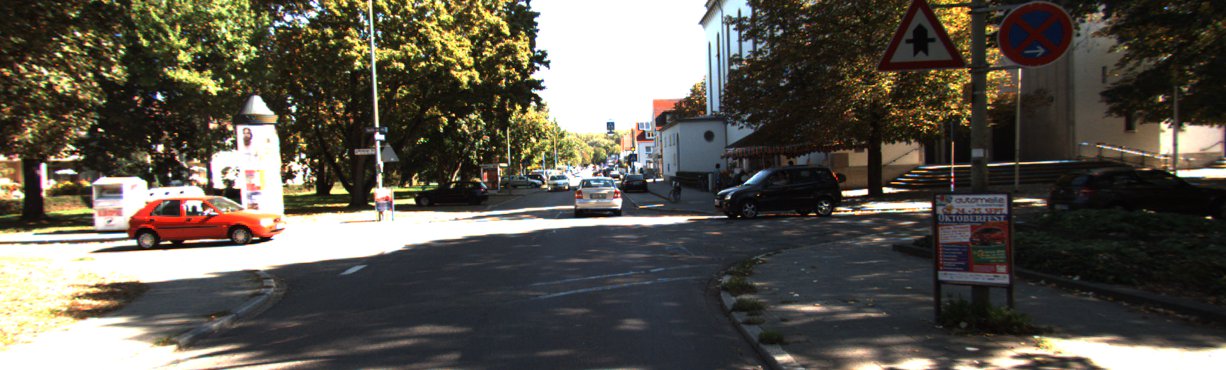} &
    \includegraphics[width=0.15\textwidth]{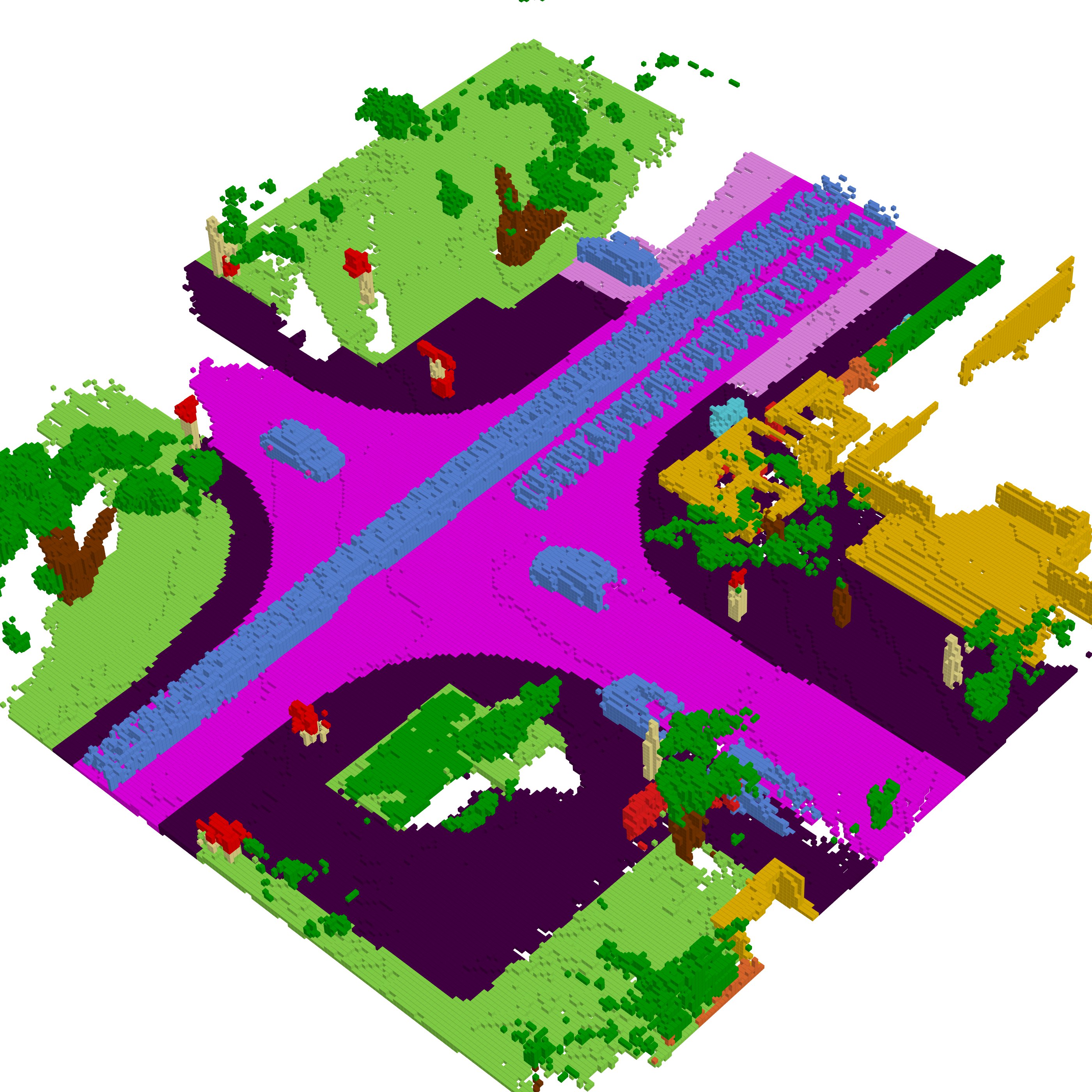} &
    \includegraphics[width=0.15\textwidth]{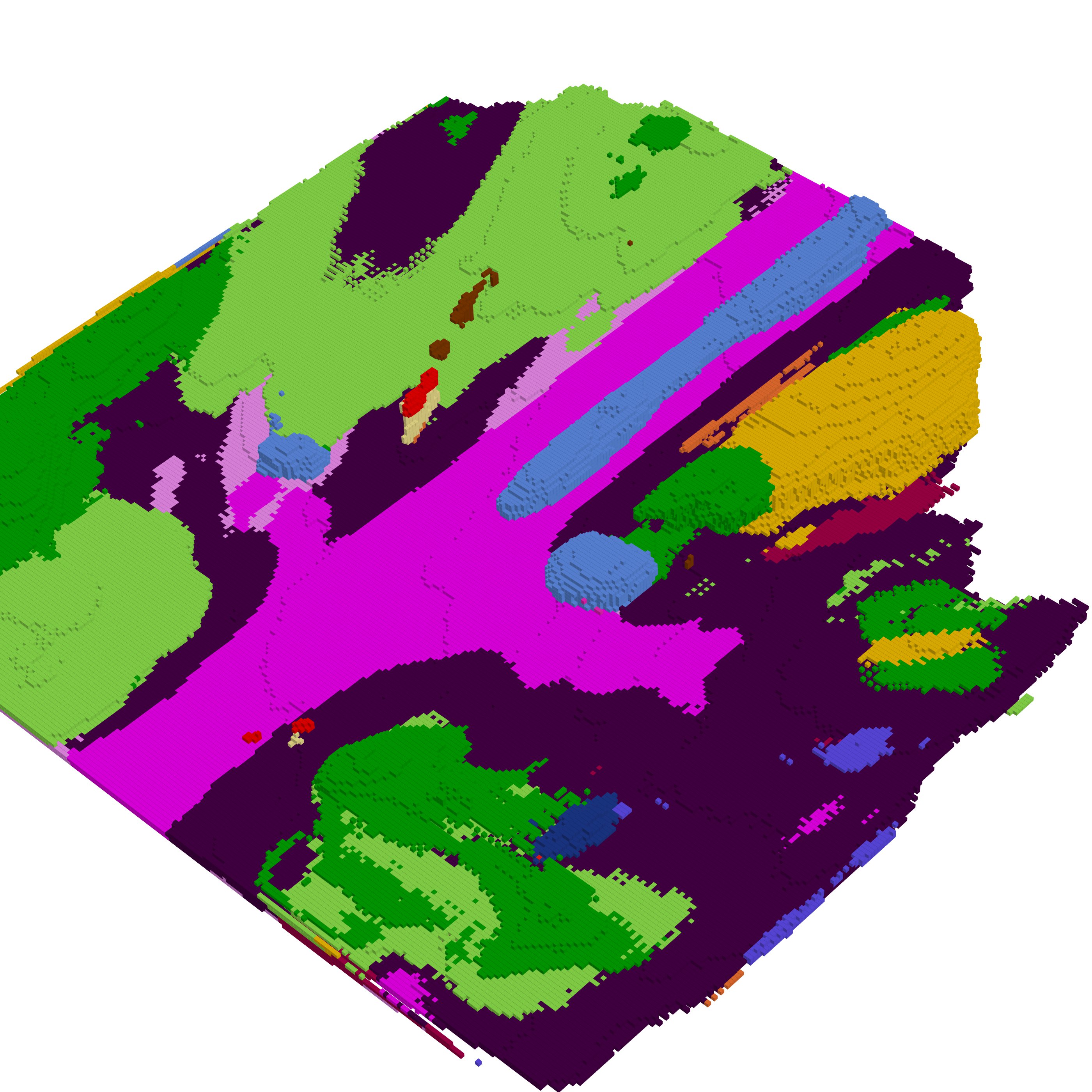} &
    \includegraphics[width=0.15\textwidth]{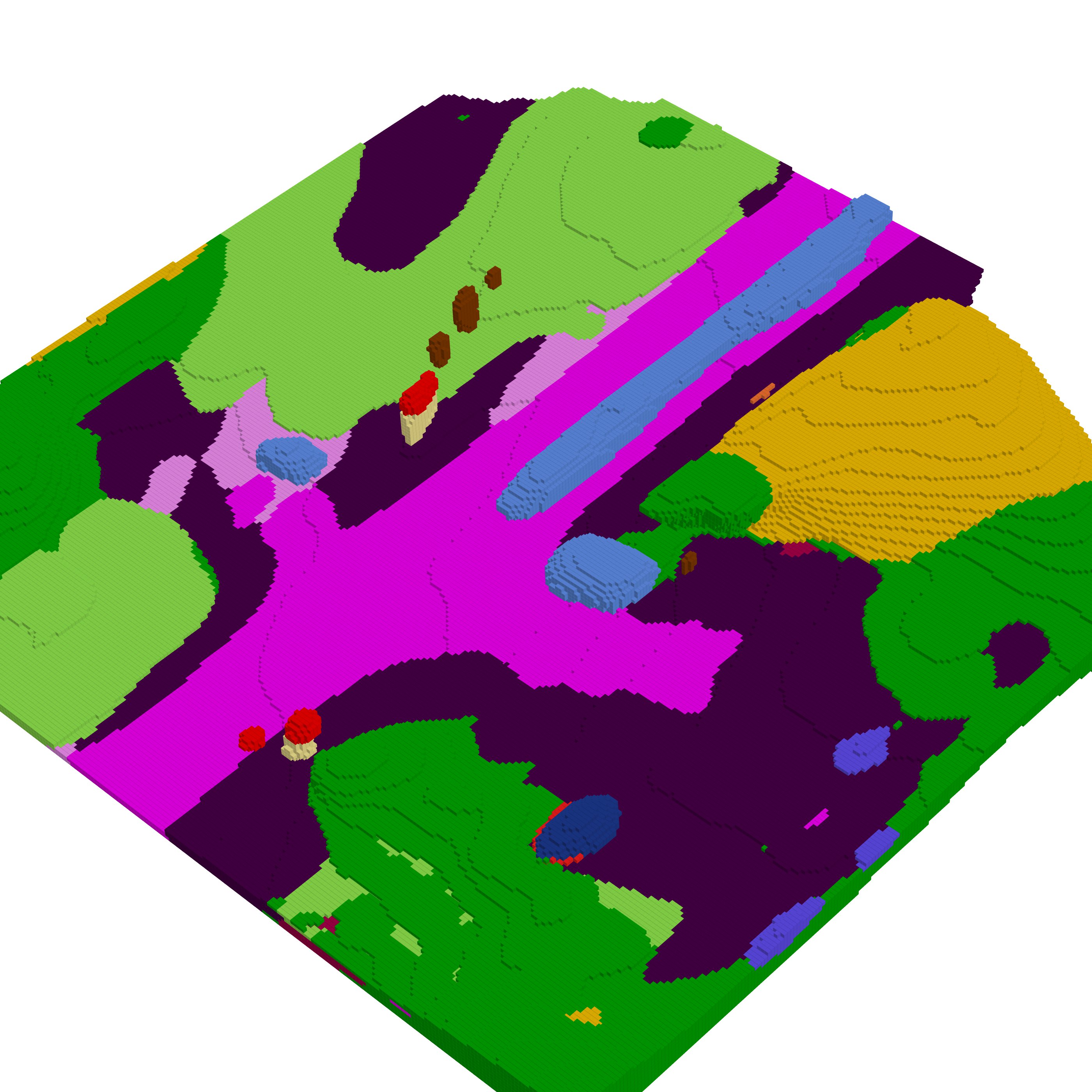} &
    \includegraphics[width=0.15\textwidth]{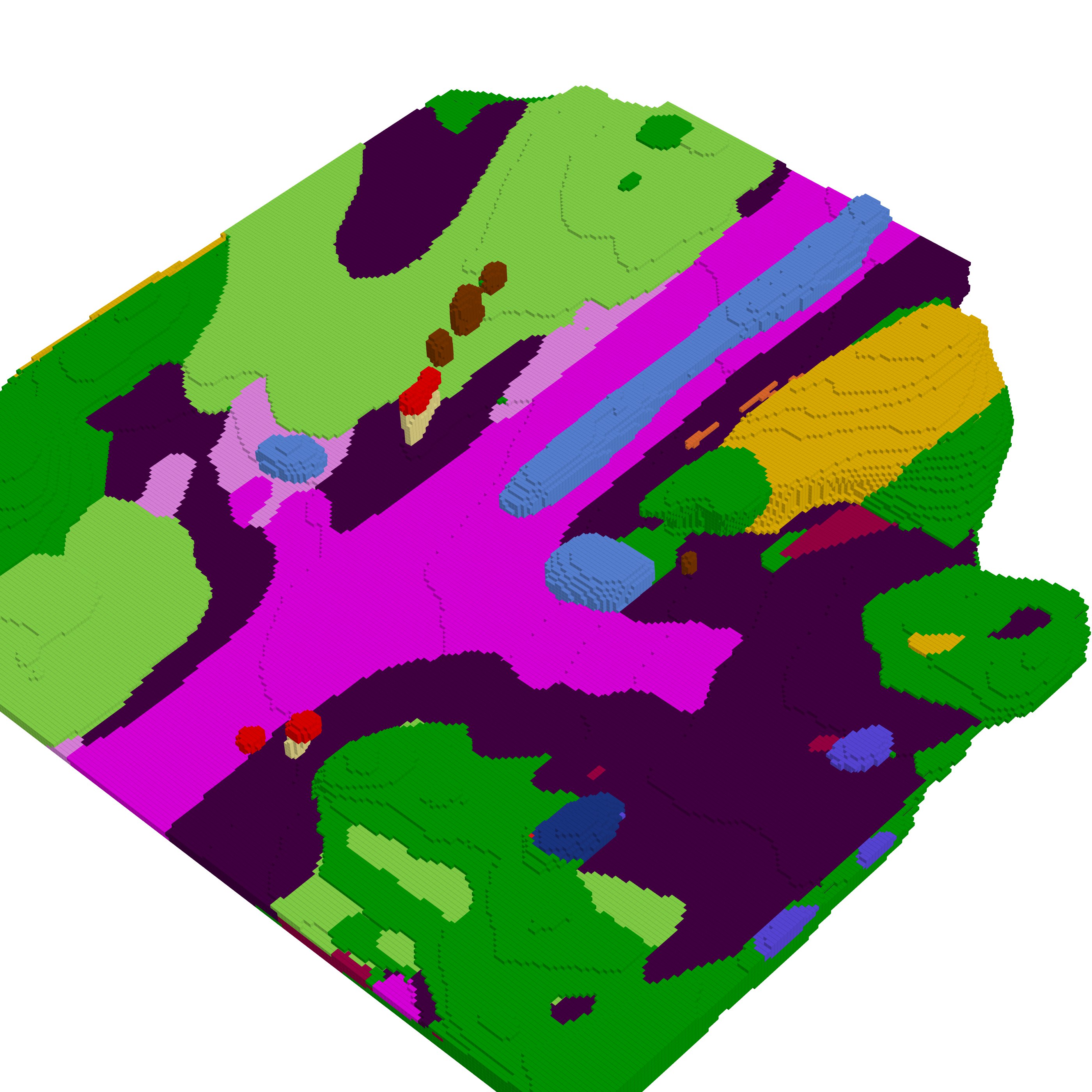} &
    \includegraphics[width=0.15\textwidth]{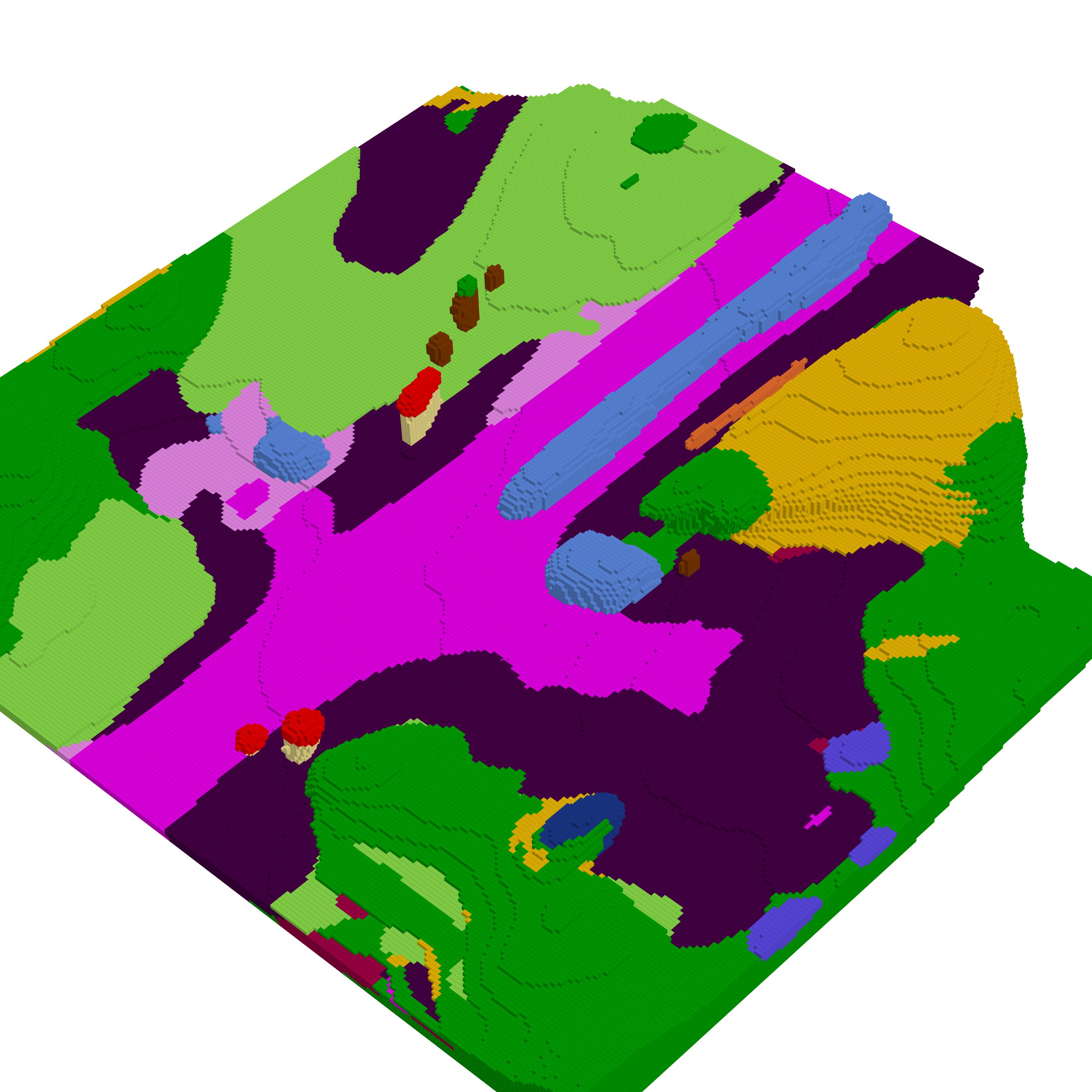} \\
    [0.3em]

    % ---- Row 2 ----
    \rgbpic{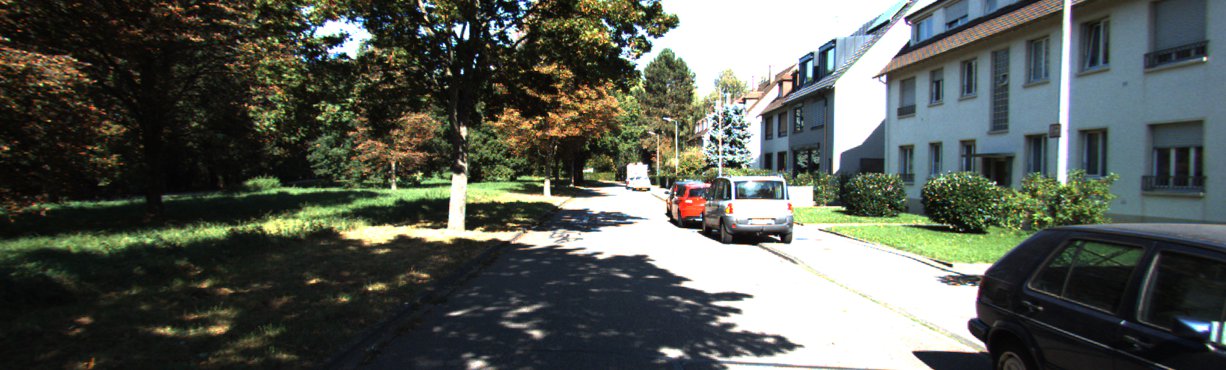} &
    \includegraphics[width=0.15\textwidth]{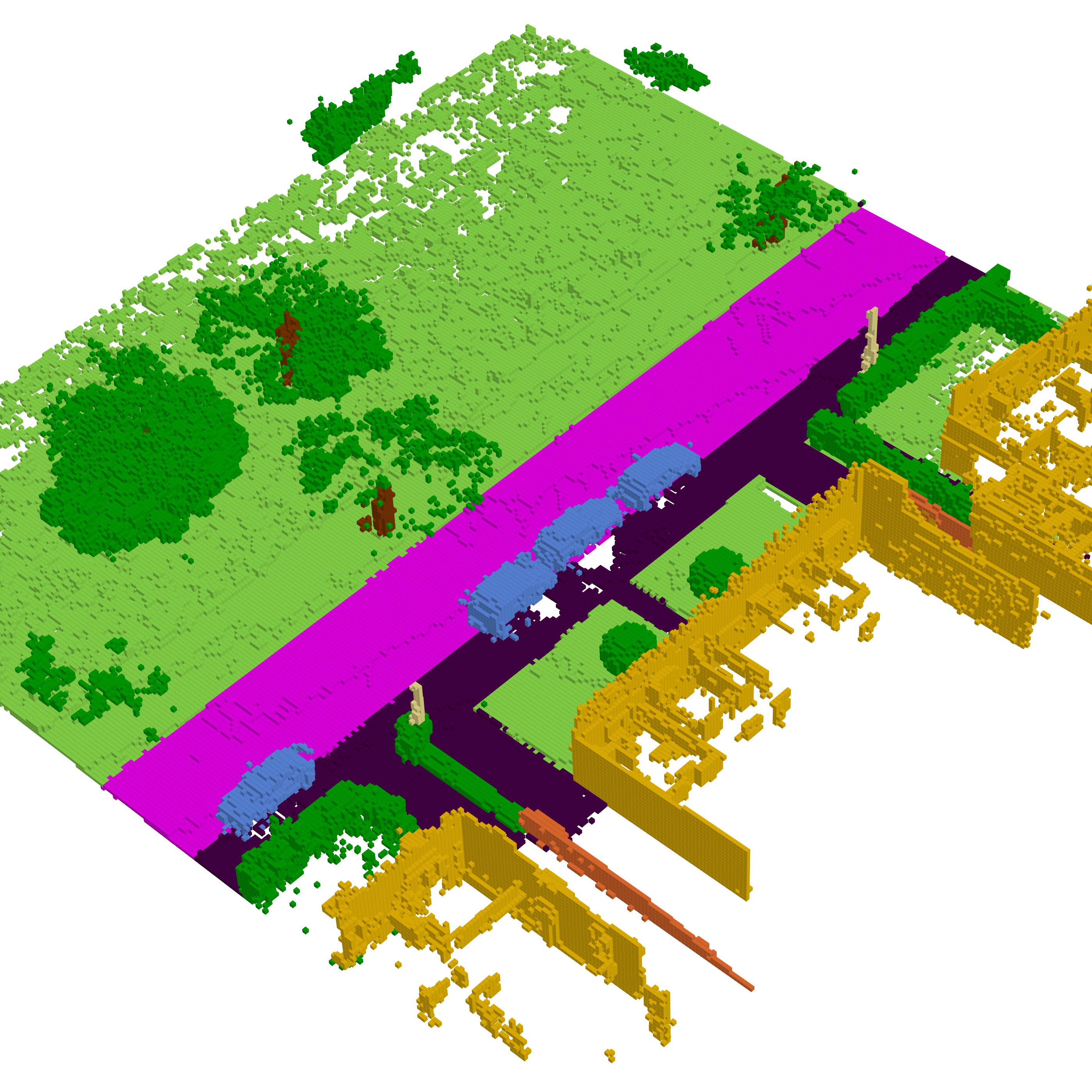} &
    \includegraphics[width=0.15\textwidth]{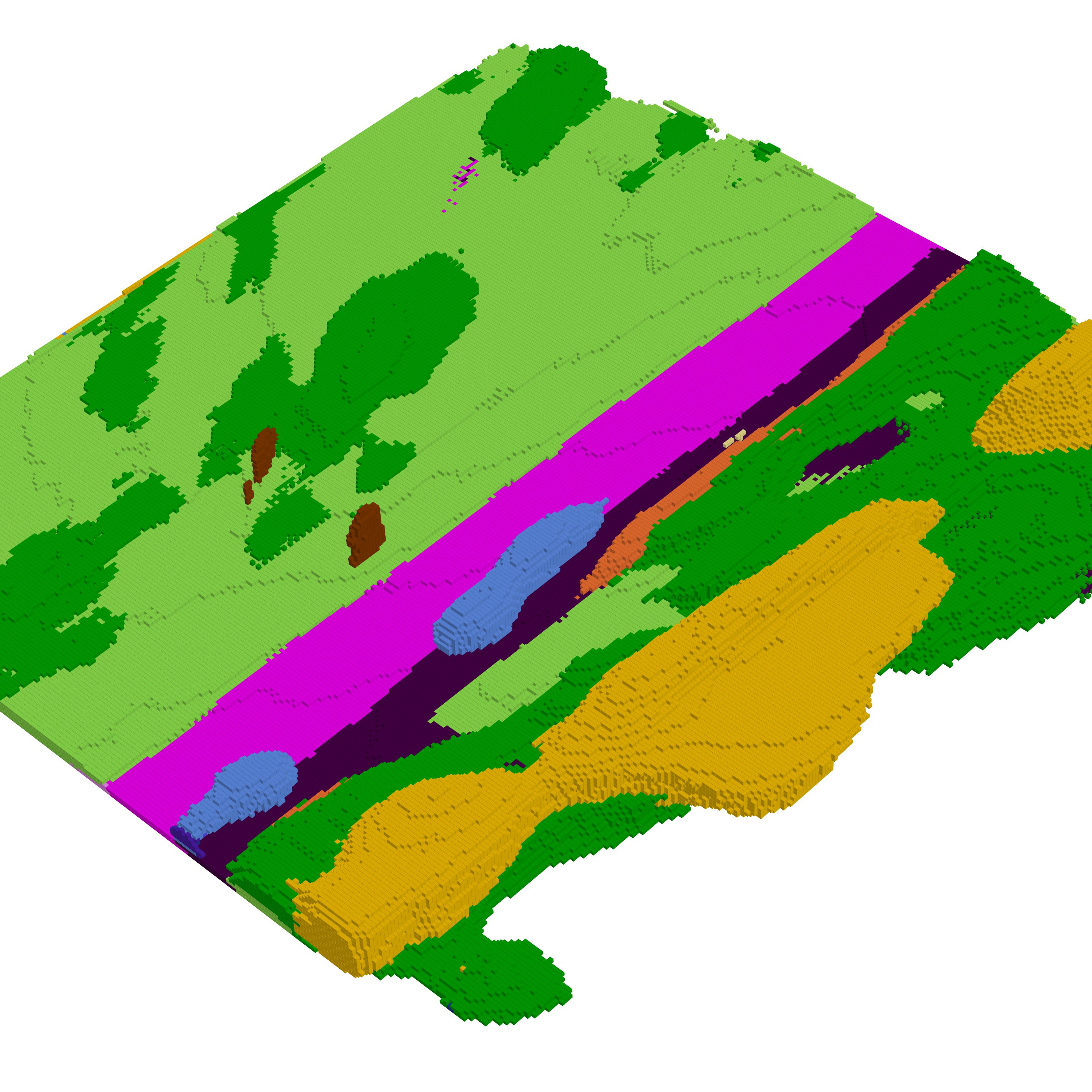} &
    \includegraphics[width=0.15\textwidth]{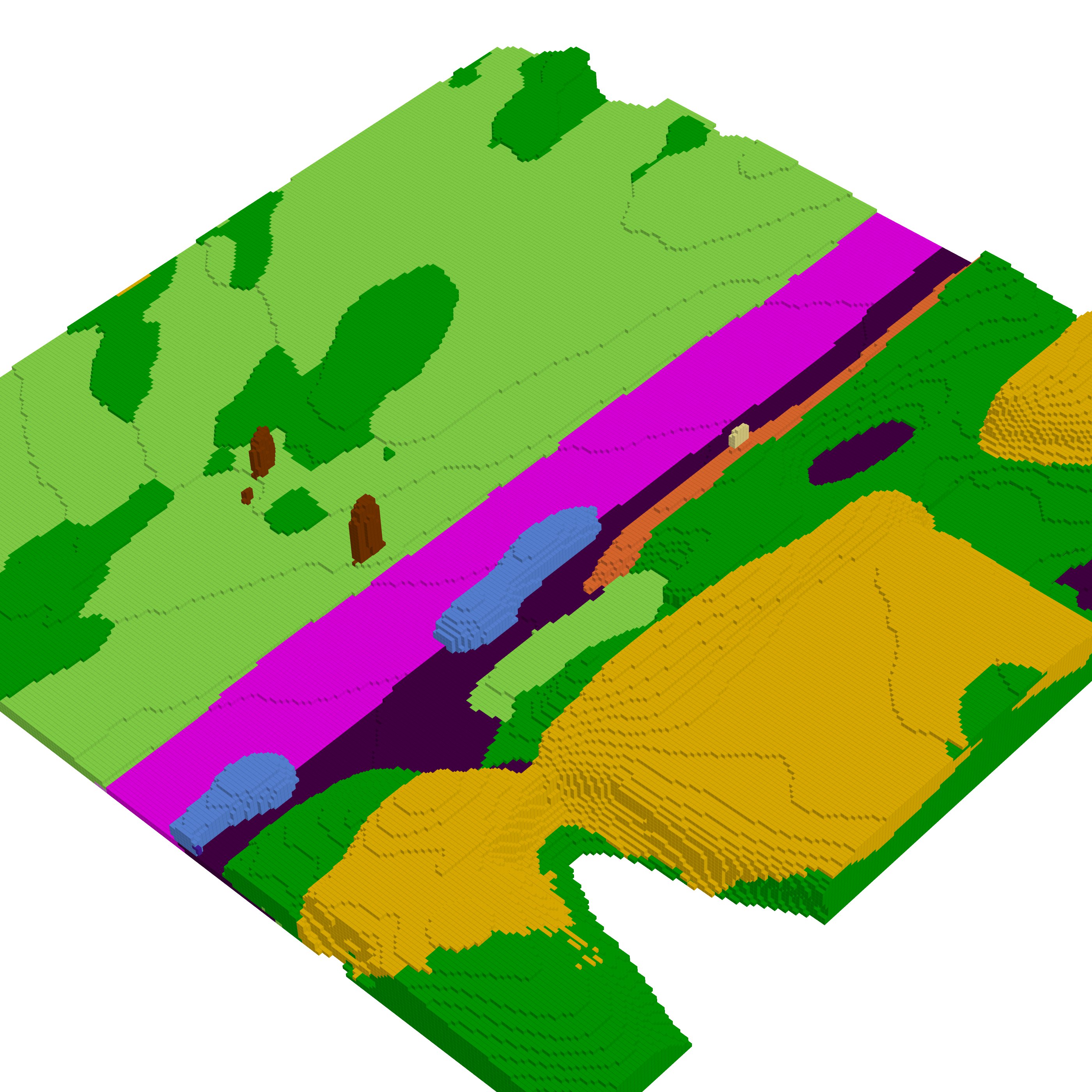} &
    \includegraphics[width=0.15\textwidth]{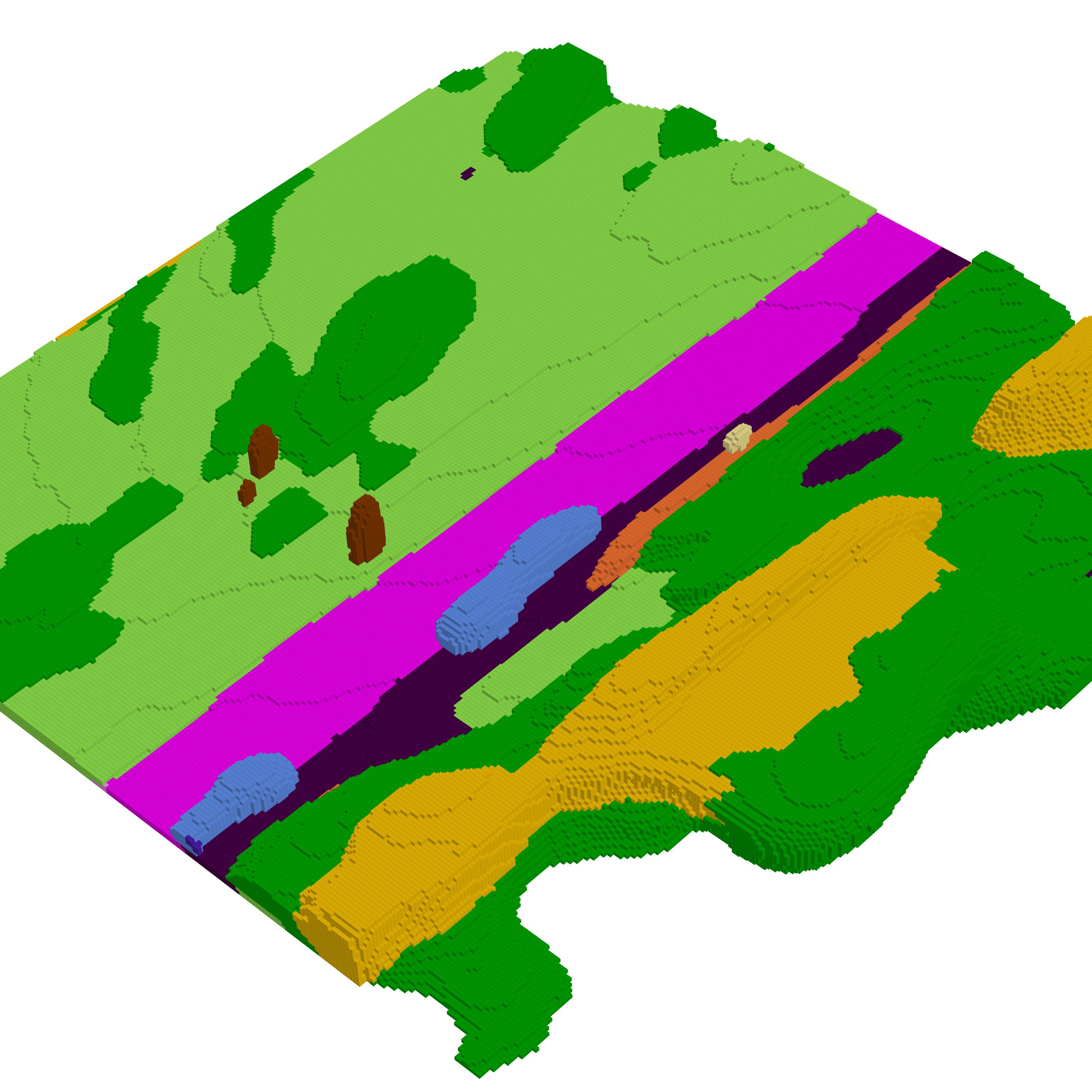} &
    \includegraphics[width=0.15\textwidth]{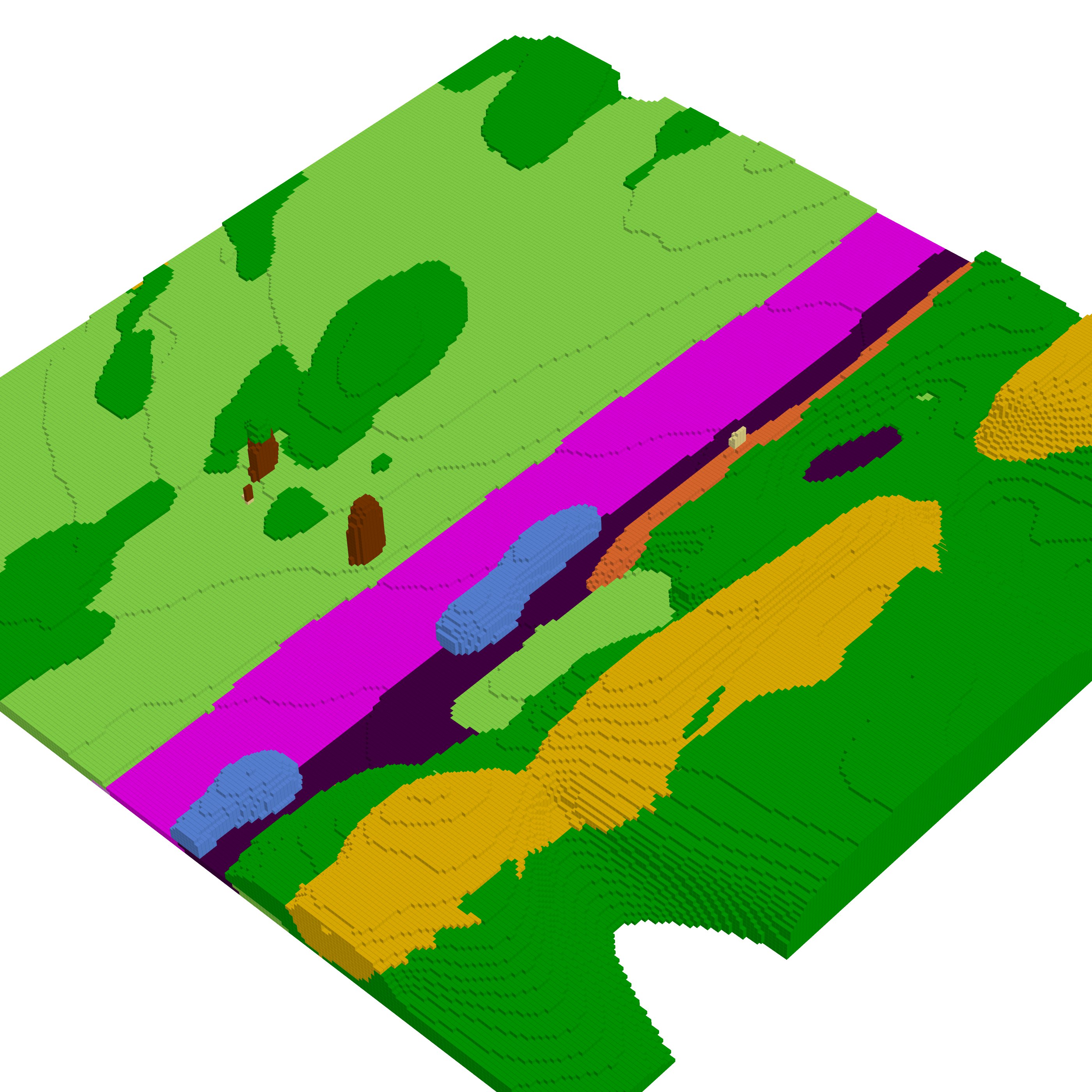} \\
    [0.3em]

    % ---- Row 3 ----
    \rgbpic{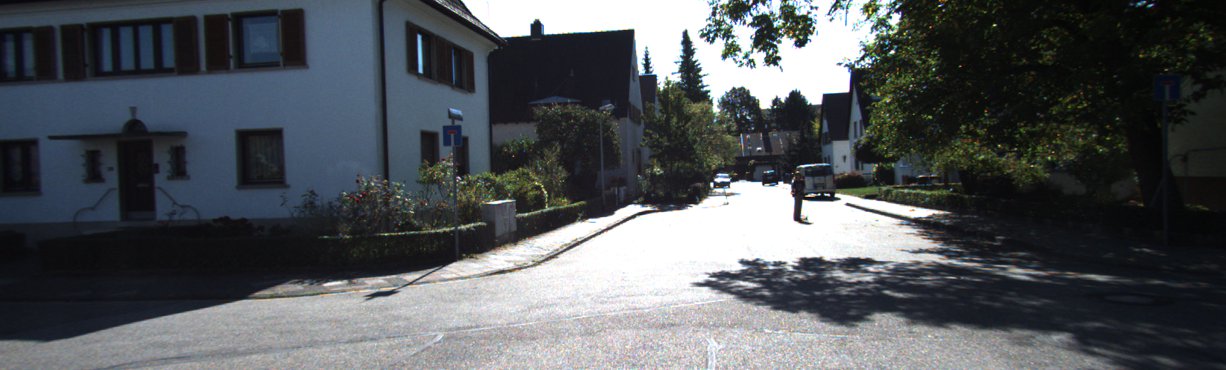} &
    \includegraphics[width=0.15\textwidth]{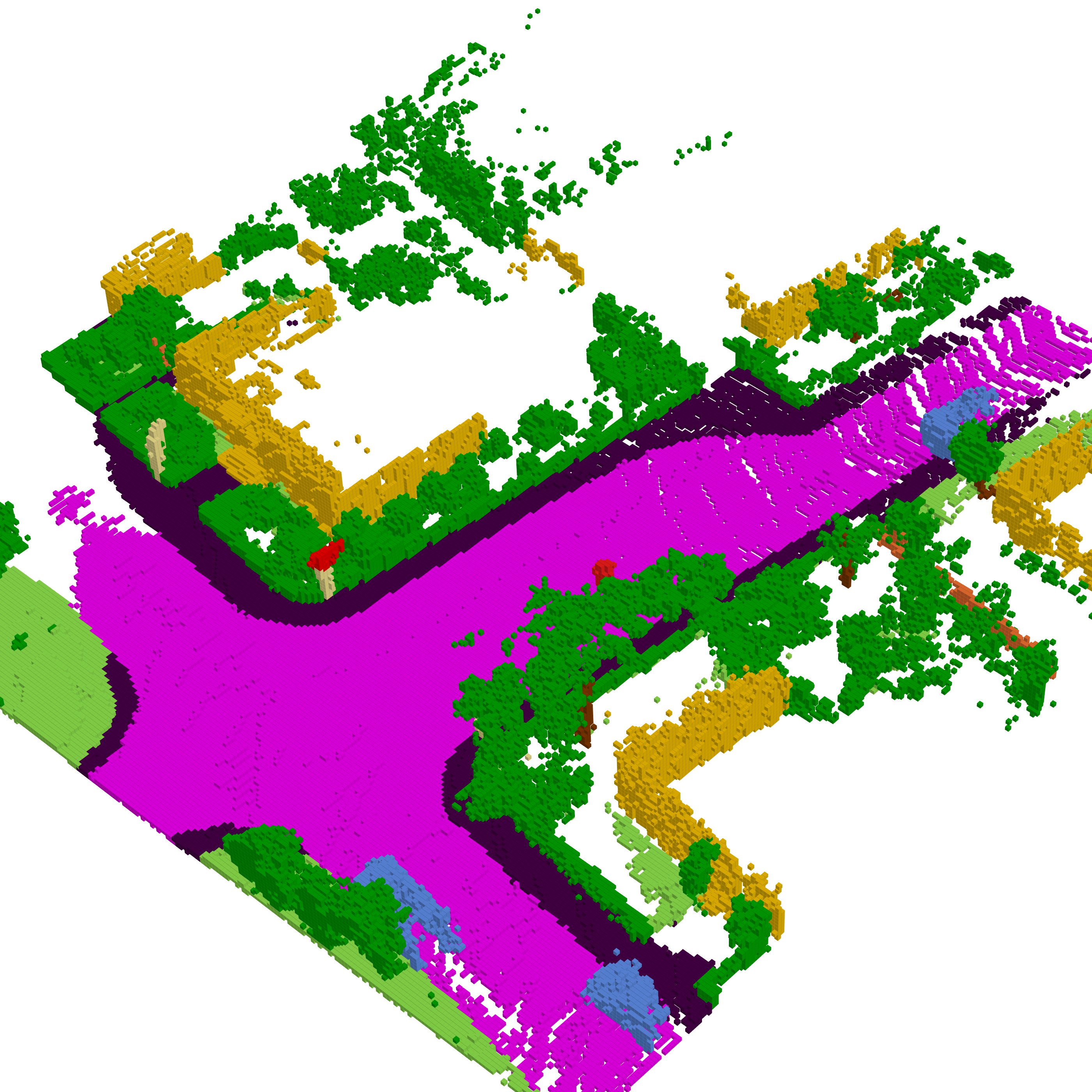} &
    \includegraphics[width=0.15\textwidth]{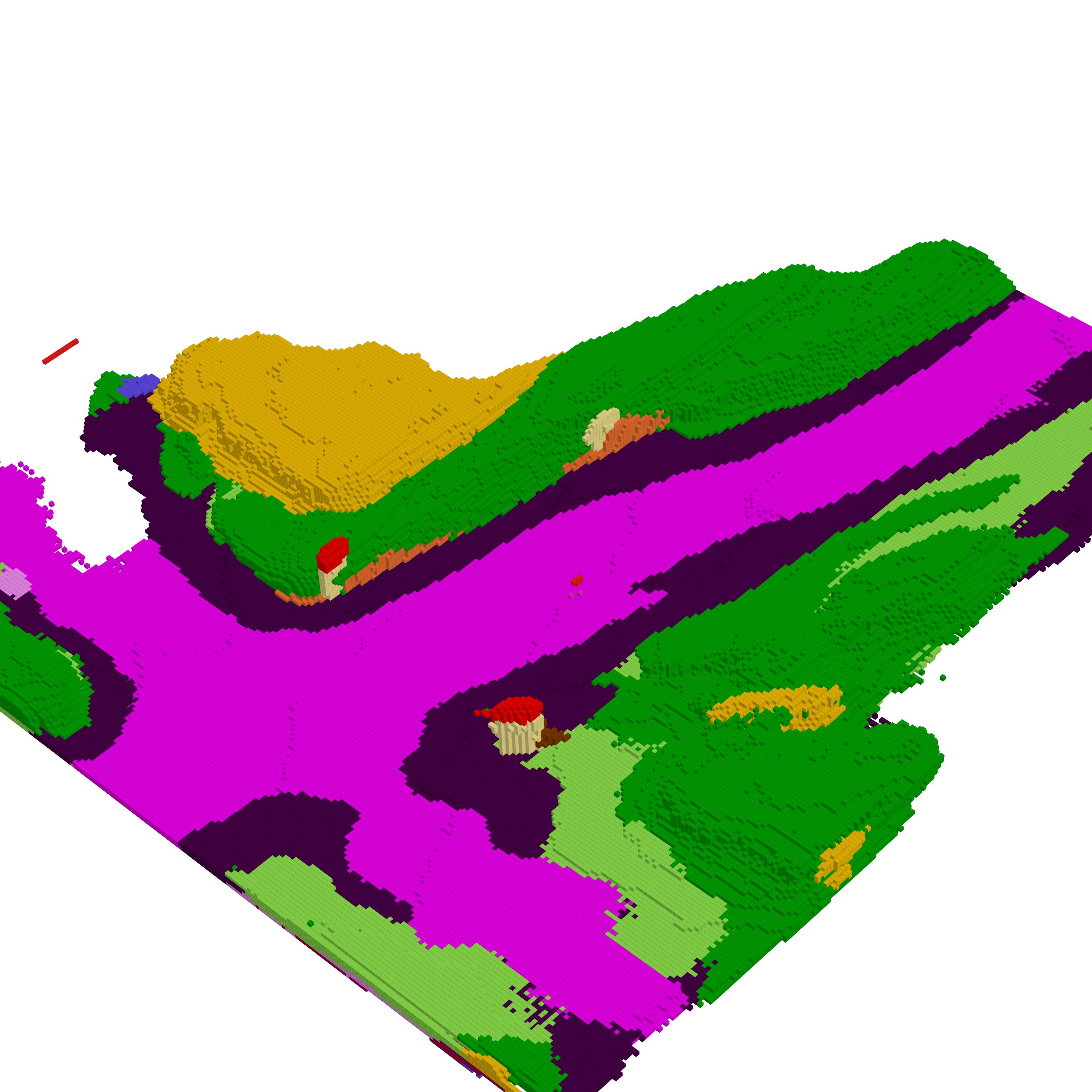} &
    \includegraphics[width=0.15\textwidth]{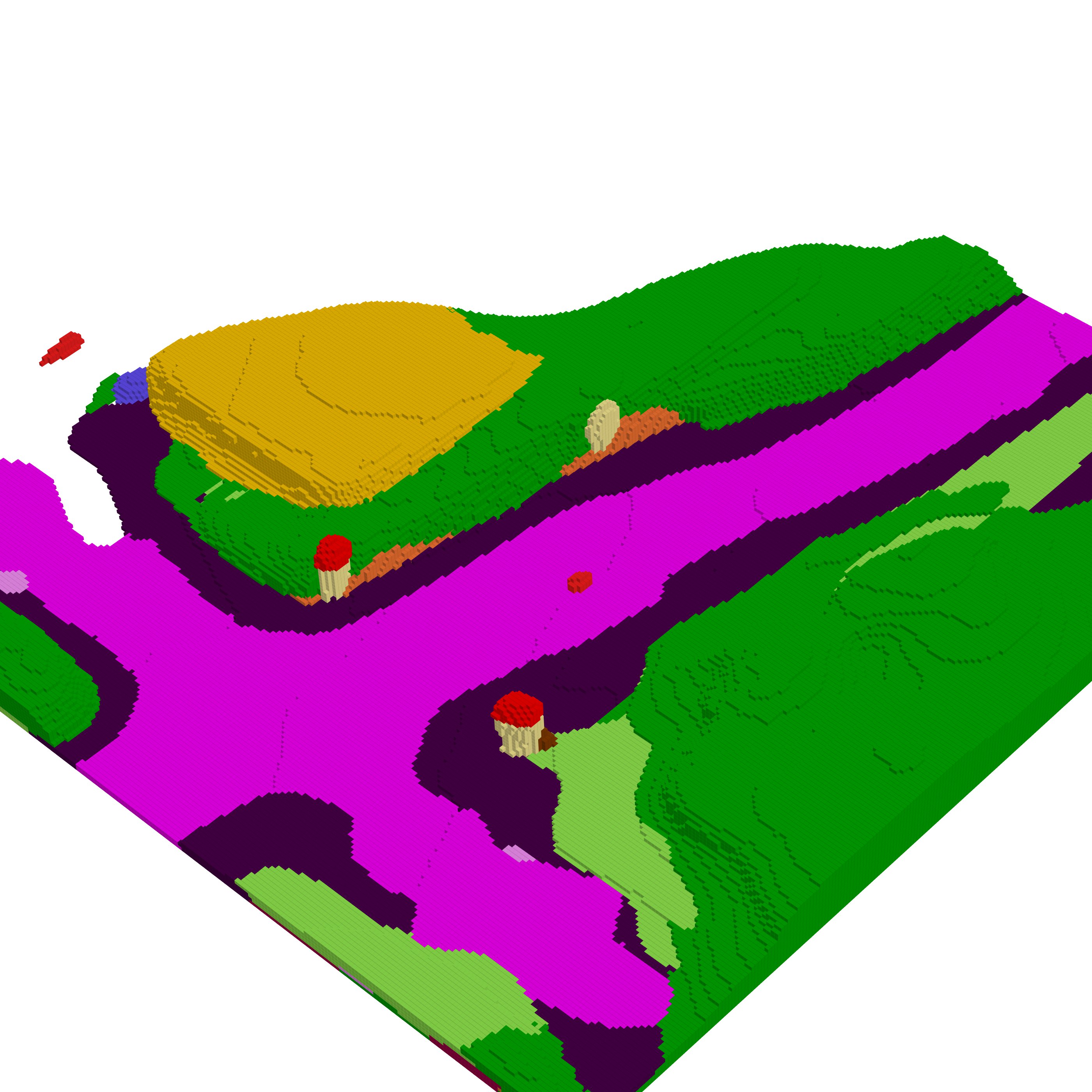} &
    \includegraphics[width=0.15\textwidth]{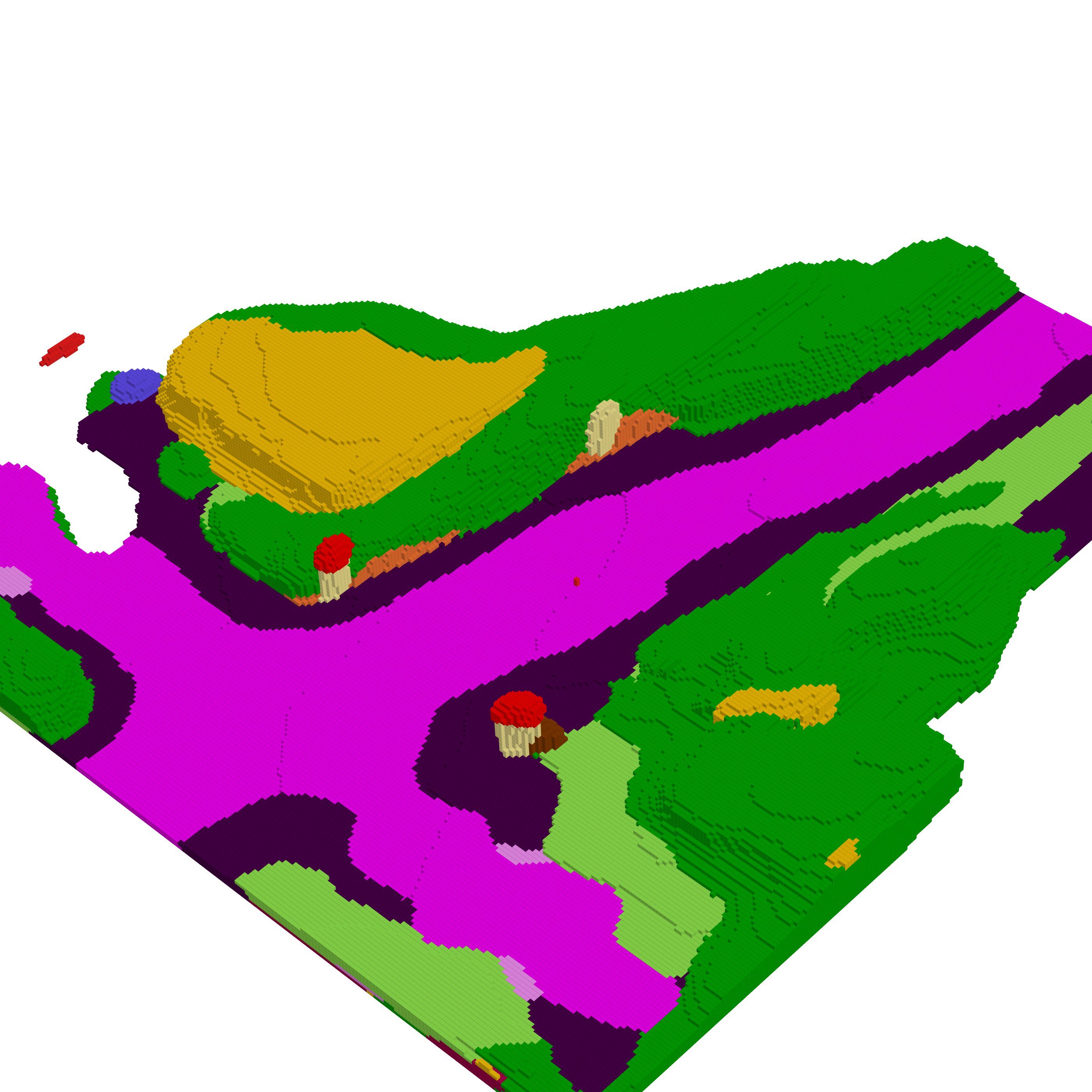} &
    \includegraphics[width=0.15\textwidth]{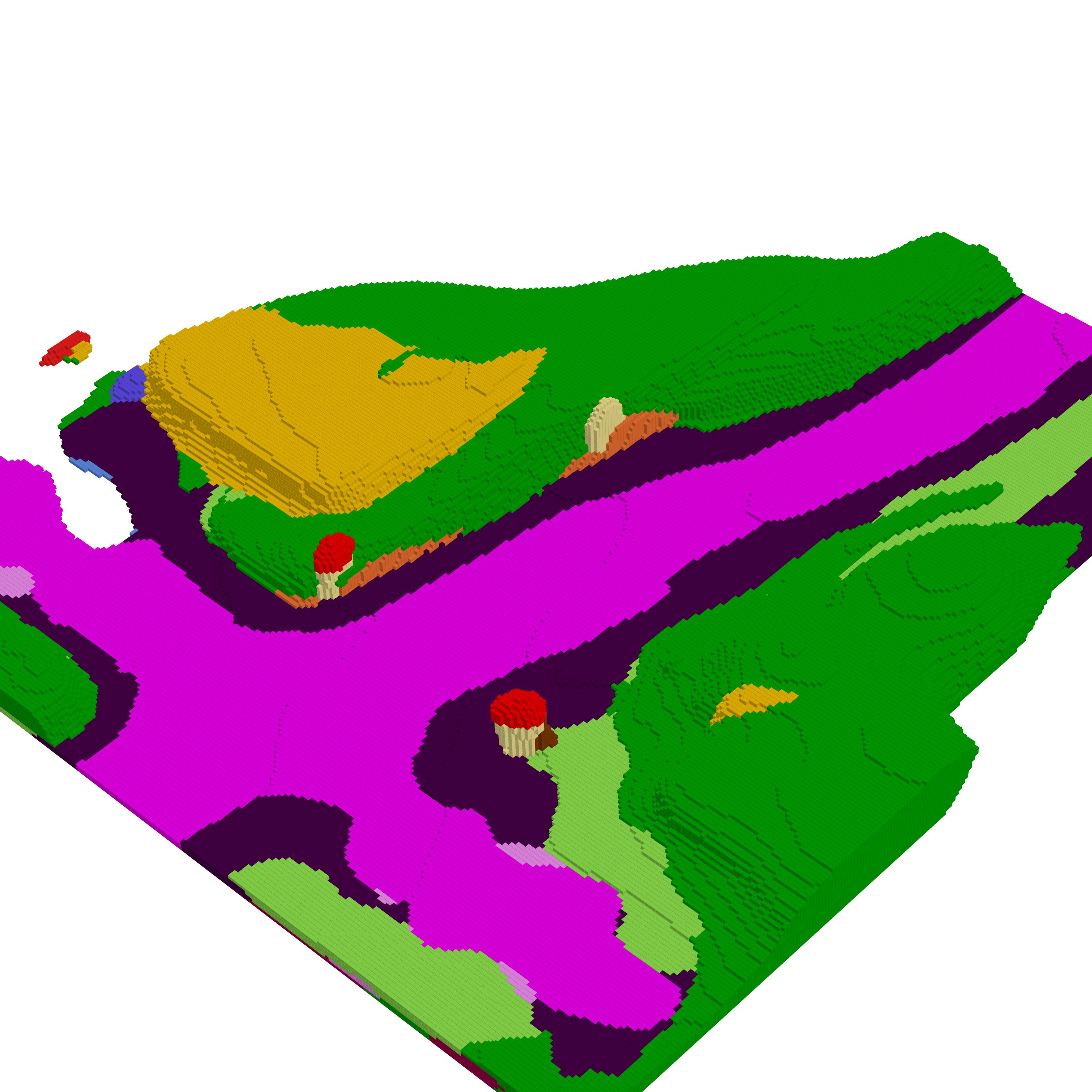} \\
    [0.4em]

    % ---- Legend row ----
    {\small RGB} &
    {\small Ground Truth} &
    {\small MonoScene} &
    {\small + 3D U\mbox{-}Net} &
    {\small + PNAM} &
    {\small + VLGM}
  \end{tabular}

  \caption{Qualitative results of ESSC-RM on MonoScene~\cite{2021_MonoScene}
  on the SemanticKITTI~\cite{2019_SemanticKITTI,2021_SemanticKITTI_2} validation set.
  ESSC-RM enhances the baseline prediction by completing missing structures,
  improving geometric consistency, and sharpening small semantic objects.}
  \label{fig:qualitative_MonoScene}
\end{figure}

\subsubsection{Qualitative Results}\label{sec:qual_results}

Figures~\ref{fig:qualitative_CGFormer} and \ref{fig:qualitative_MonoScene} present qualitative results of ESSC-RM applied respectively to CGFormer and MonoScene on the SemanticKITTI validation set.
Each row displays the input RGB image, ground truth, the prediction of the baseline model, and the refined outputs after integrating the 3D U-Net, PNAM, and VLGM modules.

Across both baselines, the refinement module consistently reduces holes and misclassifications in occluded or boundary regions, restores missing vegetation and structures at scene edges, and produces smoother and more coherent semantic layouts.
On large-scale structures such as roads and buildings, PNAM and VLGM further improve geometric regularity, yielding cleaner contours and more stable surface predictions.
For small-scale objects like traffic signs and poles, text-derived priors in VLGM highlight distinctive semantic regions, while PNAM enhances local aggregation and sharpens object boundaries.

These results demonstrate that ESSC-RM provides robust and generalizable refinement across different SSC backbones.

\section{Conclusion}
\label{sec:conclusion}
In summary, while ESSC-RM improves semantic scene completion by refining voxel predictions with PNAM and VLGM, several challenges remain.
First, the refinement module still relies on 3D convolutions and attention, leading to non-negligible latency and memory overhead.
Second, our evaluation is currently centered on SemanticKITTI, and broader generalization remains constrained by differences in voxel resolution, label taxonomy, and scene layout, which may require re-training or lightweight structural adaptation.
Moreover, while we validate the plug-and-play behavior on two representative SSC backbones (CGFormer and MonoScene), extending the plug-in evaluation to additional backbones is still limited by the training and storage overhead of voxel-level refinement. % [R1-2, R2-2]
Third, PNAM and VLGM are incorporated as largely independent components without a unified fusion mechanism, and emphasizing semantic correction may slightly compromise geometric completeness, resulting in minor degradations in SC-IoU. % [R2-1]

Future work will therefore explore lightweight and efficient representations (e.g., sparse convolution, tri-plane features, and Gaussian voxelization), knowledge distillation for compact deployment, as well as structured pruning and quantization-aware optimization to further reduce latency and memory footprint. % [R1-5]
We will also investigate adapter-based transfer across datasets, and broaden plug-in evaluation across diverse SSC backbones to further substantiate generality. % [R1-2, R2-3]
In addition, we will study adaptive fusion layers that more tightly couple local geometric attention with textual priors.
Finally, integrating generative priors (e.g., CVAE- or diffusion-based models) to pre-complete sparse voxels, together with extensive evaluation on diverse real-world benchmarks, may further improve the robustness, practicality, and scalability of ESSC-RM.

\section{Acknowledgments}
This research was funded by the Natural Science Foundation of Tianjin (No. 24PTLYHZ00290).

\bibliographystyle{Frontiers-Harvard} %  Many Frontiers journals use the Harvard referencing system (Author-date), to find the style and resources for the journal you are submitting to: https://zendesk.frontiersin.org/hc/en-us/articles/360017860337-Frontiers-Reference-Styles-by-Journal. For Humanities and Social Sciences articles please include page numbers in the in-text citations 
\bibliography{test}

@ARTICLE{9787342,
  author={Cao, Hu and Chen, Guang and Li, Zhijun and Hu, Yingbai and Knoll, Alois},
  journal={IEEE Transactions on Instrumentation and Measurement}, 
  title={NeuroGrasp: Multimodal Neural Network With Euler Region Regression for Neuromorphic Vision-Based Grasp Pose Estimation}, 
  year={2022},
  volume={71},
  number={},
  pages={1-11},
  doi={10.1109/TIM.2022.3179469}}

@ARTICLE{9546775,
  author={Cao, Hu and Chen, Guang and Xia, Jiahao and Zhuang, Genghang and Knoll, Alois},
  journal={IEEE Sensors Journal}, 
  title={Fusion-Based Feature Attention Gate Component for Vehicle Detection Based on Event Camera}, 
  year={2021},
  volume={21},
  number={21},
  pages={24540-24548},
  doi={10.1109/JSEN.2021.3115016}}

@inproceedings{cao2024embracing,
  title={Embracing events and frames with hierarchical feature refinement network for object detection},
  author={Cao, Hu and Zhang, Zehua and Xia, Yan and Li, Xinyi and Xia, Jiahao and Chen, Guang and Knoll, Alois},
  booktitle={European Conference on Computer Vision},
  pages={161--177},
  year={2024},
  organization={Springer}
}

@ARTICLE{10292927,
  author={Cao, Hu and Qu, Zhongnan and Chen, Guang and Li, Xinyi and Thiele, Lothar and Knoll, Alois},
  journal={IEEE Transactions on Artificial Intelligence}, 
  title={GhostViT: Expediting Vision Transformers via Cheap Operations}, 
  year={2024},
  volume={5},
  number={6},
  pages={2517-2525},
  doi={10.1109/TAI.2023.3326795}}

@ARTICLE{10584449,
  author={Cao, Hu and Chen, Guang and Zhao, Hengshuang and Jiang, Dongsheng and Zhang, Xiaopeng and Tian, Qi and Knoll, Alois},
  journal={IEEE Transactions on Intelligent Transportation Systems}, 
  title={SDPT: Semantic-Aware Dimension-Pooling Transformer for Image Segmentation}, 
  year={2024},
  volume={25},
  number={11},
  pages={15934-15946},
  doi={10.1109/TITS.2024.3417813}}

@article{2019_DL43DPC,
  author       = {Yulan Guo and
                  Hanyun Wang and
                  Qingyong Hu and
                  Hao Liu and
                  Li Liu and
                  Mohammed Bennamoun},
  title        = {Deep Learning for 3D Point Clouds: {A} Survey},
  journal      = {CoRR},
  volume       = {abs/1912.12033},
  year         = {2019},
  url          = {http://arxiv.org/abs/1912.12033},
  eprinttype    = {arXiv},
  eprint       = {1912.12033},
  bibsource    = {dblp computer science bibliography, https://dblp.org}
}

@article{2021_3DSSC,
  author       = {Luis Roldao and
                  Raoul de Charette and
                  Anne Verroust{-}Blondet},
  title        = {3D Semantic Scene Completion: a Survey},
  journal      = {CoRR},
  volume       = {abs/2103.07466},
  year         = {2021},
  url          = {https://arxiv.org/abs/2103.07466},
  eprinttype    = {arXiv},
  eprint       = {2103.07466},
  bibsource    = {dblp computer science bibliography, https://dblp.org}
}

@INPROCEEDINGS{2024_SADP,
  author={Zhao, Hui and Li, Xin and Xu, Cheng and Xu, Bingxin and Liu, Hongzhe},
  booktitle={2024 IEEE 24th International Conference on Software Quality, Reliability, and Security Companion (QRS-C)}, 
  title={A Survey of Automatic Driving Environment Perception}, 
  year={2024},
  volume={},
  number={},
  pages={1038-1047},
  doi={10.1109/QRS-C63300.2024.00137}
}

@ARTICLE{2020_SAD,
  author={Yurtsever, Ekim and Lambert, Jacob and Carballo, Alexander and Takeda, Kazuya},
  journal={IEEE Access}, 
  title={A Survey of Autonomous Driving: Common Practices and Emerging Technologies}, 
  year={2020},
  volume={8},
  number={},
  pages={58443-58469},
  doi={10.1109/ACCESS.2020.2983149}}

@article{2022_3DOD4AD,
  author       = {Xinzhu Ma and
                  Wanli Ouyang and
                  Andrea Simonelli and
                  Elisa Ricci},
  title        = {3D Object Detection from Images for Autonomous Driving: {A} Survey},
  journal      = {CoRR},
  volume       = {abs/2202.02980},
  year         = {2022},
  url          = {https://arxiv.org/abs/2202.02980},
  eprinttype    = {arXiv},
  eprint       = {2202.02980},
  bibsource    = {dblp computer science bibliography, https://dblp.org}
}

@article{2016_SSCNet,
  author       = {Shuran Song and
                  Fisher Yu and
                  Andy Zeng and
                  Angel X. Chang and
                  Manolis Savva and
                  Thomas A. Funkhouser},
  title        = {Semantic Scene Completion from a Single Depth Image},
  journal      = {CoRR},
  volume       = {abs/1611.08974},
  year         = {2016},
  url          = {http://arxiv.org/abs/1611.08974},
  eprinttype    = {arXiv},
  eprint       = {1611.08974},
  bibsource    = {dblp computer science bibliography, https://dblp.org}
}

@article{2020_LMSCNet,
  author       = {Luis Rold{\~{a}}o and
                  Raoul de Charette and
                  Anne Verroust{-}Blondet},
  title        = {LMSCNet: Lightweight Multiscale 3D Semantic Completion},
  journal      = {CoRR},
  volume       = {abs/2008.10559},
  year         = {2020},
  url          = {https://arxiv.org/abs/2008.10559},
  eprinttype    = {arXiv},
  eprint       = {2008.10559},
  bibsource    = {dblp computer science bibliography, https://dblp.org}
}

@article{2020_JS3CNet,
  author       = {Xu Yan and
                  Jiantao Gao and
                  Jie Li and
                  Ruimao Zhang and
                  Zhen Li and
                  Rui Huang and
                  Shuguang Cui},
  title        = {Sparse Single Sweep LiDAR Point Cloud Segmentation via Learning Contextual
                  Shape Priors from Scene Completion},
  journal      = {CoRR},
  volume       = {abs/2012.03762},
  year         = {2020},
  url          = {https://arxiv.org/abs/2012.03762},
  eprinttype    = {arXiv},
  eprint       = {2012.03762},
  bibsource    = {dblp computer science bibliography, https://dblp.org}
}

@article{2023_SCPNet,
  title={SCPNet: Semantic Scene Completion on Point Cloud},
  author={Zhaoyang Xia and You-Chen Liu and Xin Li and Xinge Zhu and Yuexin Ma and Yikang Li and Yuenan Hou and Y. Qiao},
  journal={2023 IEEE/CVF Conference on Computer Vision and Pattern Recognition (CVPR)},
  year={2023},
  pages={17642-17651},
  url={https://api.semanticscholar.org/CorpusID:257496877}
}

@inproceedings{2024_TALoS,
 author = {Jang, Hyun-Kurl and Kim, Jihun and Kweon, Hyeokjun and Yoon, Kuk-Jin},
 booktitle = {Advances in Neural Information Processing Systems},
 doi = {10.52202/079017-2361},
 editor = {A. Globerson and L. Mackey and D. Belgrave and A. Fan and U. Paquet and J. Tomczak and C. Zhang},
 pages = {74211--74232},
 publisher = {Curran Associates, Inc.},
 title = {TALoS: Enhancing Semantic Scene Completion via Test-time Adaptation on the Line of Sight},
 url = {https://proceedings.neurips.cc/paper_files/paper/2024/file/87571720167f7e88827c40e468e3101f-Paper-Conference.pdf},
 volume = {37},
 year = {2024}
}

@article{2021_MonoScene,
  author       = {Anh{-}Quan Cao and
                  Raoul de Charette},
  title        = {MonoScene: Monocular 3D Semantic Scene Completion},
  journal      = {CoRR},
  volume       = {abs/2112.00726},
  year         = {2021},
  url          = {https://arxiv.org/abs/2112.00726},
  eprinttype    = {arXiv},
  eprint       = {2112.00726},
  bibsource    = {dblp computer science bibliography, https://dblp.org}
}

@InProceedings{2023_VoxFormer,
      title={VoxFormer: Sparse Voxel Transformer for Camera-based 3D Semantic Scene Completion}, 
      author={Li, Yiming and Yu, Zhiding and Choy, Christopher and Xiao, Chaowei and Alvarez, Jose M and Fidler, Sanja and Feng, Chen and Anandkumar, Anima},
      booktitle = {Proceedings of the IEEE/CVF Conference on Computer Vision and Pattern Recognition (CVPR)},
      year={2023},
        eprint={2302.12251},
      archivePrefix={arXiv},
      primaryClass={cs.CV},
      url={https://arxiv.org/abs/2302.12251},
}

@article{2023_OccFormer,
  title={OccFormer: Dual-path Transformer for Vision-based 3D Semantic Occupancy Prediction},
  author={Yunpeng Zhang and Zhengbiao Zhu and Dalong Du},
  journal={2023 IEEE/CVF International Conference on Computer Vision (ICCV)},
  year={2023},
  pages={9399-9409},
  url={https://api.semanticscholar.org/CorpusID:258059891}
}

@article{2024_DepthSSC,
  title={DepthSSC: Monocular 3D Semantic Scene Completion via Depth-Spatial Alignment and Voxel Adaptation},
  author={Jiawei Yao and Jusheng Zhang and Xiaochao Pan and Tong Wu and Canran Xiao},
  journal={2025 IEEE/CVF Winter Conference on Applications of Computer Vision (WACV)},
  year={2023},
  pages={2154-2163},
  url={https://api.semanticscholar.org/CorpusID:265498560}
}

@ARTICLE{2024_IAMSSC,
  author={Xiao, Haihong and Xu, Hongbin and Kang, Wenxiong and Li, Yuqiong},
  journal={IEEE Transactions on Intelligent Transportation Systems}, 
  title={Instance-Aware Monocular 3D Semantic Scene Completion}, 
  year={2024},
  volume={25},
  number={7},
  pages={6543-6554},
  doi={10.1109/TITS.2023.3344806}}

@INPROCEEDINGS{2023_Symphonies,
  author={Jiang, Haoyi and Cheng, Tianheng and Gao, Naiyu and Zhang, Haoyang and Lin, Tianwei and Liu, Wenyu and Wang, Xinggang},
  booktitle={2024 IEEE/CVF Conference on Computer Vision and Pattern Recognition (CVPR)}, 
  title={Symphonize 3D Semantic Scene Completion with Contextual Instance Queries}, 
  year={2024},
  volume={},
  number={},
  pages={20258-20267},
  doi={10.1109/CVPR52733.2024.01915}}

@inproceedings{2023_CGFormer,
 title = {Context and Geometry Aware Voxel Transformer for Semantic Scene Completion},
 author = {Yu, Zhu and Zhang, Runmin and Ying, Jiacheng and Yu, Junchen and Hu, Xiaohai and Luo, Lun and Cao, Si-Yuan and Shen, Hui-liang},
 booktitle = {Advances in Neural Information Processing Systems},
 pages = {1531--1555},
 volume = {37},
 year = {2024}
}

@article{2019_M3DRPN,
  author       = {Garrick Brazil and
                  Xiaoming Liu},
  title        = {{M3D-RPN:} Monocular 3D Region Proposal Network for Object Detection},
  journal      = {CoRR},
  volume       = {abs/1907.06038},
  year         = {2019},
  url          = {http://arxiv.org/abs/1907.06038},
  eprinttype    = {arXiv},
  eprint       = {1907.06038},
  bibsource    = {dblp computer science bibliography, https://dblp.org}
}

@article{2019_CenterNet,
  author       = {Kaiwen Duan and
                  Song Bai and
                  Lingxi Xie and
                  Honggang Qi and
                  Qingming Huang and
                  Qi Tian},
  title        = {CenterNet: Keypoint Triplets for Object Detection},
  journal      = {CoRR},
  volume       = {abs/1904.08189},
  year         = {2019},
  url          = {http://arxiv.org/abs/1904.08189},
  eprinttype    = {arXiv},
  eprint       = {1904.08189},
  bibsource    = {dblp computer science bibliography, https://dblp.org}
}

@article{2018_ROI_10D,
  author       = {Fabian Manhardt and
                  Wadim Kehl and
                  Adrien Gaidon},
  title        = {{ROI-10D:} Monocular Lifting of 2D Detection to 6D Pose and Metric
                  Shape},
  journal      = {CoRR},
  volume       = {abs/1812.02781},
  year         = {2018},
  url          = {http://arxiv.org/abs/1812.02781},
  eprinttype    = {arXiv},
  eprint       = {1812.02781},
  bibsource    = {dblp computer science bibliography, https://dblp.org}
}

@INPROCEEDINGS{2018_MultiFusion,
  author={Xu, Bin and Chen, Zhenzhong},
  booktitle={2018 IEEE/CVF Conference on Computer Vision and Pattern Recognition}, 
  title={Multi-level Fusion Based 3D Object Detection from Monocular Images}, 
  year={2018},
  volume={},
  number={},
  pages={2345-2353},
  doi={10.1109/CVPR.2018.00249}}

@article{2018_Pseudo_LiDAR,
author       = {Yan Wang and
                Wei{-}Lun Chao and
                Divyansh Garg and
                Bharath Hariharan and
                Mark Campbell and
                Kilian Q. Weinberger},
title        = {Pseudo-LiDAR from Visual Depth Estimation: Bridging the Gap in 3D
                Object Detection for Autonomous Driving},
journal      = {CoRR},
volume       = {abs/1812.07179},
year         = {2018},
url          = {http://arxiv.org/abs/1812.07179},
eprinttype    = {arXiv},
eprint       = {1812.07179},
bibsource    = {dblp computer science bibliography, https://dblp.org}
}

@article{2016_3DBB,
  author       = {Arsalan Mousavian and
                  Dragomir Anguelov and
                  John Flynn and
                  Jana Kosecka},
  title        = {3D Bounding Box Estimation Using Deep Learning and Geometry},
  journal      = {CoRR},
  volume       = {abs/1612.00496},
  year         = {2016},
  url          = {http://arxiv.org/abs/1612.00496},
  eprinttype    = {arXiv},
  eprint       = {1612.00496},
  bibsource    = {dblp computer science bibliography, https://dblp.org}
}

@article{2018_JM3D,
  author       = {Hou{-}Ning Hu and
                  Qi{-}Zhi Cai and
                  Dequan Wang and
                  Ji Lin and
                  Min Sun and
                  Philipp Kr{\"{a}}henb{\"{u}}hl and
                  Trevor Darrell and
                  Fisher Yu},
  title        = {Joint Monocular 3D Vehicle Detection and Tracking},
  journal      = {CoRR},
  volume       = {abs/1811.10742},
  year         = {2018},
  url          = {http://arxiv.org/abs/1811.10742},
  eprinttype    = {arXiv},
  eprint       = {1811.10742},
  bibsource    = {dblp computer science bibliography, https://dblp.org}
}

@INPROCEEDINGS{2014_3DB,
  author={Zia, Muhammad Zeeshan and Stark, Michael and Schindler, Konrad},
  booktitle={2014 IEEE Conference on Computer Vision and Pattern Recognition}, 
  title={Are Cars Just 3D Boxes? Jointly Estimating the 3D Shape of Multiple Objects}, 
  year={2014},
  volume={},
  number={},
  pages={3678-3685},
  doi={10.1109/CVPR.2014.470}}

@article{2018_PyramidStero,
  author       = {Jia{-}Ren Chang and
                  Yong{-}Sheng Chen},
  title        = {Pyramid Stereo Matching Network},
  journal      = {CoRR},
  volume       = {abs/1803.08669},
  year         = {2018},
  url          = {http://arxiv.org/abs/1803.08669},
  eprinttype    = {arXiv},
  eprint       = {1803.08669},
  bibsource    = {dblp computer science bibliography, https://dblp.org}
}

@article{2019_SteroRCNN,
  author       = {Peiliang Li and
                  Xiaozhi Chen and
                  Shaojie Shen},
  title        = {Stereo {R-CNN} based 3D Object Detection for Autonomous Driving},
  journal      = {CoRR},
  volume       = {abs/1902.09738},
  year         = {2019},
  url          = {http://arxiv.org/abs/1902.09738},
  eprinttype    = {arXiv},
  eprint       = {1902.09738},
  bibsource    = {dblp computer science bibliography, https://dblp.org}
}

@article{2019_Pseudo_LiDAR,
  author       = {Yurong You and
                  Yan Wang and
                  Wei{-}Lun Chao and
                  Divyansh Garg and
                  Geoff Pleiss and
                  Bharath Hariharan and
                  Mark Campbell and
                  Kilian Q. Weinberger},
  title        = {Pseudo-LiDAR++: Accurate Depth for 3D Object Detection in Autonomous
                  Driving},
  journal      = {CoRR},
  volume       = {abs/1906.06310},
  year         = {2019},
  url          = {http://arxiv.org/abs/1906.06310},
  eprinttype    = {arXiv},
  eprint       = {1906.06310},
  bibsource    = {dblp computer science bibliography, https://dblp.org}
}

@article{2020_DSGN,
  author       = {Yilun Chen and
                  Shu Liu and
                  Xiaoyong Shen and
                  Jiaya Jia},
  title        = {{DSGN:} Deep Stereo Geometry Network for 3D Object Detection},
  journal      = {CoRR},
  volume       = {abs/2001.03398},
  year         = {2020},
  url          = {https://arxiv.org/abs/2001.03398},
  eprinttype    = {arXiv},
  eprint       = {2001.03398},
  bibsource    = {dblp computer science bibliography, https://dblp.org}
}

@article{2023_3DOD,
  title   = {3D Object Detection for Autonomous Driving: A Comprehensive Survey},
  author  = {Mao, Jun and Shi, Shaoshuai and Wang, Xiaogang and others},
  journal = {International Journal of Computer Vision},
  volume  = {131},
  pages   = {1909--1963},
  year    = {2023},
  doi     = {10.1007/s11263-023-01790-1},
  url     = {https://doi.org/10.1007/s11263-023-01790-1}
}

@article{2020_LSS,
  author       = {Jonah Philion and
                  Sanja Fidler},
  title        = {Lift, Splat, Shoot: Encoding Images From Arbitrary Camera Rigs by
                  Implicitly Unprojecting to 3D},
  journal      = {CoRR},
  volume       = {abs/2008.05711},
  year         = {2020},
  url          = {https://arxiv.org/abs/2008.05711},
  eprinttype    = {arXiv},
  eprint       = {2008.05711},
  bibsource    = {dblp computer science bibliography, https://dblp.org}
}

@article{2021_BEVDet,
  author       = {Junjie Huang and
                  Guan Huang and
                  Zheng Zhu and
                  Dalong Du},
  title        = {BEVDet: High-performance Multi-camera 3D Object Detection in Bird-Eye-View},
  journal      = {CoRR},
  volume       = {abs/2112.11790},
  year         = {2021},
  url          = {https://arxiv.org/abs/2112.11790},
  eprinttype    = {arXiv},
  eprint       = {2112.11790},
  bibsource    = {dblp computer science bibliography, https://dblp.org}
}

@misc{2022_BEVFormer,
      title={BEVFormer: Learning Bird's-Eye-View Representation from Multi-Camera Images via Spatiotemporal Transformers}, 
      author={Zhiqi Li and Wenhai Wang and Hongyang Li and Enze Xie and Chonghao Sima and Tong Lu and Qiao Yu and Jifeng Dai},
      year={2022},
      eprint={2203.17270},
      archivePrefix={arXiv},
      primaryClass={cs.CV},
      url={https://arxiv.org/abs/2203.17270}, 
}

@article{2021_DETR3D,
  author       = {Yue Wang and
                  Vitor Guizilini and
                  Tianyuan Zhang and
                  Yilun Wang and
                  Hang Zhao and
                  Justin Solomon},
  title        = {{DETR3D:} 3D Object Detection from Multi-view Images via 3D-to-2D
                  Queries},
  journal      = {CoRR},
  volume       = {abs/2110.06922},
  year         = {2021},
  url          = {https://arxiv.org/abs/2110.06922},
  eprinttype    = {arXiv},
  eprint       = {2110.06922},
  bibsource    = {dblp computer science bibliography, https://dblp.org}
}

@article{2017_AIAYN,
  author       = {Ashish Vaswani and
                  Noam Shazeer and
                  Niki Parmar and
                  Jakob Uszkoreit and
                  Llion Jones and
                  Aidan N. Gomez and
                  Lukasz Kaiser and
                  Illia Polosukhin},
  title        = {Attention Is All You Need},
  journal      = {CoRR},
  volume       = {abs/1706.03762},
  year         = {2017},
  url          = {http://arxiv.org/abs/1706.03762},
  eprinttype    = {arXiv},
  eprint       = {1706.03762},
  bibsource    = {dblp computer science bibliography, https://dblp.org}
}

@inproceedings{2022_SDETR,
  author = {Doll, Simon and Schulz, Richard and Schneider, Lukas and Benzin, Viviane and Enzweiler Markus and Lensch, Hendrik P.A.},
  title = {SpatialDETR: Robust Scalable Transformer-Based 3D Object Detection from Multi-View Camera Images with Global Cross-Sensor Attention},
  booktitle = {European Conference on Computer Vision(ECCV)},
  year = {2022}
}

@misc{2023_TPVFormer,
      title={Tri-Perspective View for Vision-Based 3D Semantic Occupancy Prediction}, 
      author={Yuanhui Huang and Wenzhao Zheng and Yunpeng Zhang and Jie Zhou and Jiwen Lu},
      year={2023},
      eprint={2302.07817},
      archivePrefix={arXiv},
      primaryClass={cs.CV},
      url={https://arxiv.org/abs/2302.07817}, 
}

@article{2017_VoxelNet,
  author       = {Yin Zhou and
                  Oncel Tuzel},
  title        = {VoxelNet: End-to-End Learning for Point Cloud Based 3D Object Detection},
  journal      = {CoRR},
  volume       = {abs/1711.06396},
  year         = {2017},
  url          = {http://arxiv.org/abs/1711.06396},
  eprinttype    = {arXiv},
  eprint       = {1711.06396},
  bibsource    = {dblp computer science bibliography, https://dblp.org}
}

@article{2015_2DUNET,
  author       = {Olaf Ronneberger and
                  Philipp Fischer and
                  Thomas Brox},
  title        = {U-Net: Convolutional Networks for Biomedical Image Segmentation},
  journal      = {CoRR},
  volume       = {abs/1505.04597},
  year         = {2015},
  url          = {http://arxiv.org/abs/1505.04597},
  eprinttype    = {arXiv},
  eprint       = {1505.04597},
  bibsource    = {dblp computer science bibliography, https://dblp.org}
}

@article{2016_3DUNET,
  author       = {{\"{O}}zg{\"{u}}n {\c{C}}i{\c{c}}ek and
                  Ahmed Abdulkadir and
                  Soeren S. Lienkamp and
                  Thomas Brox and
                  Olaf Ronneberger},
  title        = {3D U-Net: Learning Dense Volumetric Segmentation from Sparse Annotation},
  journal      = {CoRR},
  volume       = {abs/1606.06650},
  year         = {2016},
  url          = {http://arxiv.org/abs/1606.06650},
  eprinttype    = {arXiv},
  eprint       = {1606.06650},
  bibsource    = {dblp computer science bibliography, https://dblp.org}
}

@article{2019_SemanticKITTI,
  author       = {Jens Behley and
                  Martin Garbade and
                  Andres Milioto and
                  Jan Quenzel and
                  Sven Behnke and
                  Cyrill Stachniss and
                  Juergen Gall},
  title        = {A Dataset for Semantic Segmentation of Point Cloud Sequences},
  journal      = {CoRR},
  volume       = {abs/1904.01416},
  year         = {2019},
  url          = {http://arxiv.org/abs/1904.01416},
  eprinttype    = {arXiv},
  eprint       = {1904.01416},
  bibsource    = {dblp computer science bibliography, https://dblp.org}
}

@article{2021_SemanticKITTI_2,
author = {Jens Behley and Martin Garbade and Andres Milioto and Jan Quenzel and Sven Behnke and Jürgen Gall and Cyrill Stachniss},
title ={Towards 3D LiDAR-based semantic scene understanding of 3D point cloud sequences: The SemanticKITTI Dataset},
journal = {The International Journal of Robotics Research},
volume = {40},
number = {8-9},
pages = {959-967},
year = {2021},
doi = {10.1177/02783649211006735},
URL = {https://doi.org/10.1177/02783649211006735},
eprint = {  https://doi.org/10.1177/02783649211006735}
}

@INPROCEEDINGS{2012_KITTI,
  author={Geiger, Andreas and Lenz, Philip and Urtasun, Raquel},
  booktitle={2012 IEEE Conference on Computer Vision and Pattern Recognition}, 
  title={Are we ready for autonomous driving? The KITTI vision benchmark suite}, 
  year={2012},
  volume={},
  number={},
  pages={3354-3361},
  doi={10.1109/CVPR.2012.6248074}}

@misc{2024_HASSC,
      title={Not All Voxels Are Equal: Hardness-Aware Semantic Scene Completion with Self-Distillation}, 
      author={Song Wang and Jiawei Yu and Wentong Li and Wenyu Liu and Xiaolu Liu and Junbo Chen and Jianke Zhu},
      year={2024},
      eprint={2404.11958},
      archivePrefix={arXiv},
      primaryClass={cs.CV},
      url={https://arxiv.org/abs/2404.11958}, 
}

@article{2024_H2GFormer, 
title={H2GFormer: Horizontal-to-Global Voxel Transformer for 3D Semantic Scene Completion}, 
volume={38},
url={https://ojs.aaai.org/index.php/AAAI/article/view/28384}, 
DOI={10.1609/aaai.v38i6.28384}, 
number={6}, 
journal={Proceedings of the AAAI Conference on Artificial Intelligence}, 
author={Wang, Yu and Tong, Chao}, 
year={2024}, 
month={Mar.}, 
pages={5722-5730} }

@article{2017_AdamW,
  author       = {Ilya Loshchilov and
                  Frank Hutter},
  title        = {Fixing Weight Decay Regularization in Adam},
  journal      = {CoRR},
  volume       = {abs/1711.05101},
  year         = {2017},
  url          = {http://arxiv.org/abs/1711.05101},
  eprinttype    = {arXiv},
  eprint       = {1711.05101},
  bibsource    = {dblp computer science bibliography, https://dblp.org}
}

@article{2017_OneCycleLR,
  author       = {Leslie N. Smith and
                  Nicholay Topin},
  title        = {Super-Convergence: Very Fast Training of Residual Networks Using Large
                  Learning Rates},
  journal      = {CoRR},
  volume       = {abs/1708.07120},
  year         = {2017},
  url          = {http://arxiv.org/abs/1708.07120},
  eprinttype    = {arXiv},
  eprint       = {1708.07120},
  bibsource    = {dblp computer science bibliography, https://dblp.org}
}

@inproceedings{2024_SemCity,
    title={SemCity: Semantic Scene Generation with Triplane Diffusion},
    author={Lee, Jumin and Lee, Sebin and Jo, Changho and Im, Woobin and Seon, Juhyeong and Yoon, Sung-Eui},
    booktitle={Proceedings of the IEEE/CVF conference on computer vision and pattern recognition},
    year={2024},
    doi ={10.48550/arXiv.2403.07773}
}

@inproceedings{hassani2024faster,
 author = {Hassani, Ali and Hwu, Wen-mei and Shi, Humphrey},
 booktitle = {Advances in Neural Information Processing Systems},
 doi = {10.52202/079017-2065},
 editor = {A. Globerson and L. Mackey and D. Belgrave and A. Fan and U. Paquet and J. Tomczak and C. Zhang},
 pages = {64717--64734},
 publisher = {Curran Associates, Inc.},
 title = {Faster Neighborhood Attention: Reducing the O(n\^{}2) Cost of Self Attention at the Threadblock Level},
 url = {https://proceedings.neurips.cc/paper_files/paper/2024/file/76e952a4e83d97186d3f55eef6a3a367-Paper-Conference.pdf},
 volume = {37},
 year = {2024}
}

@inproceedings{hassani2023neighborhood,
  title        = {Neighborhood Attention Transformer},
  author       = {Ali Hassani and Steven Walton and Jiachen Li and Shen Li and Humphrey Shi},
  year         = 2023,
  booktitle    = {IEEE/CVF Conference on Computer Vision and Pattern Recognition (CVPR)}
}

@misc{hassani2022dilated,
  title        = {Dilated Neighborhood Attention Transformer},
  author       = {Ali Hassani and Humphrey Shi},
  year         = 2022,
  url          = {https://arxiv.org/abs/2209.15001},
  eprint       = {2209.15001},
  archiveprefix = {arXiv},
  primaryclass = {cs.CV}
}

@inproceedings{2023_PNA,
author = {Liu, Ting and Wei, Yunchao and Zhang, Yanning},
title = {Progressive neighborhood aggregation for semantic segmentation refinement},
year = {2023},
isbn = {978-1-57735-880-0},
publisher = {AAAI Press},
url = {https://doi.org/10.1609/aaai.v37i2.25262},
doi = {10.1609/aaai.v37i2.25262},
booktitle = {Proceedings of the Thirty-Seventh AAAI Conference on Artificial Intelligence and Thirty-Fifth Conference on Innovative Applications of Artificial Intelligence and Thirteenth Symposium on Educational Advances in Artificial Intelligence},
articleno = {193},
numpages = {9},
series = {AAAI'23/IAAI'23/EAAI'23}
}

@misc{2023_MiniGPT,
      title={MiniGPT-4: Enhancing Vision-Language Understanding with Advanced Large Language Models}, 
      author={Deyao Zhu and Jun Chen and Xiaoqian Shen and Xiang Li and Mohamed Elhoseiny},
      year={2023},
      eprint={2304.10592},
      archivePrefix={arXiv},
      primaryClass={cs.CV},
      url={https://arxiv.org/abs/2304.10592}, 
}

@inproceedings{2023_BLIP2,
author = {Li, Junnan and Li, Dongxu and Savarese, Silvio and Hoi, Steven},
title = {BLIP-2: bootstrapping language-image pre-training with frozen image encoders and large language models},
year = {2023},
publisher = {JMLR.org},
booktitle = {Proceedings of the 40th International Conference on Machine Learning},
articleno = {814},
numpages = {13},
location = {Honolulu, Hawaii, USA},
series = {ICML'23}
}

@inproceedings{2023_InstructBLIP,
 author = {Dai, Wenliang and Li, Junnan and LI, DONGXU and Tiong, Anthony and Zhao, Junqi and Wang, Weisheng and Li, Boyang and Fung, Pascale N and Hoi, Steven},
 booktitle = {Advances in Neural Information Processing Systems},
 editor = {A. Oh and T. Naumann and A. Globerson and K. Saenko and M. Hardt and S. Levine},
 pages = {49250--49267},
 publisher = {Curran Associates, Inc.},
 title = {InstructBLIP: Towards General-purpose Vision-Language Models with Instruction Tuning},
 url = {https://proceedings.neurips.cc/paper_files/paper/2023/file/9a6a435e75419a836fe47ab6793623e6-Paper-Conference.pdf},
 volume = {36},
 year = {2023}
}

@inproceedings{2023_LLaVA,
author = {Liu, Haotian and Li, Chunyuan and Wu, Qingyang and Lee, Yong Jae},
title = {Visual instruction tuning},
year = {2023},
publisher = {Curran Associates Inc.},
address = {Red Hook, NY, USA},
booktitle = {Proceedings of the 37th International Conference on Neural Information Processing Systems},
articleno = {1516},
numpages = {25},
location = {New Orleans, LA, USA},
series = {NIPS '23}
}

@INPROCEEDINGS{2024_LLaVA_2,
  author={Liu, Haotian and Li, Chunyuan and Li, Yuheng and Lee, Yong Jae},
  booktitle={2024 IEEE/CVF Conference on Computer Vision and Pattern Recognition (CVPR)}, 
  title={Improved Baselines with Visual Instruction Tuning}, 
  year={2024},
  volume={},
  number={},
  pages={26286-26296},
  doi={10.1109/CVPR52733.2024.02484}}

@misc{2024_GPT4,
      title={GPT-4 Technical Report}, 
      author={OpenAI and Josh Achiam and Steven Adler and Sandhini Agarwal and Lama Ahmad and Ilge Akkaya and Florencia Leoni Aleman and Diogo Almeida and Janko Altenschmidt and Sam Altman and Shyamal Anadkat and Red Avila and Igor Babuschkin and Suchir Balaji and Valerie Balcom and Paul Baltescu and Haiming Bao and Mohammad Bavarian and Jeff Belgum and Irwan Bello and Jake Berdine and Gabriel Bernadett-Shapiro and Christopher Berner and Lenny Bogdonoff and Oleg Boiko and Madelaine Boyd and Anna-Luisa Brakman and Greg Brockman and Tim Brooks and Miles Brundage and Kevin Button and Trevor Cai and Rosie Campbell and Andrew Cann and Brittany Carey and Chelsea Carlson and Rory Carmichael and Brooke Chan and Che Chang and Fotis Chantzis and Derek Chen and Sully Chen and Ruby Chen and Jason Chen and Mark Chen and Ben Chess and Chester Cho and Casey Chu and Hyung Won Chung and Dave Cummings and Jeremiah Currier and Yunxing Dai and Cory Decareaux and Thomas Degry and Noah Deutsch and Damien Deville and Arka Dhar and David Dohan and Steve Dowling and Sheila Dunning and Adrien Ecoffet and Atty Eleti and Tyna Eloundou and David Farhi and Liam Fedus and Niko Felix and Simón Posada Fishman and Juston Forte and Isabella Fulford and Leo Gao and Elie Georges and Christian Gibson and Vik Goel and Tarun Gogineni and Gabriel Goh and Rapha Gontijo-Lopes and Jonathan Gordon and Morgan Grafstein and Scott Gray and Ryan Greene and Joshua Gross and Shixiang Shane Gu and Yufei Guo and Chris Hallacy and Jesse Han and Jeff Harris and Yuchen He and Mike Heaton and Johannes Heidecke and Chris Hesse and Alan Hickey and Wade Hickey and Peter Hoeschele and Brandon Houghton and Kenny Hsu and Shengli Hu and Xin Hu and Joost Huizinga and Shantanu Jain and Shawn Jain and Joanne Jang and Angela Jiang and Roger Jiang and Haozhun Jin and Denny Jin and Shino Jomoto and Billie Jonn and Heewoo Jun and Tomer Kaftan and Łukasz Kaiser and Ali Kamali and Ingmar Kanitscheider and Nitish Shirish Keskar and Tabarak Khan and Logan Kilpatrick and Jong Wook Kim and Christina Kim and Yongjik Kim and Jan Hendrik Kirchner and Jamie Kiros and Matt Knight and Daniel Kokotajlo and Łukasz Kondraciuk and Andrew Kondrich and Aris Konstantinidis and Kyle Kosic and Gretchen Krueger and Vishal Kuo and Michael Lampe and Ikai Lan and Teddy Lee and Jan Leike and Jade Leung and Daniel Levy and Chak Ming Li and Rachel Lim and Molly Lin and Stephanie Lin and Mateusz Litwin and Theresa Lopez and Ryan Lowe and Patricia Lue and Anna Makanju and Kim Malfacini and Sam Manning and Todor Markov and Yaniv Markovski and Bianca Martin and Katie Mayer and Andrew Mayne and Bob McGrew and Scott Mayer McKinney and Christine McLeavey and Paul McMillan and Jake McNeil and David Medina and Aalok Mehta and Jacob Menick and Luke Metz and Andrey Mishchenko and Pamela Mishkin and Vinnie Monaco and Evan Morikawa and Daniel Mossing and Tong Mu and Mira Murati and Oleg Murk and David Mély and Ashvin Nair and Reiichiro Nakano and Rajeev Nayak and Arvind Neelakantan and Richard Ngo and Hyeonwoo Noh and Long Ouyang and Cullen O'Keefe and Jakub Pachocki and Alex Paino and Joe Palermo and Ashley Pantuliano and Giambattista Parascandolo and Joel Parish and Emy Parparita and Alex Passos and Mikhail Pavlov and Andrew Peng and Adam Perelman and Filipe de Avila Belbute Peres and Michael Petrov and Henrique Ponde de Oliveira Pinto and Michael and Pokorny and Michelle Pokrass and Vitchyr H. Pong and Tolly Powell and Alethea Power and Boris Power and Elizabeth Proehl and Raul Puri and Alec Radford and Jack Rae and Aditya Ramesh and Cameron Raymond and Francis Real and Kendra Rimbach and Carl Ross and Bob Rotsted and Henri Roussez and Nick Ryder and Mario Saltarelli and Ted Sanders and Shibani Santurkar and Girish Sastry and Heather Schmidt and David Schnurr and John Schulman and Daniel Selsam and Kyla Sheppard and Toki Sherbakov and Jessica Shieh and Sarah Shoker and Pranav Shyam and Szymon Sidor and Eric Sigler and Maddie Simens and Jordan Sitkin and Katarina Slama and Ian Sohl and Benjamin Sokolowsky and Yang Song and Natalie Staudacher and Felipe Petroski Such and Natalie Summers and Ilya Sutskever and Jie Tang and Nikolas Tezak and Madeleine B. Thompson and Phil Tillet and Amin Tootoonchian and Elizabeth Tseng and Preston Tuggle and Nick Turley and Jerry Tworek and Juan Felipe Cerón Uribe and Andrea Vallone and Arun Vijayvergiya and Chelsea Voss and Carroll Wainwright and Justin Jay Wang and Alvin Wang and Ben Wang and Jonathan Ward and Jason Wei and CJ Weinmann and Akila Welihinda and Peter Welinder and Jiayi Weng and Lilian Weng and Matt Wiethoff and Dave Willner and Clemens Winter and Samuel Wolrich and Hannah Wong and Lauren Workman and Sherwin Wu and Jeff Wu and Michael Wu and Kai Xiao and Tao Xu and Sarah Yoo and Kevin Yu and Qiming Yuan and Wojciech Zaremba and Rowan Zellers and Chong Zhang and Marvin Zhang and Shengjia Zhao and Tianhao Zheng and Juntang Zhuang and William Zhuk and Barret Zoph},
      year={2024},
      eprint={2303.08774},
      archivePrefix={arXiv},
      primaryClass={cs.CL},
      url={https://arxiv.org/abs/2303.08774}, 
}

@article{2022_LSEG,
  author       = {Boyi Li and
                  Kilian Q. Weinberger and
                  Serge J. Belongie and
                  Vladlen Koltun and
                  Ren{\'{e}} Ranftl},
  title        = {Language-driven Semantic Segmentation},
  journal      = {CoRR},
  volume       = {abs/2201.03546},
  year         = {2022},
  url          = {https://arxiv.org/abs/2201.03546},
  eprinttype    = {arXiv},
  eprint       = {2201.03546},
  bibsource    = {dblp computer science bibliography, https://dblp.org}
}

@article{2023_GDINO,
  title={Grounding dino: Marrying dino with grounded pre-training for open-set object detection},
  author={Liu, Shilong and Zeng, Zhaoyang and Ren, Tianhe and Li, Feng and Zhang, Hao and Yang, Jie and Li, Chunyuan and Yang, Jianwei and Su, Hang and Zhu, Jun and others},
  journal={arXiv preprint arXiv:2303.05499},
  year={2023}
}

@article{2024_GSAM,
  title={Grounded SAM: Assembling Open-World Models for Diverse Visual Tasks},
  author={Tianhe Ren and Shilong Liu and Ailing Zeng and Jing Lin and Kunchang Li and He Cao and Jiayu Chen and Xinyu Huang and Yukang Chen and Feng Yan and Zhaoyang Zeng and Hao Zhang and Feng Li and Jie Yang and Hongyang Li and Qing Jiang and Lei Zhang},
  journal={ArXiv},
  year={2024},
  volume={abs/2401.14159},
  url={https://api.semanticscholar.org/CorpusID:267212047}
}

@misc{2024_Text_IF,
      title={Text-IF: Leveraging Semantic Text Guidance for Degradation-Aware and Interactive Image Fusion}, 
      author={Xunpeng Yi and Han Xu and Hao Zhang and Linfeng Tang and Jiayi Ma},
      year={2024},
      eprint={2403.16387},
      archivePrefix={arXiv},
      primaryClass={cs.CV},
      url={https://arxiv.org/abs/2403.16387}, 
}

@misc{2024_MultiAtt_RSSC,
      title={Multimodal Remote Sensing Scene Classification Using VLMs and Dual-Cross Attention Networks}, 
      author={Jinjin Cai and Kexin Meng and Baijian Yang and Gang Shao},
      year={2024},
      eprint={2412.02531},
      archivePrefix={arXiv},
      primaryClass={cs.CV},
      url={https://arxiv.org/abs/2412.02531}, 
}

@InProceedings{2024_PaSCo,
      title={PaSCo: Urban 3D Panoptic Scene Completion with Uncertainty Awareness}, 
      author={Anh-Quan Cao and Angela Dai and Raoul de Charette},
      year={2024},
      booktitle = {CVPR}
}

@article{2024_HybridOcc,
   title={HybridOcc: NeRF Enhanced Transformer-Based Multi-Camera 3D Occupancy Prediction},
   volume={9},
   ISSN={2377-3774},
   url={http://dx.doi.org/10.1109/LRA.2024.3416798},
   DOI={10.1109/lra.2024.3416798},
   number={9},
   journal={IEEE Robotics and Automation Letters},
   publisher={Institute of Electrical and Electronics Engineers (IEEE)},
   author={Zhao, Xiao and Chen, Bo and Sun, Mingyang and Yang, Dingkang and Wang, Youxing and Zhang, Xukun and Li, Mingcheng and Kou, Dongliang and Wei, Xiaoyi and Zhang, Lihua},
   year={2024},
   month=sep, pages={7867–7874} }

@article{2016_InstanceNorm3D,
  author       = {Dmitry Ulyanov and
                  Andrea Vedaldi and
                  Victor S. Lempitsky},
  title        = {Instance Normalization: The Missing Ingredient for Fast Stylization},
  journal      = {CoRR},
  volume       = {abs/1607.08022},
  year         = {2016},
  url          = {http://arxiv.org/abs/1607.08022},
  eprinttype    = {arXiv},
  eprint       = {1607.08022},
  bibsource    = {dblp computer science bibliography, https://dblp.org}
}

@article{2015_LeakyReLU,
  author       = {Bing Xu and
                  Naiyan Wang and
                  Tianqi Chen and
                  Mu Li},
  title        = {Empirical Evaluation of Rectified Activations in Convolutional Network},
  journal      = {CoRR},
  volume       = {abs/1505.00853},
  year         = {2015},
  url          = {http://arxiv.org/abs/1505.00853},
  eprinttype    = {arXiv},
  eprint       = {1505.00853},
  bibsource    = {dblp computer science bibliography, https://dblp.org}
}

@article{2025_VLME2E,
  title={VLM-E2E: Enhancing End-to-End Autonomous Driving with Multimodal Driver Attention Fusion},
  author={Pei Liu and Haipeng Liu and Haichao Liu and Xin Liu and Jinxin Ni and Jun Ma},
  journal={ArXiv},
  year={2025},
  volume={abs/2502.18042},
  url={https://api.semanticscholar.org/CorpusID:276580793}
}

@inproceedings{2024_LongCLIP,
  title     = {Long-CLIP: Unlocking the Long-Text Capability of CLIP},
  author    = {Zhang, B. and Zhang, P. and Dong, X. and Zang, Y. and Wang, J.},
  booktitle = {Computer Vision -- ECCV 2024},
  editor    = {Leonardis, A. and Ricci, E. and Roth, S. and Russakovsky, O. and Sattler, T. and Varol, G.},
  series    = {Lecture Notes in Computer Science},
  volume    = {15109},
  publisher = {Springer},
  address   = {Cham},
  year      = {2024},
  doi       = {10.1007/978-3-031-72983-6_18},
  url       = {https://doi.org/10.1007/978-3-031-72983-6_18}
}

@misc{2023_EVACLIP,
      title={EVA-CLIP: Improved Training Techniques for CLIP at Scale}, 
      author={Quan Sun and Yuxin Fang and Ledell Wu and Xinlong Wang and Yue Cao},
      year={2023},
      eprint={2303.15389},
      archivePrefix={arXiv},
      primaryClass={cs.CV},
      url={https://arxiv.org/abs/2303.15389}, 
}

@misc{2024_EVACLIP,
      title={EVA-CLIP-18B: Scaling CLIP to 18 Billion Parameters}, 
      author={Quan Sun and Jinsheng Wang and Qiying Yu and Yufeng Cui and Fan Zhang and Xiaosong Zhang and Xinlong Wang},
      year={2024},
      eprint={2402.04252},
      archivePrefix={arXiv},
      primaryClass={cs.CV},
      url={https://arxiv.org/abs/2402.04252}, 
}

@inproceedings{
2024_JinaCLIP,
title={Jina {CLIP}: Your {CLIP} Model Is Also Your Text Retriever},
author={Han Xiao and Georgios Mastrapas and Bo Wang},
booktitle={Multi-modal Foundation Model meets Embodied AI Workshop @ ICML2024},
year={2024},
url={https://openreview.net/forum?id=lSDkG98goM}
}

@misc{2024_JinaCLIP_2,
      title={jina-clip-v2: Multilingual Multimodal Embeddings for Text and Images}, 
      author={Andreas Koukounas and Georgios Mastrapas and Bo Wang and Mohammad Kalim Akram and Sedigheh Eslami and Michael Günther and Isabelle Mohr and Saba Sturua and Scott Martens and Nan Wang and Han Xiao},
      year={2024},
      eprint={2412.08802},
      archivePrefix={arXiv},
      primaryClass={cs.CL},
      url={https://arxiv.org/abs/2412.08802}, 
}

@article{2023_SAM,
  title={Segment Anything},
  author={Kirillov, Alexander and Mintun, Eric and Ravi, Nikhila and Mao, Hanzi and Rolland, Chloe and Gustafson, Laura and Xiao, Tete and Whitehead, Spencer and Berg, Alexander C. and Lo, Wan-Yen and Doll{\'a}r, Piotr and Girshick, Ross},
  journal={arXiv:2304.02643},
  year={2023}
}

@article{2021_SSASC,
  author       = {Xuemeng Yang and
                  Hao Zou and
                  Xin Kong and
                  Tianxin Huang and
                  Yong Liu and
                  Wanlong Li and
                  Feng Wen and
                  Hongbo Zhang},
  title        = {Semantic Segmentation-assisted Scene Completion for LiDAR Point Clouds},
  journal      = {CoRR},
  volume       = {abs/2109.11453},
  year         = {2021},
  url          = {https://arxiv.org/abs/2109.11453},
  eprinttype    = {arXiv},
  eprint       = {2109.11453},
  bibsource    = {dblp computer science bibliography, https://dblp.org}
}

@InProceedings{2021_CLIP,
  title = 	 {Learning Transferable Visual Models From Natural Language Supervision},
  author =       {Radford, Alec and Kim, Jong Wook and Hallacy, Chris and Ramesh, Aditya and Goh, Gabriel and Agarwal, Sandhini and Sastry, Girish and Askell, Amanda and Mishkin, Pamela and Clark, Jack and Krueger, Gretchen and Sutskever, Ilya},
  booktitle = 	 {Proceedings of the 38th International Conference on Machine Learning},
  pages = 	 {8748--8763},
  year = 	 {2021},
  editor = 	 {Meila, Marina and Zhang, Tong},
  volume = 	 {139},
  series = 	 {Proceedings of Machine Learning Research},
  month = 	 {18--24 Jul},
  publisher =    {PMLR},
  url = 	 {https://proceedings.mlr.press/v139/radford21a.html}
}

@Article{2022_LiDAR_Survey,
AUTHOR = {Danni Wu and Zichen Liang and Guang Chen},
TITLE = {Deep learning for LiDAR-only and LiDAR-fusion 3D perception: a survey},
JOURNAL = {Intelligence \& Robotics},
VOLUME = {2},
YEAR = {2022},
NUMBER = {2},
ARTICLE-NUMBER = {105-29},
URL = {https://www.oaepublish.com/articles/ir.2021.20},
ISSN = {2770-3541},
DOI = {10.20517/ir.2021.20}
}

@ARTICLE{2025_LiDAR_Survey,
  author={Lin, Shih-Lin and Wu, Jun-Yi},
  journal={IEEE Access}, 
  title={Enhancing Lidar-Based 3D Classification Through an Improved Deep Learning Framework With Residual Connections}, 
  year={2025},
  volume={13},
  number={},
  pages={42836-42849},
  doi={10.1109/ACCESS.2025.3547942}
}

@article{2016_PointNet,
  author       = {Charles Ruizhongtai Qi and
                  Hao Su and
                  Kaichun Mo and
                  Leonidas J. Guibas},
  title        = {PointNet: Deep Learning on Point Sets for 3D Classification and Segmentation},
  journal      = {CoRR},
  volume       = {abs/1612.00593},
  year         = {2016},
  url          = {http://arxiv.org/abs/1612.00593},
  eprinttype    = {arXiv},
  eprint       = {1612.00593},
  bibsource    = {dblp computer science bibliography, https://dblp.org}
}

@article{2017_PointNet_pp,
  author       = {Charles Ruizhongtai Qi and
                  Li Yi and
                  Hao Su and
                  Leonidas J. Guibas},
  title        = {PointNet++: Deep Hierarchical Feature Learning on Point Sets in a
                  Metric Space},
  journal      = {CoRR},
  volume       = {abs/1706.02413},
  year         = {2017},
  url          = {http://arxiv.org/abs/1706.02413},
  eprinttype    = {arXiv},
  eprint       = {1706.02413},
  bibsource    = {dblp computer science bibliography, https://dblp.org}
}

@Article{2018_SECOND,
AUTHOR = {Yan, Yan and Mao, Yuxing and Li, Bo},
TITLE = {SECOND: Sparsely Embedded Convolutional Detection},
JOURNAL = {Sensors},
VOLUME = {18},
YEAR = {2018},
NUMBER = {10},
ARTICLE-NUMBER = {3337},
URL = {https://www.mdpi.com/1424-8220/18/10/3337},
PubMedID = {30301196},
ISSN = {1424-8220},
DOI = {10.3390/s18103337}
}

@article{2018_PointPillars,
  author       = {Alex H. Lang and
                  Sourabh Vora and
                  Holger Caesar and
                  Lubing Zhou and
                  Jiong Yang and
                  Oscar Beijbom},
  title        = {PointPillars: Fast Encoders for Object Detection from Point Clouds},
  journal      = {CoRR},
  volume       = {abs/1812.05784},
  year         = {2018},
  url          = {http://arxiv.org/abs/1812.05784},
  eprinttype    = {arXiv},
  eprint       = {1812.05784},
  bibsource    = {dblp computer science bibliography, https://dblp.org}
}

@article{2018_PointRCNN,
  author       = {Shaoshuai Shi and
                  Xiaogang Wang and
                  Hongsheng Li},
  title        = {PointRCNN: 3D Object Proposal Generation and Detection from Point
                  Cloud},
  journal      = {CoRR},
  volume       = {abs/1812.04244},
  year         = {2018},
  url          = {http://arxiv.org/abs/1812.04244},
  eprinttype    = {arXiv},
  eprint       = {1812.04244},
  bibsource    = {dblp computer science bibliography, https://dblp.org}
}

@article{2019_PVRCNN,
  author       = {Shaoshuai Shi and
                  Chaoxu Guo and
                  Li Jiang and
                  Zhe Wang and
                  Jianping Shi and
                  Xiaogang Wang and
                  Hongsheng Li},
  title        = {{PV-RCNN:} Point-Voxel Feature Set Abstraction for 3D Object Detection},
  journal      = {CoRR},
  volume       = {abs/1912.13192},
  year         = {2019},
  url          = {http://arxiv.org/abs/1912.13192},
  eprinttype    = {arXiv},
  eprint       = {1912.13192},
  bibsource    = {dblp computer science bibliography, https://dblp.org}
}

@article{2020_VoxelRCNN,
  author       = {Jiajun Deng and
                  Shaoshuai Shi and
                  Peiwei Li and
                  Wengang Zhou and
                  Yanyong Zhang and
                  Houqiang Li},
  title        = {Voxel {R-CNN:} Towards High Performance Voxel-based 3D Object Detection},
  journal      = {CoRR},
  volume       = {abs/2012.15712},
  year         = {2020},
  url          = {https://arxiv.org/abs/2012.15712},
  eprinttype    = {arXiv},
  eprint       = {2012.15712},
  bibsource    = {dblp computer science bibliography, https://dblp.org}
}

@article{2020_AB3DMOT,
  author       = {Xinshuo Weng and
                  Jianren Wang and
                  David Held and
                  Kris Kitani},
  title        = {{AB3DMOT:} {A} Baseline for 3D Multi-Object Tracking and New Evaluation
                  Metrics},
  journal      = {CoRR},
  volume       = {abs/2008.08063},
  year         = {2020},
  url          = {https://arxiv.org/abs/2008.08063},
  eprinttype    = {arXiv},
  eprint       = {2008.08063},
  bibsource    = {dblp computer science bibliography, https://dblp.org}
}

@Article{2023_3DMOT,
AUTHOR = {Cho, Minho and Kim, Euntai},
TITLE = {3D LiDAR Multi-Object Tracking with Short-Term and Long-Term Multi-Level Associations},
JOURNAL = {Remote Sensing},
VOLUME = {15},
YEAR = {2023},
NUMBER = {23},
ARTICLE-NUMBER = {5486},
URL = {https://www.mdpi.com/2072-4292/15/23/5486},
ISSN = {2072-4292},
DOI = {10.3390/rs15235486}
}

@INPROCEEDINGS{2019_RangeNet_pp,
  author={Milioto, Andres and Vizzo, Ignacio and Behley, Jens and Stachniss, Cyrill},
  booktitle={2019 IEEE/RSJ International Conference on Intelligent Robots and Systems (IROS)}, 
  title={RangeNet ++: Fast and Accurate LiDAR Semantic Segmentation}, 
  year={2019},
  volume={},
  number={},
  pages={4213-4220},
  doi={10.1109/IROS40897.2019.8967762}}

@article{2020_Cylinder3D,
  author       = {Hui Zhou and
                  Xinge Zhu and
                  Xiao Song and
                  Yuexin Ma and
                  Zhe Wang and
                  Hongsheng Li and
                  Dahua Lin},
  title        = {Cylinder3D: An Effective 3D Framework for Driving-scene LiDAR Semantic
                  Segmentation},
  journal      = {CoRR},
  volume       = {abs/2008.01550},
  year         = {2020},
  url          = {https://arxiv.org/abs/2008.01550},
  eprinttype    = {arXiv},
  eprint       = {2008.01550},
  bibsource    = {dblp computer science bibliography, https://dblp.org}
}

@InProceedings{2012_NYUDV2,
author="Silberman, Nathan
and Hoiem, Derek
and Kohli, Pushmeet
and Fergus, Rob",
editor="Fitzgibbon, Andrew
and Lazebnik, Svetlana
and Perona, Pietro
and Sato, Yoichi
and Schmid, Cordelia",
title="Indoor Segmentation and Support Inference from RGBD Images",
booktitle="Computer Vision -- ECCV 2012",
year="2012",
publisher="Springer Berlin Heidelberg",
address="Berlin, Heidelberg",
pages="746--760",
isbn="978-3-642-33715-4"
}

@misc{2024_SSCBench,
      title={SSCBench: A Large-Scale 3D Semantic Scene Completion Benchmark for Autonomous Driving}, 
      author={Yiming Li and Sihang Li and Xinhao Liu and Moonjun Gong and Kenan Li and Nuo Chen and Zijun Wang and Zhiheng Li and Tao Jiang and Fisher Yu and Yue Wang and Hang Zhao and Zhiding Yu and Chen Feng},
      year={2024},
      eprint={2306.09001},
      archivePrefix={arXiv},
      primaryClass={cs.CV}, 
      doi = {10.1109/IROS58592.2024.10802143}
}

@article{2024_MMAF,
    title={Multimodal Alignment and Fusion: A Survey},
    author={Li, Songtao and Tang, Hao},
    url={https://arxiv.org/abs/2411.17040},
    doi={10.48550/ARXIV.2411.17040},
    publisher={arXiv},
    year={2024},
    month={Jan},
}

@ARTICLE{2020_CMA,
  author={Xu, Xing and Wang, Tan and Yang, Yang and Zuo, Lin and Shen, Fumin and Shen, Heng Tao},
  journal={IEEE Transactions on Neural Networks and Learning Systems}, 
  title={Cross-Modal Attention With Semantic Consistence for Image–Text Matching}, 
  year={2020},
  volume={31},
  number={12},
  pages={5412-5425},
  doi={10.1109/TNNLS.2020.2967597}}

@article{2025_DynamicRepresentations,
title = {Vision-based 3D semantic scene completion via capture dynamic representations},
journal = {Knowledge-Based Systems},
volume = {331},
pages = {114550},
year = {2026},
issn = {0950-7051},
doi = {https://doi.org/10.1016/j.knosys.2025.114550},
url = {https://www.sciencedirect.com/science/article/pii/S0950705125015898},
author = {Meng Wang and Fan Wu and Yunchuan Qin and Ruihui Li and Zhuo Tang and Kenli Li}
}

@misc{2025_VLScene,
      title={VLScene: Vision-Language Guidance Distillation for Camera-Based 3D Semantic Scene Completion}, 
      author={Meng Wang and Huilong Pi and Ruihui Li and Yunchuan Qin and Zhuo Tang and Kenli Li},
      year={2025},
      eprint={2503.06219},
      archivePrefix={arXiv},
      primaryClass={cs.CV},
      url={https://arxiv.org/abs/2503.06219}, 
}

@article{2017_Lovasz,
  author       = {Maxim Berman and
                  Matthew B. Blaschko},
  title        = {Optimization of the Jaccard index for image segmentation with the
                  Lov{\'{a}}sz hinge},
  journal      = {CoRR},
  volume       = {abs/1705.08790},
  year         = {2017},
  url          = {http://arxiv.org/abs/1705.08790},
  eprinttype    = {arXiv},
  eprint       = {1705.08790},
  bibsource    = {dblp computer science bibliography, https://dblp.org}
}

@article{2025_SkipMamba, 
    title={Skip Mamba Diffusion for Monocular 3D Semantic Scene Completion}, 
    volume={39}, 
    url={https://ojs.aaai.org/index.php/AAAI/article/view/32547}, 
    DOI={10.1609/aaai.v39i5.32547}, 
    number={5}, 
    journal={Proceedings of the AAAI Conference on Artificial Intelligence}, 
    author={Liang, Li and Akhtar, Naveed and Vice, Jordan and Kong, Xiangrui and Mian, Ajmal Saeed}, 
    year={2025}, 
    month={Apr.}, 
    pages={5155-5163} 
}

@inproceedings{2025_ProtoOcc,
  title={Protoocc: Accurate, efficient 3d occupancy prediction using dual branch encoder-prototype query decoder},
  author={Kim, Jungho and Kang, Changwon and Lee, Dongyoung and Choi, Sehwan and Choi, Jun Won},
  booktitle={Proceedings of the AAAI Conference on Artificial Intelligence},
  volume={39},
  pages={4284--4292},
  year={2025}
}

@article{2025_BiPVL,
  title={BiPVL-Seg: Bidirectional Progressive Vision-Language Fusion with Global-Local Alignment for Medical Image Segmentation},
  author={Sultan, Rafi Ibn and Zhu, Hui and Li, Chengyin and Zhu, Dongxiao},
  journal={arXiv preprint arXiv:2503.23534},
  year={2025}
}

%%% Make sure to upload the bib file along with the tex file and PDF
%%% Please see the test.bib file for some examples of references

\end{document}

\onecolumn
\firstpage{1}

\title[Supplementary Material]{{\helveticaitalic{Supplementary Material}}}

\maketitle
\section{Ablation Study}\label{chapter:AblationStudy}

This appendix reports additional ablation studies conducted to understand the effect of individual components in ESSC-RM. We investigate (i) feature enhancement block (FEB) design, (ii) multi-scale supervision, (iii) choices of vision--language models (VLMs), text semantic encoders, and fusion strategies, and (iv) the influence of learning rate and loss combinations. All experiments use CGFormer~\cite{2023_CGFormer} as the base semantic scene completion model unless otherwise specified.

\subsection{Feature Enhancement Block (FEB)}

ESSC-RM employs a 3D U-Net encoder with FEBs following SemCity~\cite{2024_SemCity}. Each FEB consists of an initial ConvBlock, a spatial downsampling layer, and an optional post-downsampling ConvBlock. To analyze how these architectural choices affect feature quality, two dimensions are ablated: (1) MaxPooling vs.\ convolutional downsampling, and (2) whether a second ConvBlock is appended. 

Table~\ref{tab:FEB} shows that convolutional downsampling provides slightly higher mIoU than MaxPooling, and adding a second ConvBlock further improves performance for both 3D U-Net and PNAM variants. These gains come at a small cost in IoU and computational overhead, indicating a trade-off between representational richness and efficiency.

\begin{table}[htbp]
  \centering
  \begin{tabular}{@{}c|cccc|cc@{}}
  \toprule
  \multirow{2}{*}{Model} & \multicolumn{4}{c|}{FEB Components}  & \multirow{2}{*}{IoU} & \multirow{2}{*}{mIoU} \\
                         & ConvBlock1 & MaxPooling & ConvDown & ConvBlock2 & & \\ \midrule
  CGFormer + 3D U-Net    & \checkmark & \checkmark &          &            & 44.15 & 17.03 \\
  CGFormer + 3D U-Net    & \checkmark &            & \checkmark &          & 43.48 & 17.06 \\
  CGFormer + 3D U-Net    & \checkmark & \checkmark &          & \checkmark & 43.43 & 17.13 \\
  CGFormer + 3D U-Net    & \checkmark &            & \checkmark & \checkmark & 43.53 & \textbf{17.17} \\ \midrule
  CGFormer + PNAM        & \checkmark & \checkmark &          &            & 44.85 & 17.19 \\
  CGFormer + PNAM        & \checkmark &            & \checkmark &          & 44.05 & 17.09 \\
  CGFormer + PNAM        & \checkmark & \checkmark &          & \checkmark & 44.88 & 17.26 \\
  CGFormer + PNAM        & \checkmark &            & \checkmark & \checkmark & 44.33 & \textbf{17.27} \\ \bottomrule
  \end{tabular}
  \caption{Ablation study on the Feature Enhancement Block (FEB) following SemCity~\cite{2024_SemCity}.}
  \label{tab:FEB}
\end{table}

\subsection{Multi-scale Supervision}

We evaluate whether supervising voxel predictions at multiple resolutions improves semantic consistency. Following the multi-scale design in PaSCo~\cite{2024_PaSCo}, we compare training with only $1\!:\!1$ supervision against multi-scale supervision at $\{1\!:\!8,1\!:\!4,1\!:\!2,1\!:\!1\}$.

Across all refinement modules, multi-scale supervision improves mIoU (e.g., 16.87\%$\rightarrow$17.21\% for VLGM and 17.07\%$\rightarrow$17.27\% for PNAM), while IoU decreases slightly. This suggests that supervising intermediate resolutions strengthens fine-grained semantic discrimination.

\begin{table}[htbp]
  \centering
  \begin{tabular}{@{}c|cccc|cc@{}}
  \toprule
  \multirow{2}{*}{Model} & \multicolumn{4}{c|}{Multi-scale Supervision} 
                         & \multirow{2}{*}{IoU} & \multirow{2}{*}{mIoU} \\
                         & 1:8 & 1:4 & 1:2 & 1:1 & & \\ \midrule
  CGFormer + 3D U-Net &  &  &  & \checkmark & 44.21 & 16.98 \\
  CGFormer + 3D U-Net & \checkmark & \checkmark & \checkmark & \checkmark & 43.53 & \textbf{17.17} \\ \midrule
  CGFormer + VLGM     &  &  &  & \checkmark & 44.09 & 16.87 \\
  CGFormer + VLGM     & \checkmark & \checkmark & \checkmark & \checkmark & 43.20 & \textbf{17.21} \\ \midrule
  CGFormer + PNAM     &  &  &  & \checkmark & 44.44 & 17.07 \\
  CGFormer + PNAM     & \checkmark & \checkmark & \checkmark & \checkmark & 44.33 & \textbf{17.27} \\ \bottomrule
  \end{tabular}
  \caption{Ablation study on multi-scale supervision following PaSCo~\cite{2024_PaSCo}.}
  \label{tab:MSS}
\end{table}

\subsection{Text Annotation Generation}

\begin{figure}[htbp]
  \centering
  \includegraphics[width=\textwidth]{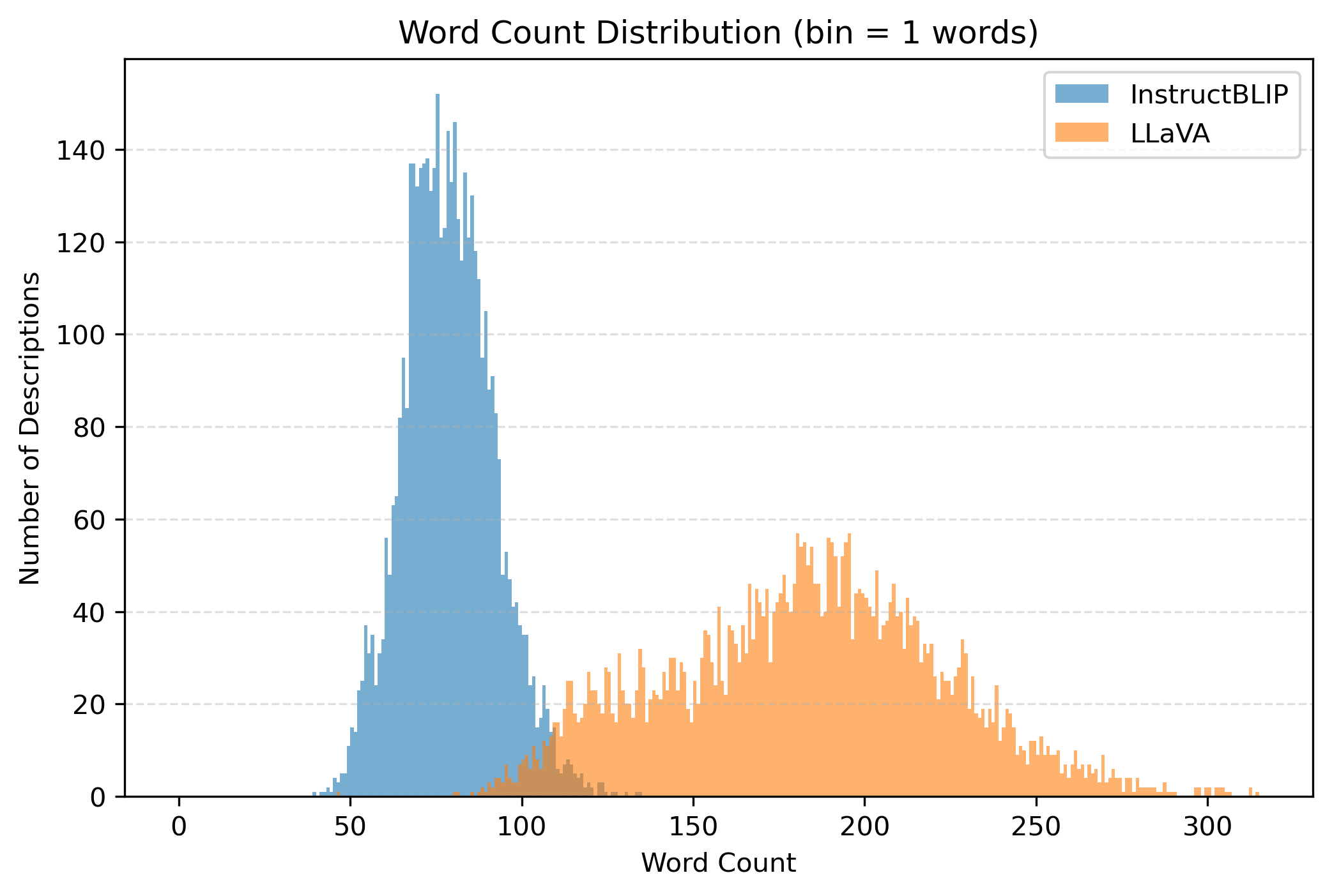}
  \caption{Word count distribution of text outputs from InstructBLIP~\cite{2023_InstructBLIP} and LLaVA~\cite{2023_LLaVA,2024_LLaVA_2}.}
  \label{fig:WordDistribution}
\end{figure}

Two VLMs are compared for generating scene descriptions: InstructBLIP~\cite{2023_InstructBLIP} and LLaVA~\cite{2023_LLaVA,2024_LLaVA_2}. As shown in Figure~\ref{fig:WordDistribution} and Table~\ref{tab:VLM}, LLaVA generates considerably longer descriptions (average 182 words vs.\ 78), often with narrative-style content, whereas InstructBLIP produces concise object-focused descriptions with more accurate attributes.

InstructBLIP is based on BLIP-2~\cite{2023_BLIP2} with Vicuna-7B as the LLM, while LLaVA uses a ViT-based visual encoder and a larger Yi-34B LLM. The larger language model in LLaVA results in much higher inference cost (42.25s vs.\ 13.67s per frame). Therefore, InstructBLIP is preferable for on-the-fly annotation, while LLaVA is mainly used offline to evaluate the effect of richer textual priors.

\begin{table}[htbp]
  \centering
  \begin{tabular}{@{}c|cc|c@{}}
  \toprule
  VLM & Vision Encoder & LLM & Inference Time \\ \midrule
  InstructBLIP & ViT-G/14 & Vicuna-7B & 13.67s \\
  LLaVA        & ViT-L/14 & Yi-34B    & 42.25s \\ \bottomrule
  \end{tabular}
  \caption{ Comparison of VLM architectures and inference latency for InstructBLIP~\cite{2023_InstructBLIP} and LLaVA~\cite{2023_LLaVA,2024_LLaVA_2}.}
  \label{tab:VLM}
\end{table}

\subsection{Text Semantic Encoder}

We evaluate several text encoders using texts from both VLMs: Q-Former from BLIP-2~\cite{2023_BLIP2}, CLIP~\cite{2021_CLIP}, LongCLIP~\cite{2024_LongCLIP}, and JinaCLIP/JinaCLIP\_1024~\cite{2024_JinaCLIP,2024_JinaCLIP_2}. Table~\ref{tab:TSE} shows that Q-Former combined with DCAM yields the highest mIoU for both InstructBLIP (17.18\%) and LLaVA (17.21\%). Higher-dimensional encoders do not consistently improve mIoU, suggesting that fusion quality matters more than feature dimensionality.

\begin{table}[htbp]
  \centering
  \resizebox{\textwidth}{!}{%
  \begin{tabular}{@{}c|c|cc|c|cc@{}}
  \toprule
  Baseline & VLM & Text Encoder & Feature Dim. & Fuser & IoU & mIoU \\ \midrule
  \multirow{5}{*}{CGFormer} 
      & InstructBLIP & Q-Former~\cite{2023_BLIP2} & $N \times L \times 256$ & DCAM & 44.02 & \textbf{17.18} \\
      &              & CLIP~\cite{2021_CLIP}      & $N \times 512$ & SIGM & 44.10 & 17.11 \\
      &              & LongCLIP~\cite{2024_LongCLIP}  & $N \times 768$ & SIGM & 44.18 & 17.04 \\
      &              & JinaCLIP~\cite{2024_JinaCLIP,2024_JinaCLIP_2} & $N \times 512$ & SIGM & 42.83 & 17.11 \\
      &              & JinaCLIP\_1024~\cite{2024_JinaCLIP,2024_JinaCLIP_2} & $N \times 1024$ & SIGM & 43.26 & 17.07 \\ \midrule
  \multirow{5}{*}{CGFormer} 
      & LLaVA~\cite{2023_LLaVA,2024_LLaVA_2} & Q-Former~\cite{2023_BLIP2} & $N \times L \times 256$ & DCAM & 43.20 & \textbf{17.21} \\
      &       & CLIP~\cite{2021_CLIP}      & -- & -- & -- & -- \\
      &       & LongCLIP~\cite{2024_LongCLIP}  & -- & -- & -- & -- \\
      &       & JinaCLIP~\cite{2024_JinaCLIP,2024_JinaCLIP_2}  & $N \times 512$ & SIGM & 43.51 & 17.11 \\
      &       & JinaCLIP\_1024~\cite{2024_JinaCLIP,2024_JinaCLIP_2} & $N \times 1024$ & SIGM & 43.88 & 17.08 \\ \bottomrule
  \end{tabular}}
  \caption{ Ablation study on text semantic encoders, including Q-Former~\cite{2023_BLIP2}, CLIP~\cite{2021_CLIP}, LongCLIP~\cite{2024_LongCLIP}, and JinaCLIP variants~\cite{2024_JinaCLIP,2024_JinaCLIP_2}.}
  \label{tab:TSE}
\end{table}

\subsection{Text Semantic Fusion Strategy}

\begin{figure}[htbp]
  \centering
  \includegraphics[width=0.8\textwidth]{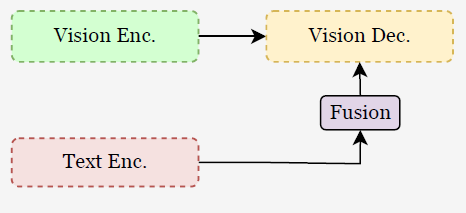}
  \caption{Illustration of text--vision fusion strategies inspired by BiPVL~\cite{2025_BiPVL}.}
  \label{fig:2x2grid}
\end{figure}

We examine where to inject text semantics within the refinement module. Figure~\ref{fig:2x2grid} illustrates representative fusion schemes following BiPVL-style designs~\cite{2025_BiPVL}. Since the text encoder is frozen, bidirectional encoder fusion is not applicable and late fusion conflicts with skip connections. We therefore test encoder-only, decoder-only, and encoder+decoder fusion.

Encoder-only fusion gives the highest IoU, whereas decoder-only fusion slightly improves mIoU. The best mIoU is obtained when fusing at both encoder and decoder, indicating complementary benefits from low-level and high-level semantic guidance (Table~\ref{tab:TSF}).

\begin{table}[htbp]
  \centering
  \begin{tabular}{@{}c|c|cc|cc@{}}
  \toprule
  Baseline & VLM & Encoder Fusion & Decoder Fusion & IoU & mIoU \\ \midrule
  CGFormer & InstructBLIP~\cite{2023_InstructBLIP} & \checkmark &  & 44.40 & 17.10 \\
           &               &  & \checkmark & 44.23 & 17.13 \\
           &               & \checkmark & \checkmark & 44.02 & \textbf{17.18} \\ \midrule
  CGFormer & LLaVA~\cite{2023_LLaVA,2024_LLaVA_2} & \checkmark &  & 43.87 & 17.08 \\
           &       &  & \checkmark & 43.55 & 17.16 \\
           &       & \checkmark & \checkmark & 43.20 & \textbf{17.21} \\ \bottomrule
  \end{tabular}
  \caption{Ablation study on text semantic fusion placement, comparing encoder-only, decoder-only, and encoder+decoder fusion strategies~\cite{2025_BiPVL}.}
  \label{tab:TSF}
\end{table}

\subsection{Learning Rate}

We evaluate three learning rates ($3\times10^{-4}$, $1\times10^{-4}$, $5\times10^{-5}$). As shown in Table~\ref{tab:LR}, all refinement modules achieve their best mIoU with the smallest learning rate. Since the refinement module updates an already well-formed 3D representation from CGFormer~\cite{2023_CGFormer}, a smaller step size helps stabilize optimization and prevents overfitting.

\begin{table}[htbp]
  \centering
  \begin{tabular}{@{}c|ccc|cc@{}}
  \toprule
  Model & 3e-4 & 1e-4 & 5e-5 & IoU & mIoU \\ \midrule
  CGFormer + 3D U-Net & \checkmark &  &  & 45.71 & 16.87 \\
  CGFormer + 3D U-Net &  & \checkmark &  & 44.72 & 16.92 \\
  CGFormer + 3D U-Net &  &  & \checkmark & 43.53 & \textbf{17.17} \\ \midrule
  CGFormer + VLGM     & \checkmark &  &  & 45.46 & 17.05 \\
  CGFormer + VLGM     &  & \checkmark &  & 44.52 & 17.03 \\
  CGFormer + VLGM     &  &  & \checkmark & 43.20 & \textbf{17.21} \\ \midrule
  CGFormer + PNAM     & \checkmark &  &  & 45.18 & 17.13 \\
  CGFormer + PNAM     &  & \checkmark &  & 44.75 & 17.16 \\
  CGFormer + PNAM     &  &  & \checkmark & 44.33 & \textbf{17.27} \\ \bottomrule
  \end{tabular}
  \caption{Ablation study on learning rate for ESSC-RM built on CGFormer~\cite{2023_CGFormer}.}
  \label{tab:LR}
\end{table}

\subsection{Loss Design}

Finally, we test the impact of adding Lovász-softmax loss~\cite{2017_Lovasz} on top of the default cross-entropy (CE) and SCAL. Lovász-softmax has been adopted in several recent SSC works~\cite{2025_DynamicRepresentations,2025_SkipMamba,2025_ProtoOcc} due to its direct optimization of an IoU surrogate. Table~\ref{tab:Loss} shows that introducing Lovász-softmax slightly decreases mIoU for all refinement modules and does not improve IoU. SCAL provides strong voxel-level consistency supervision, and additional IoU-oriented losses do not yield further gains on SemanticKITTI~\cite{2019_SemanticKITTI,2021_SemanticKITTI_2}.

\begin{table}[htbp]
  \centering
  \begin{tabular}{@{}c|ccc|cc@{}}
  \toprule
  Model & CE & SCAL & Lovász-softmax~\cite{2017_Lovasz} & IoU & mIoU \\ \midrule
  CGFormer + 3D U-Net & \checkmark & \checkmark &  & 43.53 & \textbf{17.17} \\
  CGFormer + 3D U-Net & \checkmark & \checkmark & \checkmark & 44.15 & 16.94 \\ \midrule
  CGFormer + VLGM     & \checkmark & \checkmark &  & 43.20 & \textbf{17.21} \\
  CGFormer + VLGM     & \checkmark & \checkmark & \checkmark & 42.84 & 17.12 \\ \midrule
  CGFormer + PNAM     & \checkmark & \checkmark &  & 44.33 & \textbf{17.27} \\
  CGFormer + PNAM     & \checkmark & \checkmark & \checkmark & 43.74 & 17.18 \\ \bottomrule
  \end{tabular}
  \caption{Ablation study on loss configurations, including Lovász-softmax~\cite{2017_Lovasz} as in~\cite{2025_DynamicRepresentations,2025_SkipMamba,2025_ProtoOcc}.}
  \label{tab:Loss}
\end{table}

%%% If you don't add the figures in the LaTeX files, please upload them when submitting the article.
%%% Frontiers will add the figures at the end of the provisional pdf automatically
%%% The use of LaTeX coding to draw Diagrams/Figures/Structures should be avoided. They should be external callouts including graphics.

\bibliographystyle{Frontiers-Harvard} 
\bibliography{test}